\documentclass[10pt,journal,compsoc]{IEEEtran}

\usepackage{cite}
\usepackage{amsmath,amssymb,amsfonts}
\usepackage{graphicx}
\usepackage{textcomp}
\usepackage{xcolor, array, colortbl}

\usepackage[utf8]{inputenc}
\usepackage[T1]{fontenc}    
\usepackage{url}          
\usepackage{booktabs}      
\usepackage{amsfonts}      
\usepackage{nicefrac}    
\usepackage{microtype}     

\usepackage{bbding}
\usepackage{pifont}
\usepackage{amssymb}
 
\newcommand{\xmarkg}{\textcolor{gray}{\ding{55}}\xspace}%
\newcommand{\cmark}{\ding{51}}
\newcommand{\tabincell}[2]{}
\usepackage{bm}
\usepackage{comment}
\usepackage{enumerate} 
\usepackage{enumitem}
\usepackage{mathrsfs}
\usepackage{makeidx}      
\usepackage{graphicx}       
\usepackage{multicol}       
\usepackage{subfigure}
\usepackage{epsfig,amssymb,latexsym}
\usepackage{psfrag}
\usepackage[ruled,vlined]{algorithm2e}
\definecolor{lightgray}{rgb}{.91,.91,.91}
\definecolor{deepred}{rgb}{0.698,0.133,0.133}
\definecolor{deepgreen}{rgb}{0,0.392,0}
\definecolor{deepblue}{rgb}{0,0,0.392}

\usepackage{makecell}	
\usepackage{fancyhdr}
\usepackage{multirow}
\usepackage{rotating}
\usepackage{color}
\usepackage{mathtools}
\usepackage{amsmath}
\usepackage{tensor}

\usepackage[pagebackref=false,breaklinks=true,colorlinks,bookmarks=false,linkcolor=red,urlcolor=blue,citecolor=blue]{hyperref}

\newcommand{\mf}{\mathbf}

\newcommand{\mr}{\mathrm}

\usepackage{amsthm}

\ifCLASSINFOpdf
\else
\fi

\begin{document}
\title{Crafting Your Evolving Dreams: Concept-Incremental Versatile Customization}

\author{Jiahua Dong,
        Wenqi Liang,
        Hongliu Li,
        Yang Cong,
        Duzhen Zhang,
        Hanbin Zhao,  \\
        Henghui Ding, 
        Yulun Zhang,  
        Salman Khan,
        Fahad Shahbaz Khan

\IEEEcompsocitemizethanks{
\IEEEcompsocthanksitem Jiahua Dong, Duzhen Zhang, Salman Khan and Fahad Shahbaz Khan are with the Mohamed bin Zayed University of Artificial Intelligence, Abu Dhabi, UAE. Email: dongjiahua1995@gmail.com, \{duzhen.zhang, salman.khan, fahad.khan\}@mbzuai.ac.ae. \protect\\
\vspace{-8pt}
\IEEEcompsocthanksitem Wenqi Liang is with the University of Trento, Trento, Italy. Email: liangwenqi0123@gmail.com. \protect\\
\vspace{-8pt}
\IEEEcompsocthanksitem Hongliu Li is with the Department of Civil and Environmental Engineering, The Hong Kong Polytechnic University, Hong Kong, China. Email: hongliuli1994@gmail.com. \protect\\
\vspace{-8pt}
\IEEEcompsocthanksitem Yang Cong is with the South China University of Technology, Guangzhou, China. Email: congyang81@gmail.com. \protect\\
\vspace{-8pt}
\IEEEcompsocthanksitem Hanbin Zhao is with the College of Computer Science and Technology, Zhejiang University, Hangzhou, China.
Email: zhaohanbin@zju.edu.cn.  \protect\\
\vspace{-8pt}
\IEEEcompsocthanksitem Henghui Ding is with the Institute of Big Data, Fudan University, Shanghai, China. Email: henghui.ding@gmail.com.
\protect\\
\vspace{-8pt}
\IEEEcompsocthanksitem Yulun Zhang is with the Shanghai Jiao Tong University, Shanghai, China. Email: yulun100@gmail.com. \protect\\
}

\thanks{Manuscript received April 19, 2005; revised August 26, 2015.}
\thanks{This work was supported by the National Natural Science Foundation of China under Grants 62501386 and 62472104, the CCF-Tencent Rhino-Bird Open Research Fund, and the CAAI-Tencent Rhino-Bird Open Research Fund. This work was also supported by the AI Hundred Schools Program and carried out using the Ascend AI technology stack. }
\thanks{Jiahua Dong and Wenqi Liang contributed equally. }
\thanks{The corresponding author is Dr. Hongliu Li.}
}

\IEEEtitleabstractindextext{
\begin{abstract}
Custom diffusion models (CDMs) have garnered significant interest owing to their remarkable capacity for generating personalized concepts. However, the majority of CDMs unrealistically presume that the user's collection of personalized concepts is static and incapable of incremental growth over time. Furthermore, they exhibit significant catastrophic forgetting and concept neglect of previously learned concepts when incrementally learning a sequence of new ones. To resolve the above challenges, we develop a novel \underline{C}ontinually \underline{C}ustomizable \underline{D}iffusion \underline{M}odel (CCDM), enabling users to perform concept-incremental versatile customization. Specifically, we design an attribute-decoupled LoRA (AD-LoRA) module and a relevance-guided AD-LoRA aggregation strategy
to mitigate catastrophic forgetting. They can preserve concept-specific attributes of each task and leverage beneficial inter-task correlations to enhance the continual learning of new customization tasks. Additionally, to address the challenge of concept neglect, we propose a controllable regional context synthesis strategy that performs multi-concept composition in alignment with user-provided conditions. This strategy enhances the overall consistency in multi-concept synthesis by guaranteeing semantic independence between user-defined regions and their smooth boundary transitions. 
Experiments show our CCDM exhibits significant improvements over baseline methods. 

\end{abstract}

}

\maketitle

\IEEEdisplaynontitleabstractindextext
\IEEEpeerreviewmaketitle

\IEEEraisesectionheading{\section{Introduction}
\label{sec: introduction}}
\IEEEPARstart{L}{atent} diffusion models (LDMs) \cite{podell2024sdxl, xie2024sana, Hu_2025_CVPR} have emerged as an effective approach for high-quality content generation across various domains, including text-to-image synthesis \cite{10377881}, video generation \cite{Wu_2025_CVPR}, and 3D object creation \cite{Yang_2025_CVPR}. By encoding input data into a latent space using a pretrained autoencoder, LDMs perform the denoising-based diffusion process in a lower-dimensional domain \cite{mou2023t2i}. This design significantly reduces computational and memory costs while preserving the semantic richness required for photorealistic generation \cite{karras2022elucidating, Henschel_2025_CVPR}. Despite their impressive performance, standard LDMs are typically trained on large-scale, general-purpose datasets \cite{Meng_2025_CVPR, wang2023lavie} and struggle to synthesize user-specific concepts without extensive retraining. To address this limitation, custom diffusion models (CDMs) \cite{yang2024loracomposer, hu2024storyagentcus, Ma_2025_CVPR, he2025cameractrl} have become a promising paradigm by using low-rank adaptation (LoRA) \cite{hu2022lora} to finetune LDMs \cite{Meng_2025_CVPR, Blattmann_2023_CVPR} on users' personalized concepts. They expand the visual-language vocabulary of pretrained LDMs \cite{10377881} by associating user-defined concepts with desired subjects \cite{Qin_2025_CVPR, li2024motrans}.

\begin{figure}[t]
\centering
\includegraphics[width = 1.0\linewidth]
{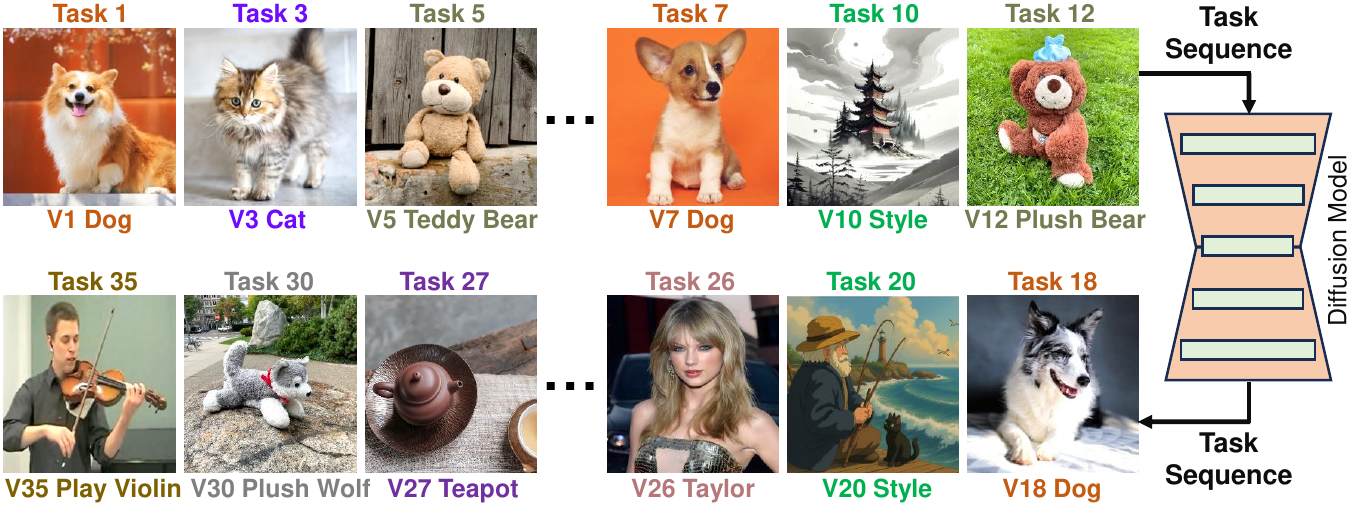}
\vspace{-23pt}
\caption{Illustration of the proposed CIVC problem. }
\label{fig: teaser_CIVC}
\end{figure}

\begin{figure*}[t]
\centering
\includegraphics[width = 1.0\linewidth]
{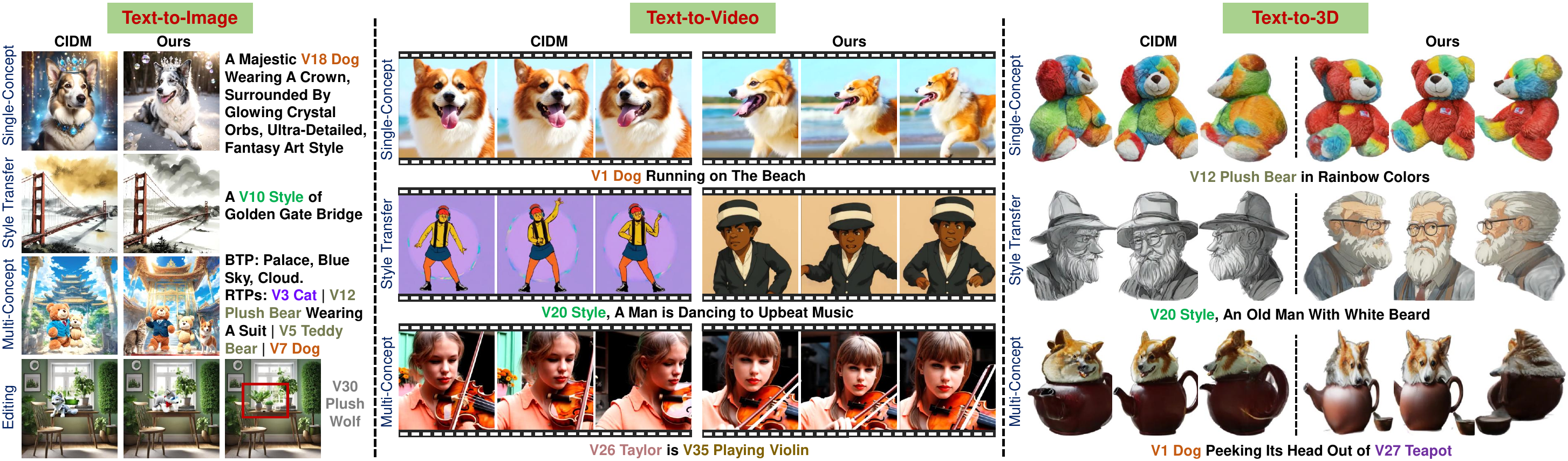}
\vspace{-23pt}
\caption{Demonstration of our model's scalability in supporting versatile concept customization tasks, including single/multi-concept synthesis, editing, and style transfer across text-to-image, text-to-video, and text-to-3D domains under the CIVC problem.   } 
\label{fig: teaser_exp}
\end{figure*}

Typically, the majority of CDMs \cite{ruiz2023dreambooth, wu2024customcrafter, Petrov_2025_CVPR} are built on the assumption that the set of each user’s personalized concepts remains constant and cannot continuously expand over time. However, this assumption contradicts practical scenarios, where users require a growing collection of new personalized concepts and wish to continually generate new concepts to reflect their evolving life experiences. To address such scenarios, existing CDMs \cite{chen2024disenstudio, 101007978031_72907_23, Lu_2024_CVPR, Huang_2025_CVPR} often rely on retaining all training data of previously learned concepts and leverage LoRA \cite{hu2022lora} to finetune pretrained LDMs \cite{Lin_2023_CVPR, Hu_2025_CVPR, Wu_2025_CVPR}. Yet, as the volume of learned concepts accumulates, this strategy renders CDMs increasingly computationally intensive and more vulnerable to serious privacy issues \cite{le_etal2023antidreambooth}. If existing CDMs store all previously acquired low-rank parameters associated with old concepts and linearly fuse them for the continual learning of new concepts \cite{wu2024mixture, zhong2024multi}, they may suffer from substantial degradation in their ability to preserve the distinctive characteristics of old personalized concepts, thereby hindering versatile concept customization \cite{smith2023continual}. This phenomenon is widely known as catastrophic forgetting of previously learned concepts \cite{Guo_2025_CVPR}. 
Furthermore, in practical applications, users often seek precise control over the subjects and contexts linked to multiple previously learned concepts within generated images, videos, or 3D scenes for multi-concept composition, using conditions they provide (such as bounding boxes and scribbles \cite{gu2023mixofshow}). This requirement poses a significant challenge to CDMs \cite{yang2024loracomposer}, which frequently face the issue of concept neglect \cite{11453592116_Chefer}, whereby certain old concepts fail to appear during the synthesis of multiple concepts.

\begin{table}[t]
\centering
\footnotesize
\setlength{\tabcolsep}{1.3mm}
\renewcommand{\arraystretch}{1.04}
\caption{Differences between our model and competing methods, where CL, SC, MC, ST, and ET denote continual learning, single-concept, multi-concept, style transfer, and editing, respectively. } 
\vspace{-3mm}
\resizebox{\linewidth}{!}{
\begin{tabular}{l|c|cccc|ccc|ccc}
\toprule
\makecell[c]{\multirow{2}{*}{Baseline Methods}} & \multirow{2}{*}{CL} & \multicolumn{4}{c|}{Text-to-Image} &\multicolumn{3}{c|}{Text-to-Video} & \multicolumn{3}{c}{Text-to-3D} \\
 &  & SC & MC & ST & ET & SC & MC & ST & SC & MC & ST \\
\midrule
CIDM \cite{NEURIPS2024_DongJH} (NeurIPS'2024) & \textcolor{orange}{\cmark} & \textcolor{deepred}{\cmark}&\textcolor{deepred}{\cmark} &\textcolor{deepred}{\cmark} &\textcolor{deepred}{\cmark} & & & & & \\
MuseumMaker \cite{10965859} (TIP'2025) & \textcolor{orange}{\cmark} & & & \textcolor{deepred}{\cmark} & & & & & & \\
CPG \cite{Guo_2025_CVPR} (CVPR'2025) & \textcolor{orange}{\cmark} &\textcolor{deepred}{\cmark} &\textcolor{deepred}{\cmark} & & & & & & &  \\ 
Stylemaster \cite{ye2025stylemaster} (CVPR'2025) &  & & & & & & & \textcolor{deepgreen}{\cmark} &  &  \\
DualReal \cite{wang2025dualreal} (ICCV'2025) &  & & & & & \textcolor{deepgreen}{\cmark} & \textcolor{deepgreen}{\cmark} & & &   \\
Dream3D \cite{Raj_2023_ICCV} (ICCV'2023) &  & & & & & & & & \textcolor{deepblue}{\cmark} &\textcolor{deepblue}{\cmark} &\textcolor{deepblue}{\cmark}  \\
TV-3DG \cite{11081388} (TPAMI'2025) &  & & & & & & & & \textcolor{deepblue}{\cmark} &\textcolor{deepblue}{\cmark} &\textcolor{deepblue}{\cmark} \\
\midrule

\textbf{Ours} (\textbf{CCDM}) & \textcolor{orange}{\cmark} &\textcolor{deepred}{\cmark} &\textcolor{deepred}{\cmark} & \textcolor{deepred}{\cmark} &\textcolor{deepred}{\cmark} &\textcolor{deepgreen}{\cmark} &\textcolor{deepgreen}{\cmark} & \textcolor{deepgreen}{\cmark} &\textcolor{deepblue}{\cmark} &\textcolor{deepblue}{\cmark} &\textcolor{deepblue}{\cmark}  \\
\bottomrule
\end{tabular}
}
\label{tab: difference_comparison}
\vspace{-3mm}
\end{table}

To address the practical scenarios mentioned above, this work introduces a novel problem termed \underline{C}oncept-\underline{I}ncremental \underline{V}ersatile \underline{C}ustomization (CIVC). As illustrated in Fig.~\ref{fig: teaser_CIVC}, CIVC enables CDMs \cite{Qin_2025_CVPR}  to continually learn and generate a series of new personalized concepts in a concept-incremental manner, while also supporting versatile concept customization tasks, including single/multi-concept synthesis, editing and style transfer across text-to-image \cite{Zhu_Li_Ma_He_Li_2025}, text-to-video \cite{Huang_2025_CVPR} and text-to-3D \cite{Ma_2025_CVPR} domains. 
Our work significantly differs from existing baselines and offers strong scalability to perform a variety of different concept customization tasks (see Tab.~\ref{tab: difference_comparison}).
Additionally, during multi-concept composition, it allows users to guide the generation process by specifying the desired contexts, objects, and their intended spatial locations. Overall, the CIVC problem proposed in this paper faces the following two challenges: 
\begin{itemize}
\item \textbf{Catastrophic Forgetting} refers to the substantial loss of unique attributes of old concepts when existing CDMs \cite{ruiz2023dreambooth, Petrov_2025_CVPR, wu2024mixture} linearly merge their low-rank weights to incrementally learn new concepts. If new concepts (\emph{e.g.}, golden retriever) are semantically similar to previously learned ones (\emph{e.g.}, labrador retriever), CDMs tend to exhibit more severe forgetting of those old concepts. 

\item \textbf{Concept Neglect} indicates the failure of some intended concepts to be properly synthesized at user-specified spatial locations during multi-concept composition. As multiple personalized concepts are composed together within generated images, videos, or 3D scenes, the risk increases that certain concepts may be overlooked, incorrectly blended, or visually overshadowed by others. 
\end{itemize}

To address the challenges of catastrophic forgetting and concept neglect in CIVC, we propose a novel \underline{C}ontinually \underline{C}ustomizable \underline{D}iffusion \underline{M}odel (CCDM) in this work. 
\textbf{Firstly}, we develop an attribute-decoupled LoRA (AD-LoRA) module and a relevance-guided AD-LoRA aggregation strategy to alleviate catastrophic forgetting of previous concepts. The AD-LoRA module is designed to automatically retain concept-specific attributes while filtering out redundant characteristics. This is achieved by masking the LoRA parameters corresponding to irrelevant properties. By assessing semantic relevance, the relevance-guided AD-LoRA aggregation strategy dynamically fuses low-rank updates from both current and previous tasks, exploiting beneficial inter-task correlations to facilitate the continual learning of new customization tasks. 
\textbf{Secondly}, to address concept neglect during multi-concept composition, we design a controllable regional context synthesis strategy that controls generated contexts to meet user-specified conditions. This module leverages a unified attention mask to restrict self-attention among image features exclusively to user-specified concept regions, while allowing cross-attention from image features to the textual embeddings of those regions. It promotes semantic independence across regions and seamless transitions at their boundaries, enhancing the holistic consistency of multi-concept synthesis. 
Comprehensive evaluations verify the efficacy of our CCDM in addressing diverse CIVC tasks across text-to-image, text-to-video, and text-to-3D domains (see Fig.~\ref{fig: teaser_exp}). Our core contributions are listed as follows: 
\begin{itemize}
\item We formulate a practical Concept-Incremental Versatile Customization (CIVC) problem, characterized by challenges like catastrophic forgetting and concept neglect. 

\item We design a novel Continually Customizable Diffusion Model (CCDM) to tackle the challenges of CIVC, enabling the continual customization of new concepts. 

\item We propose attribute-decoupled LoRA (AD-LoRA) and relevance-guided AD-LoRA aggregation to alleviate catastrophic forgetting by retaining concept-specific attributes and exploiting positive inter-task correlations. 

\item We develop a controllable regional context synthesis strategy to mitigate concept neglect in multi-concept generation by enabling semantic independence across regions and smooth transitions at their boundaries. 
\end{itemize}

Building upon our prior research \cite{NEURIPS2024_DongJH}, this paper presents substantial improvements in both methodology and experiments: 1) Compared with \cite{NEURIPS2024_DongJH}, we propose a novel attribute-decoupled LoRA (AD-LoRA) module to identify which attributes of personalized concepts are distinctive. This AD-LoRA module preserves only the unique concept attributes to mitigate catastrophic forgetting. 
2) The relevance-guided AD-LoRA aggregation merges the low-rank updates from current and previous tasks based on their semantic relevance. It explores positive inter-task relations to facilitate continual learning of new tasks. 
3) We devise a controllable regional context synthesis strategy that uses a unified attention mask to limit self-attention to user-defined regions while enabling cross-attention to their textual embeddings for multi-concept synthesis. In contrast to \cite{NEURIPS2024_DongJH}, it promotes greater semantic independence across regions and smoother transitions at their boundaries to tackle concept neglect.
4) We extend \cite{NEURIPS2024_DongJH} to diverse CIVC tasks spanning text-to-image, text-to-video, and text-to-3D domains.  
5) Our CCDM reduces the trainable parameters of \cite{NEURIPS2024_DongJH} by 35\% but achieves better customization performance than \cite{NEURIPS2024_DongJH}. 
6) More comparative evaluations, ablation results, and in-depth analyses are presented to prove the efficacy of our CCDM in handling the CIVC problem.

\section{Related Work}
\textbf{Continual Learning} (CL) \cite{Wu_2025_CVPR_Elegant_GIFT, Douillard_2022_CVPR, 101007978_3_03119809_0_36} allows AI systems to continuously accumulate new knowledge from a stream of tasks while preserving previously learned information \cite{900901934533}. Modern regularization approaches impose optimization constraints on the weight updates of previous tasks to maintain previously learned knowledge \cite{kirkpatrick2017overcoming, li2017learning} or apply knowledge distillation to facilitate knowledge propagation between new and old models \cite{Dong_2022_CVPR}. Additionally, data replay methods maintain exemplars from previous categories \cite{900901934533, 10323204}, or leverage generative architectures to synthesize data representing old classes \cite{Wu_2025_CVPR_Elegant_GIFT}. These methods retrain AI systems from scratch using a combination of new-class data and synthesized data from old classes.
Dynamic architectural approaches progressively adapt their structures to continually assimilate new classes \cite{yoon2018lifelong, Douillard_2022_CVPR}, while representation-based methods utilize prompt learning \cite{101007978_3_03119809_0_36} or contrastive learning \cite{madaan2022representational} to alleviate forgetting. Moreover, CL has been applied to a variety of vision tasks, including object recognition \cite{yoon2018lifelong}, scene understanding \cite{Truong_2025_CVPR}, and multi-modal reasoning \cite{liu2025cclip}. Despite their effectiveness in visual domains, most CL methods \cite{Truong_2025_CVPR, Douillard_2022_CVPR, 10323204} cannot be directly applied to solve the continual concept customization task (\emph{e.g.}, the CIVC problem).

\textbf{Concept Customization} \cite{Zhou_2024_CVPR, chen2024disenstudio, Blattmann_2023_CVPR, Lu_2024_CVPR, 10656166} refers to the process of extending pretrained LDMs \cite{rombach2022high, Hu_2025_CVPR} to generate user-defined personalized concepts. Following \cite{ruiz2023dreambooth}, which addresses subject-driven synthesis by finetuning the full set of parameters in LDMs \cite{Blattmann_2023_CVPR}, Gal \emph{et al.} leverage textual inversion \cite{gal2023an} to encode personalized concepts into learnable word embeddings for text-to-image concept customization. \cite{kumari2022customdiffusion} enables the simultaneous training of multiple concepts or the merging of distinct diffusion models through the optimization of cross-attention parameters for multi-concept composition. Inspired by \cite{kumari2022customdiffusion}, \cite{10377873} optimizes only the singular value decomposition of latent encoding parameters to enhance customization efficiency. Building upon prior works \cite{liu2023customizable, gal2023an}, some researchers \cite{10656166, wu2024customcrafter, ye2025stylemaster, li2024motrans, wang2025dualreal} have extended these methodologies to temporal domains for personalized text-to-video synthesis. DreamVideo \cite{10656166} proposes a dual-phase training strategy that separates concept acquisition from motion modeling, and \cite{chen2024disenstudio} is designed for multi-concept video customization. For text-to-3D generation \cite{10114536646473681391, 101007978031_72907_23}, several works \cite{Raj_2023_ICCV, poole2023dreamfusion, Qin_2025_CVPR, Yang_2025_CVPR} use 2D LDMs \cite{saharia2022photorealistic, Gu_2022_CVPR} as the backbone and apply pose control to 3D objects, while others \cite{Petrov_2025_CVPR, Ma_2025_CVPR, 11081388} aim to incorporate implicit 3D priors to facilitate the generation process. Besides, \cite{smith2023continual, sun2024create, Guo_2025_CVPR} focus on continual customization tasks that continuously synthesize new concepts, yet they lack the ability to control generated contexts and are prone to concept neglect during multi-concept synthesis. To tackle this issue, \cite{10377881, mou2023t2i} use regional conditions, such as bounding boxes, to guide composition. However, the above CDMs \cite{he2025cameractrl, kumari2022customdiffusion, 10114536646473681391} fail to incrementally synthesize new concepts in CIVC scenarios, suffering from catastrophic forgetting and concept neglect.

\section{Preliminary and Problem Definition}
\textbf{Preliminary:}
To steer the generation of images, latent diffusion models (LDMs) \cite{saharia2022photorealistic, xie2024sana, Gu_2022_CVPR, rombach2022high} utilize conditional cues such as textual descriptions or images. The diffusion process takes places in a latent space, where an encoder $\mathcal{E}(\cdot)$ transforms inputs into latent representations, and a decoder $\mathcal{D}(\cdot)$ reconstructs them into images. By finetuning pretrained LDMs \cite{10114536805283687625, podell2024sdxl} with low-rank adaptation (LoRA) \cite{hu2022lora, 1555536920703693369}, custom diffusion models (CDMs) \cite{liu2023customizable, Lu_2024_CVPR, Zhou_2024_CVPR} can learn user-specific concepts. For a personalized image $\mathbf{x}$ paired with its corresponding text prompt $\mathbf{p}$, the encoder $\mathcal{E}(\cdot)$ projects $\mathbf{x}$ into a latent feature $\mathbf{z}$. The noisy version of $\mathbf{z}$ at the $t$-th ($t = 1, \dots, T$) timestep is denoted as $\mathbf{z}_t$.
The text prompt $\mathbf{p}$ is then encoded into the textual embedding $\mathbf{c} = \Psi(\mathbf{p})$ using a pretrained text encoder $\Psi(\cdot)$ (such as CLIP \cite{Ramesh2022HierarchicalTI}). 
As a result, at the $t$-th timestep, the learning objective for the customized concept pair $\{\mathbf{x}, \mathbf{p}\}$ is defined as follows: 
\begin{align}
\mathcal{L}_{\mr{CDMs}} = \mathbb{E}_{\mathbf{z}\sim\mathcal{E}(\mathbf{x}), \mathbf{c}, \epsilon\sim \mathcal{N}(0, \mathbf{I}), t}[\|\epsilon - \epsilon_{\theta^\prime} (\mathbf{z}_t| \mathbf{c}, t)\|_2^2],
\label{eq: CDM_loss}
\end{align}
where the denoising UNet $\epsilon_{\theta'}(\cdot)$ progressively denoises $\mathbf{z}_t$ by estimating $\epsilon_{\theta'}(\mathbf{z}_t|\mathbf{c}, t)$ \cite{ruiz2023dreambooth, Zeng_2024_CVPR}, which is assumed to follow a Gaussian distribution $\epsilon\sim \mathcal{N}(0, \mathbf{I})$. Additionally, the parameters $\theta^\prime = \theta_0 + \Delta\theta$ are composed of two components: the pretrained weights $\theta_0=\{\mathbf{W}_0^l\}_{l=1}^L$ from LDMs \cite{10555536920703692573, wang2025msdiffusion} and the low-rank updates $\Delta\theta=\{\Delta\mathbf{W}^l\}_{l=1}^L$ introduced by LoRA \cite{hu2022lora}. In the $l$-th ($l=1, \cdots, L$) transformer layer of $\epsilon_{\theta'}(\cdot)$, $\mathbf{W}_0^l, \Delta\mathbf{W}^l \in\mathbb{R}^{a\times b}$ represent the pretrained parameters and low-rank weights. $a$ and $b$ indicate the number of rows and columns. $\Delta\mathbf{W}^l = \mathbf{A}^l\mathbf{B}^l$ can be factorized as the product of two low-rank matrices \cite{ruiz2023dreambooth}: $\mathbf{A}^l\in\mathbb{R}^{a\times r}$ and $\mathbf{B}^l\in\mathbb{R}^{r\times b}$, with the rank $r$ satisfying the constraint $r\ll\min(a,b)$.

Nevertheless, existing CDMs \cite{11081388, gal2023an} typically presume a fixed set of user-specific concepts, which does not align with practical scenarios where users often desire to consecutively incorporate new personalized concepts reflecting their evolving preferences \cite{smith2023continual}. 
Additionally, these models are notably vulnerable to catastrophic forgetting of previously learned concepts and the neglect of certain concepts \cite{sun2024create, liu2023customizable} when adapting to an incrementally expanding set of personalized concepts for versatile customization \cite{NEURIPS2024_DongJH}.

\textbf{Problem Definition:} 
To tackle the aforementioned issues, this paper introduces an innovative problem termed Concept-Incremental Versatile Customization (CIVC). This problem involves a sequence of successive concept customization tasks, denoted as $\mathcal{T}=\{\mathcal{T}_u\}_{u=1}^U$, with $U$ indicating the total number of tasks. The $u$-th task $\mathcal{T}_u = \{\mathbf{x}_u^s, \mathbf{p}_u^s, \mathbf{y}_u^s\}_{s=1}^{N_u}$ comprises $N_u$ triplet samples, each containing an image $\mathbf{x}_u^s$, a corresponding text prompt $\mathbf{p}_u^s$, and a concept token $\mathbf{y}_u^s \in\mathcal{Y}_u$. 
Specifically, the text prompt $\mathbf{p}_u^s$ describes the visual content of $\mathbf{x}_u^s$ (\emph{e.g.}, ``photo of a [$V_*$] [$V_{\mathrm{cat}}$] drinking a cup of beer''), while $\mathbf{y}_u^s$ is the conceptual element (such as [$V_*$] [$V_{\mathrm{cat}}$]) extracted from $\mathbf{p}_u^s$. The concept space $\mathcal{Y}_u$ for the $u$-th learning task contains $C_u$ new concepts, defined as $\mathbf{y}_u = \cup_{s=1}^{N_u} \mathbf{y}_u^s$. In the CIVC setting, the concept spaces of distinct tasks are completely disjoint: $\mathcal{Y}_u\cap(\cup_{i=1}^{u{-}1}\mathcal{Y}_i)=\emptyset$. This indicates that the $C_u$ new personalized concepts introduced in the $u$-th task are entirely distinct from the $K_{u{-}1} = \sum_{i=1}^{u{-}1} C_i$ previous concepts accumulated from the preceding $u{-}1$ tasks.  
As shown in Tab.~\ref{fig: teaser_CIVC}, based on individual user preferences, each task $\mathcal{T}_u$ belongs to one of the diverse customization tasks: single/multi-concept synthesis, editing and style transfer across text-to-image \cite{10377873}, text-to-video \cite{wu2024motionbooth} and text-to-3D \cite{song2025mult} domains. This is also referred to as versatile customization, which is the main focus of this paper. 
To satisfy the practical requirements of CIVC, no memory is reserved for archiving or reconstructing the training sets $\{\mathcal{T}_u\}_{u=1}^U$ across all tasks. This design ensures that all customization tasks adhere strictly to the concept-incremental learning paradigm. 
The CIVC setting aims to facilitate continual learning of new personalized concepts to meet diverse customization needs, while effectively tackling catastrophic forgetting of previous concepts and the challenge of concept neglect.

\section{The Proposed Model}
\begin{figure}[t]
\centering
\includegraphics[width = 1.0\linewidth]
{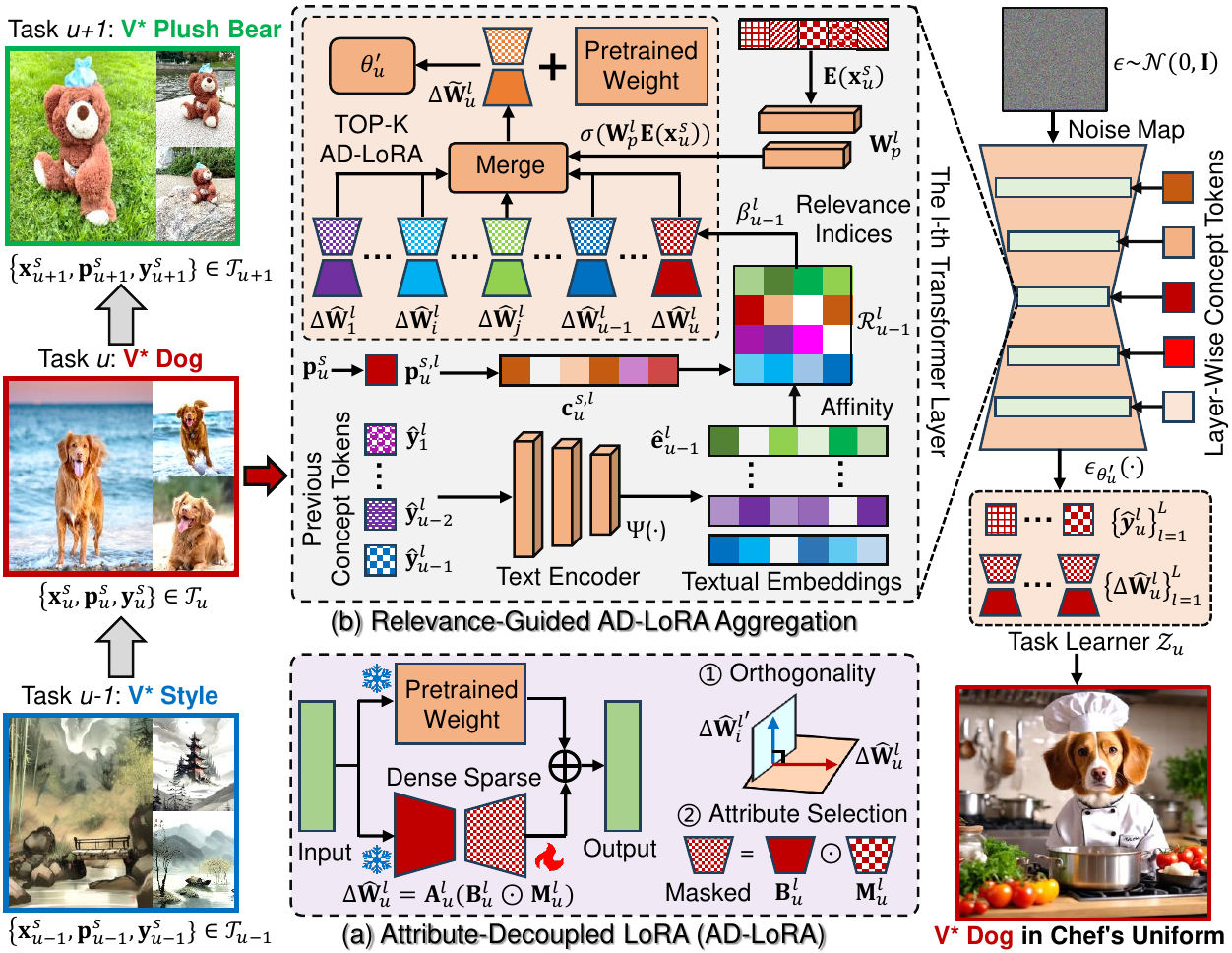}
\vspace{-8mm}
\caption{Demonstration of (a) the attribute-decoupled LoRA (AD-LoRA) module and (b) the relevance-guided AD-LoRA aggregation strategy, using text-to-image customization as an example to illustrate the pipeline of our CCDM for mitigating catastrophic forgetting in the CIVC setting. } 
\label{fig: model_pipeline_ab}
\vspace{-1mm}
\end{figure}

In this section, as shown in Figs.~\ref{fig: model_pipeline_ab}-\ref{fig: model_pipeline_c}, we first use text-to-image customization as a representative example to illustrate the pipeline of our CCDM, which resolves the challenges of catastrophic forgetting and concept neglect in the CIVC problem. Specifically, we propose an attribute-decoupled LoRA (AD-LoRA) module (Sec.~\ref{sec: AD_LoRA}) and a relevance-guided AD-LoRA aggregation strategy (Sec.~\ref{sec: AD_LoRA_Aggregation}) to overcome catastrophic forgetting, while also designing a controllable regional context synthesis strategy (Sec.~\ref{sec: regional_context_synthesis}) to alleviate concept neglect during multi-concept generation. Then, we discuss how to extend our CCDM to text-to-video (Sec.~\ref{sec: extension_text_to_video}) and text-to-3D (Sec.~\ref{sec: extension_text_to_3D}) customization under the CIVC setting.

\subsection{Attribute-Decoupled LoRA (AD-LoRA)}\label{sec: AD_LoRA}
To enable the pretrained denoising UNet $\epsilon_{\theta_0}(\cdot)$ to learn the $u$-th concept customization task $\mathcal{T}_u$, CDMs \cite{Ruiz_2024_CVPR, Nam_2024_CVPR, DBLPconfaaaiWuPZPYZLM25, podell2024sdxl, Zhu_Li_Ma_He_Li_2025} employ LoRA \cite{hu2022lora} to finetune the model on personalized data $\mathcal{T}_u = \{\mathbf{x}_u^s, \mathbf{p}_u^s, \mathbf{y}_u^s\}_{s=1}^{N_u}$. This process involves optimizing the objective in Eq.~\eqref{eq: CDM_loss}, resulting in an updated denoising UNet $\epsilon_{\theta^\prime_u}(\cdot)$. 
Here, $\theta^\prime_u = \theta_0 + \Delta\theta_u$, $\Delta\theta_u = \{\Delta\mathbf{W}_u^l\}_{l=1}^L$, and $\Delta\mathbf{W}_u^l= \mathbf{A}_u^l\mathbf{B}_u^l \in\mathbb{R}^{a\times b}$ represents the low-rank updates in the $l$-th layer, with $\mathbf{A}_u^l \in\mathbb{R}^{a\times r}$ and $\mathbf{B}_u^l \in\mathbb{R}^{r\times b}$ being the corresponding low-rank decomposition matrices. Previous studies \cite{rombach2022high, yang2024loracomposer} have demonstrated that $\Delta\theta_u$ effectively captures the majority of the personalized concept characteristics in $\mathcal{T}_u$. To continually learn new customization tasks under the CIVC setting, existing CDMs \cite{articleJinUniCanvas, Wang_Bai_Xie_Yi_Wang_Ma_2025, Raj_2023_ICCV} require retaining the low-rank weights $\{\Delta\theta_i\}_{i=1}^{u-1}$ from all preceding tasks and linearly combining them during training by assessing their respective contributions to the current task. Unfortunately, this linear combination of low-rank weights causes a significant degradation in their ability to capture the distinctive attributes of previous concepts, a phenomenon also known as catastrophic forgetting of old concepts \cite{Guo_2025_CVPR}.

In light of this, Dong \emph{et al.} \cite{NEURIPS2024_DongJH} aim to mitigate catastrophic forgetting during model training by simultaneously preserving task-specific and shared knowledge through a concept consolidation loss. However, this method fails to identify which attributes of personalized concepts are distinctive and which are irrelevant in each incremental task. Moreover, the orthogonal constraint in \cite{NEURIPS2024_DongJH} is highly dependent on hyperparameter selection, which cannot guarantee the absence of negative interference from irrelevant attributes across different tasks.
Under these circumstances, mistakenly extracting irrelevant attributes of personalized concepts from previous tasks can severely interfere with the discovery of inherent attributes in the current customization task, thereby exacerbating forgetting under the CIVC setting. 
To address the above issues, as shown in Fig.~\ref{fig: model_pipeline_ab}, we propose layer-wise concept tokens to explore a wide range of diverse attributes for personalized concepts. Additionally, we develop an attribute-decoupled LoRA (AD-LoRA) module that selectively preserves task-specific distinctive attributes while discarding irrelevant properties by zeroing out their encoding parameters in the low-rank weights. During the process of capturing concept-unique attributes, we strive to maximize the orthogonality of AD-LoRA across different tasks, thereby minimizing the negative interference of irrelevant attributes from previous concepts on learning the current task.

\textbullet\textbf{Layer-Wise Concept Tokens: }
Unlike traditional textual inversion approaches \cite{gal2023an, Zhang_2023_inst, Choi_Park_Baek_2025} that apply an identical text prompt across all transformer layers of the pretrained UNet $\epsilon_{\theta_0}(\cdot)$, we propose layer-wise concept tokens to capture diverse properties of personalized concepts during each concept customization task. Specifically, within the $u$-th task $\mathcal{T}_u$, each input tuple $\{\mathbf{x}_u^s, \mathbf{p}_u^s, \mathbf{y}_u^s\}$ contains $L$ layer-wise text prompts $\{\mathbf{p}_u^{s,l}\}_{l=1}^L$. At the $l$-th layer of $\epsilon_{\theta_u^\prime}(\cdot)$, the prompt $\mathbf{p}_u^{s,l}$ is paired with a corresponding layer-wise concept token $\mathbf{y}_u^{s,l}$. To provide a concrete instance, when processing an image $\mathbf{x}_u^s$ with its associated text prompt $\mathbf{p}_u^s$ (``photo of a [$V_*$] [$V_{\mathrm{cat}}$] drinking a cup of beer''), the text prompt for the $l$-th layer is expressed as $\mathbf{p}_u^{s,l}$ (``photo of a [$V_*^l$] [$V_{\mathrm{cat}}^l$] drinking a cup of beer''), with the corresponding layer-wise concept token $\mathbf{y}_u^{s,l}$ being [$V_*^l$] [$V_{\mathrm{cat}}^l$]. 
After $\mathbf{p}_u^s$ is employed to initialize $\{\mathbf{p}_u^{s, l}\}_{l=1}^L$, the textual embedding $\mathbf{c}_u^{s, l} = \Psi(\mathbf{p}_u^{s, l} ) \in\mathbb{R}^{N_p\times d}$, obtained by encoding $\mathbf{p}_u^{s, l}$ with the text encoder $\Psi(\cdot)$, is injected into the $l$-th transformer layer of $\epsilon_{\theta_u^\prime}(\cdot)$ via cross-attention. Here, $N_p$ is the token length and $d$ refers to the feature dimension of the textual embedding.
When incorporating LoRA and layer-wise concept tokens to optimize $\epsilon_{\theta^\prime_u}(\cdot)$ via Eq.~\eqref{eq: CDM_loss}, we can obtain varied attributes of personalized concepts within the $u$-th concept customization task. Nevertheless, it is challenging to manually determine which attributes of personalized concepts are distinctive and which are redundant in each customization task. Such irrelevant concept properties may hinder the continual learning of new tasks under the CIVC setting, thereby aggravating catastrophic forgetting of previously learned concepts.

\textbullet\textbf{AD-LoRA Module:}
To preserve concept-unique attributes while discarding irrelevant ones, we propose an attribute-decoupled LoRA (AD-LoRA) module. 
For the low-rank update $\Delta\mathbf{W}_u^l = \mathbf{A}_u^l\mathbf{B}_u^l$ of the $l$-th layer, we initialize $\mathbf{A}_u^l$ by sampling from a standard Gaussian distribution $\mathcal{N}(0, 1)$ and set $\mathbf{B}_u^l$ to zero, ensuring that $\Delta\mathbf{W}_u^l = 0$ at the beginning of learning the $u$-th task. Subsequently, we keep $\mathbf{A}_u^l$ frozen and update only the key parameters of $\mathbf{B}_u^l$ that are responsible for capturing the distinctive properties of personalized concepts, while zeroing out the encoding parameters of irrelevant attributes to facilitate the learning of the $u$-th task. To achieve this, we employ parameter importance calibration to derive a sparse mask $\mathbf{M}_u^l \in \mathbb{R}^{r \times b}$ for $\mathbf{B}_u^l$. This mask ensures that training updates are applied only to the parameters in $\mathbf{B}_u^l$ that are essential for learning the $u$-th task. For parameter importance calibration, we first finetune $\{\mathbf{B}_u^l\}_{l=1}^L$ on the $u$-th task $\mathcal{T}_u$, and then use the first-order gradient updates of $\mathbf{B}_u^l$ at the $m$-th iteration ($m=200$) to compute its Fisher information matrix $\mf{F}_{u}^l \in\mathbb{R}^{r \times b}$. It is worth noting that $\mf{F}_{u}^l$ can evaluate the importance of each parameter in learning the $u$-th task, where higher values correspond to greater importance. Subsequently, we arrange the absolute values of all elements from $\{\mathbf{F}_u^l\}_{l=1}^L$ in descending order to obtain $\mathbf{F}_u^* \in\mathbb{R}^{N_b}$, and use the sparsity ratio $\rho\in(0,1)$ to select the element ranked at the $\rho N_b$-th position in $\mathbf{F}_u^*$ as the global threshold $\tau_u$. Here, $N_b = rbL$ is relatively small since $r \ll b$. 
As a result, at the $u$-th task, the sparse mask $\mathbf{M}_u^l \in \mathbb{R}^{r \times b}$ for the $l$-th layer is defined as: 
\begin{align}
\mathbf{M}_u^l = \mathbb{I}_{|\mathbf{F}_u^l| \geq\tau_u},~~\tau_u = \mathbf{F}_u^*[\rho N_b],
\label{eq: definition_mask}
\end{align}
where $\mathbb{I}(\cdot)$ is the indicator function that sets the corresponding values in $\mathbf{M}_u^l$ to 1 if the absolute values of elements in $\mathbf{F}_u^l$ exceed $\tau_u$. Then, we apply $\mathbf{M}_u^l$ to $\mathbf{B}_u^l$ to derive a new low-rank update $\Delta\widehat{\mathbf{W}}_u^l$ for the $l$-th layer at the $u$-th task: 
\begin{align}
\Delta\widehat{\mathbf{W}}_u^l = \mathbf{A}_u^l(\mathbf{B}_u^l\odot \mathbf{M}_u^l),
\label{eq: new_low_rank_update}
\end{align}
where $\odot$ is the Hadamard product. Evidently, if the absolute values in $\mathbf{F}_u^l$ exceed $\tau_u$, the corresponding parameters in $\mathbf{B}_u^l$ are considered essential for capturing distinctive properties of the personalized concepts in the $u$-th task and are retained by the mask $\mathbf{M}_u^l$. Otherwise, they are regarded as encoding irrelevant attributes and are masked out (\emph{i.e.}, set to zero).

\textbf{Orthogonality of AD-LoRA:}
The matrices $\{\mathbf{A}_u^l\}_{l=1}^L$ in Eq.~\eqref{eq: new_low_rank_update}, initialized from a Gaussian distribution $\mathcal{N}(0, 1)$ and kept fixed during training, ensure that the low-rank updates across different layers are orthogonal for the $u$-th customization task: $\langle \Delta\widehat{\mathbf{W}}_u^l, \Delta\widehat{\mathbf{W}}_u^{\gamma} \rangle \approx 0$, where $\gamma = 1, \cdots, L$ and $\gamma \neq l$. They also ensure that the low-rank updates for the $u$-th task are approximately orthogonal to those of the $i$-th ($i=1, \cdots, u{-}1$) previous task: $\langle \Delta\widehat{\mathbf{W}}_u^l, \Delta\widehat{\mathbf{W}}_i^{l^\prime} \rangle \approx 0$, where $l^\prime = 1, \cdots, L$. The above strict orthogonality allows $\{\Delta\widehat{\mathbf{W}}_u^l\}_{l=1}^L$ to capture distinctive properties of the $u$-th task, while minimizing interference from irrelevant attributes both across layers in the current task and from previous tasks. However, such a strict orthogonality constraint may hinder the learning of semantically similar concepts with closely related attributes. Considering that adjacent layers tend to capture similar attributes, we fix the parameters of $\{\mathbf{A}_u^l\}_{l=1}^L$ in either the odd or even layers to address the above issue. 
It ensures orthogonality of AD-LoRA across non-adjacent layers and different tasks, which minimizes interference from irrelevant attributes of old concepts in learning new tasks.

\textbf{Proof:}
To prove $\langle \Delta\widehat{\mathbf{W}}_u^l, \Delta\widehat{\mathbf{W}}_i^{l^{\prime}}\rangle\approx 0$, we define $\Delta\widehat{\mathbf{W}}_i^{l^\prime} = \mathbf{A}_i^{{l^\prime}}(\mathbf{B}_i^{l^\prime}\odot \mathbf{M}_i^{l^\prime})$ via Eq.~\eqref{eq: new_low_rank_update}, and have the following deduction:
\begin{align}
\langle \Delta\widehat{\mathbf{W}}_u^l, \Delta\widehat{\mathbf{W}}_i^{l^\prime}\rangle &= \mathrm{Tr} \big( (\Delta\widehat{\mathbf{W}}_u^l)^\top \Delta\widehat{\mathbf{W}}_i^{l^\prime} \big) 
\label{eq: orthogonality_prove} \\ \nonumber
& = \mathrm{Tr}\big( (\mathbf{B}_u^l\odot \mathbf{M}_u^l)^\top (\mathbf{A}_u^l)^\top   \mathbf{A}_i^{l^\prime}(\mathbf{B}_i^{l^\prime}\odot \mathbf{M}_i^{l^\prime})\big).
\end{align}
Following \cite{goldstein2018phasemax}, we demonstrate that when both $\mathbf{A}_i^{l^\prime}$ and $\mathbf{A}_u^l$ are independently sampled from a high-dimensional standard Gaussian distribution $\mathcal{N}(0, 1)$, their inner product satisfies $\langle \mathbf{A}_u^l, \mathbf{A}_i^{l^\prime} \rangle\approx 0$ with high probability. This completes the proof that the orthogonality condition $\langle \Delta\widehat{\mathbf{W}}_u^l, \Delta\widehat{\mathbf{W}}_i^{l^\prime}\rangle\approx 0$ in Eq.~\eqref{eq: orthogonality_prove} holds. By analogous reasoning, we further verify the orthogonality relation $\langle \Delta\widehat{\mathbf{W}}_u^l, \Delta\widehat{\mathbf{W}}_u^{\gamma} \rangle \approx 0$ between the low-rank updates of different layers within the $u$-th task.

\subsection{Relevance-Guided AD-LoRA Aggregation}\label{sec: AD_LoRA_Aggregation}
Despite effectively capturing unique concept attributes within each task, the AD-LoRA module in Sec.~\ref{sec: AD_LoRA} cannot explore the underlying relationships between previously learned concepts and new concepts, nor can it use their positive relations to facilitate continual learning of new tasks in CIVC. To address the above issues, as shown in Fig.~\ref{fig: model_pipeline_ab}, we propose relevance-guided AD-LoRA aggregation. It adaptively merges the low-rank updates from the new and previous tasks based on their semantic relevance, thereby fully leveraging positive inter-task relations to enhance the learning of the $u$-th new task. Specifically, we preserve all task learners $\{\mathcal{Z}_i\}_{i=1}^{u{-}1}$ learned from the previous $(u{-}1)$ tasks to learn the $u$-th task, where $\mathcal{Z}_i = \{\Delta\widehat{\mathbf{W}}_i^l, \widehat{\mathbf{y}}_i^l\}_{l=1}^L$ indicates the task learner of the $i$-th task ($i=1, \cdots, u{-}1$), $\widehat{\mathbf{y}}_i^l = \{\widehat{\mathbf{y}}_i^{l,k}\}_{k=1}^{C_i}$, and $\widehat{\mathbf{y}}_i^{l,k}$ is the $k$-th concept token acquired in the $l$-th layer.

As aforementioned in Sec.~\ref{sec: AD_LoRA}, at the $u$-th customization task, given an input tuple $\{\mathbf{x}_u^s, \mathbf{p}_u^s, \mathbf{y}_u^s\} \in \mathcal{T}_u$ during training, we utilize the text encoder $\Psi(\cdot)$ to extract the layer-wise textual embedding $\mathbf{c}_u^{s, l} = \Psi(\mathbf{p}_u^{s, l}) \in \mathbb{R}^{N_p \times d}$ for the text prompt $\mathbf{p}_u^{s, l}$ at the $l$-th layer. 
Considering the set of stored task learners $\{\mathcal{Z}_i\}_{i=1}^{u{-}1}$, all concept tokens accumulated up to the current task can be represented as $\{\Omega_{u{-}1}^l\}_{l=1}^L$, where $\Omega_{u{-}1}^l = \cup_{i=1}^{u-1} \widehat{\mathbf{y}}_i^l \in \mathbb{R}^{K_{u{-}1}}$ comprises $K_{u{-}1} = \sum_{i=1}^{u-1} C_i$ concept tokens derived from the $l$-th layer of $\epsilon_{\theta_u^\prime}(\cdot)$. Subsequently, $\Psi(\cdot)$ maps the set of concept tokens $\{\Omega_{u{-}1}^l\}_{l=1}^L$ to the corresponding latent representations $\{\mathbf{e}_{u{-}1}^l \in\mathbb{R}^{K_{u{-}1}\times d}\}_{l=1}^L$. For each $\mathbf{e}_{u{-}1}^l$, the latent representations corresponding to the same task are averaged to derive a new representation $\widehat{\mathbf{e}}_{u{-}1}^l \in \mathbb{R}^{(u{-}1) \times d}$, where $(u{-}1)$ denotes the total number of previously learned tasks. Then we calculate the semantic affinity $\mathcal{R}_{u{-}1}^l \in\mathbb{R}^{u{-}1}$ between $\widehat{\mathbf{e}}_{u{-}1}^l$ and $\mathbf{c}_u^{s, l}$, and use $\mathcal{R}_{u{-}1}^l$ to determine the indices $\beta_{u{-}1}^l \in\mathbb{R}^{\lambda}$ of the $\lambda$ previous tasks ($\lambda\leq u{-}1$) that are most semantically similar to the $u$-th task: 
\begin{align}
\!\! \beta_{u{-}1}^l \!=\! \mathrm{max}\text{-}\mathrm{idx}_{[1:\lambda]} (\mathcal{R}_{u{-}1}^l),  \mathcal{R}_{u{-}1}^l \!=\!\max\!\big( \widehat{\mathbf{e}}_{u{-}1}^l \!\cdot\! (\mathbf{c}_u^{s, l})^\top \big),\!\!
\label{eq: select_task_idx}
\end{align}
where $\max(\cdot)$ selects the maximum value in each row, and $\mathrm{max}\text{-}\mathrm{idx}_{[1:\lambda]}(\cdot)$ obtains the indices of top $\lambda$ maximum values.

Subsequently, we leverage the underlying semantic relationships between the new task and the $\lambda$ selected previous tasks to facilitate the learning of the $u$-th task during training. Specifically, at the $l$-th layer, we dynamically merge their low-rank updates to obtain a new update $\Delta\widetilde{\mathbf{W}}_u^l$ by adaptively learning the combination weights within the network itself:
\begin{align}
\Delta\widetilde{\mathbf{W}}_u^l &= \sum_{i=1}^{\lambda+1} \sigma(\mathbf{W}_p^l\mathbf{E}(\mf{x}_u^s))_i \cdot \Delta\widehat{\mathbf{W}}_{\xi_i}^l \nonumber \\
& = \sum_{i=1}^{\lambda+1} \sigma(\mathbf{W}_p^l\mathbf{E}(\mf{x}_u^s))_i \cdot \mathbf{A}_{\xi_i}^l(\mathbf{B}_{\xi_i}^l\odot \mathbf{M}_{\xi_i}^l),
\label{eq: merge_low_rank_training}
\end{align}
where $\xi_i = \beta_{u{-}1}^l[i]$ denotes the $i$-th element of $\beta_{u{-}1}^l$. To learn the low-rank updates ${\Delta\widehat{\mathbf{W}}_u^l}$ for the $u$-th task, we include the current task index by setting $\beta_{u{-}1}^l[\lambda{+}1] = u$, resulting in $\beta_{u{-}1}^l \in \mathbb{R}^{\lambda{+}1}$. $\mathbf{E}(\mf{x}_u^s)\in\mathbb{R}^d$ is the global feature of given image $\mf{x}_u^s$, $\mathbf{W}_p^l\in\mathbb{R}^{(\lambda+1)\times d}$ is a trainable projection matrix, and $\sigma(\cdot)$ denotes the softmax function. $\sigma(\mathbf{W}_p^l\mathbf{E}(\mf{x}_u^s))_i$ represents the $i$-th element of $\sigma(\mathbf{W}_p^l\mathbf{E}(\mf{x}_u^s)) \in\mathbb{R}^{\lambda+1}$. As a result, for the 
denoising UNet $\epsilon_{\theta^\prime_u}(\cdot)$ at the $u$-th task, the parameter update $\Delta\theta_u$ can be reformulated as $\Delta\theta_u = \{\Delta\widetilde{\mathbf{W}}_u^l\}_{l=1}^L$.

\textbf{Optimization:}
When learning the $u$-th concept customization task, we freeze the parameters $\bigcup_{i=1}^\lambda \{\Delta \widehat{\mathbf{W}}_{\xi_i}^l\}_{l=1}^L$ from the $\lambda$ selected previous tasks, and optimize only the low-rank weights $\{\Delta\widehat{\mathbf{W}}_u^l\}_{l=1}^L$ and the learnable projection matrices $\{\mathbf{W}_p^l\}_{l=1}^L$ of the $u$-th task during training. This freezing operation prevents the overwriting of previously acquired knowledge, while exploiting the positive relationships between new and old tasks to learn task-specific knowledge for the $u$-th task, thereby mitigating the catastrophic forgetting of previously learned personalized concepts. In addition, selecting only the $\lambda$ most semantically similar tasks, instead of all tasks, improves training efficiency. After determining the mask $\mathbf{M}_u^l$ via Eq.~\eqref{eq: definition_mask}, the optimization update of $\mathbf{B}_u^l$ at the $l$-th layer of the UNet $\epsilon_{\theta^\prime_u}(\cdot)$ is formulated as follows: 
\begin{align}
\mathbf{B}_u^l\leftarrow \mathbf{B}_u^l - \vartheta\cdot\big( (\nabla_{\mathbf{B}_u^l}\mathcal{L}_{\mr{CDMs}})\odot \mathbf{M}_u^l\big),
\label{eq: update_B_l}
\end{align}
where $\vartheta$ denotes the learning rate to update $\mathbf{B}_u^l$ in this paper.

\textbf{Inference:}
Upon completing the learning of the $u$-th task, we store its task learner $\mathcal{Z}_u = \{\Delta\widehat{\mathbf{W}}_u^l, \widehat{\mathbf{y}}_u^l\}_{l=1}^L$ and merge it with the learners from previously learned tasks to form $\{\mathcal{Z}_i\}_{i=1}^u$ for inference. It is worth noting that the memory overhead required to retain $\mathcal{Z}_i$ (where $i$ ranges from $1$ to $u$) constitutes only 0.15\% of the total memory of $\epsilon_{\theta_0}(\cdot)$, which does not affect the practical deployment of CIVC. 
Given the text prompt for inference, denoted as $\widehat{\mathbf{p}}$, we employ it to initialize a set of layer-wise text prompts $\{\widehat{\mathbf{p}}^l\}_{l=1}^L$. These layer-wise prompts are then processed by the text encoder $\Psi(\cdot)$ to extract their corresponding textual embeddings $\widehat{\mathbf{c}} = \{\widehat{\mathbf{c}}^l \in \mathbb{R}^{N_p \times d}\}_{l=1}^L$. Subsequently, we collect $K_u = \sum_{i=1}^{u} C_i$ concept tokens from the $u$ task learners $\{\mathcal{Z}_i\}_{i=1}^u$ to obtain $\{\Omega_u^l\}_{l=1}^L$, where $\Omega_{u}^l = \cup_{i=1}^{u} \widehat{\mathbf{y}}_i^l \in \mathbb{R}^{K_u}$. After projecting $\{\Omega_u^l\}_{l=1}^L$ into latent textual embeddings via $\Psi(\cdot)$ and averaging the embeddings corresponding to the same task to obtain $\{\widehat{\mathbf{e}}_u^l \in \mathbb{R}^{u \times d}\}_{l=1}^L$, we measure the semantic correlations $\mathcal{R}_u^l =\max( \widehat{\mathbf{e}}_u^l \cdot (\widehat{\mathbf{c}}^l)^\top)\in \mathbb{R}^{u}$ between $\widehat{\mathbf{e}}_u^l$ and $\widehat{\mathbf{c}}^l$. Then, $\mathcal{R}_u^l$ is utilized to select the indices $\beta_u^l = \mathrm{max}\text{-}\mathrm{idx}_{[1:\lambda]} (\mathcal{R}_u^l) \in \mathbb{R}^{\lambda}$ of the $\lambda$ learned tasks that are most semantically similar to the given text prompt $\widehat{\mathbf{p}}$.

Since the given text prompt $\widehat{\mathbf{p}}$ is variable during testing and may differ from the training samples, we cannot store its corresponding projection matrices $\{\mathbf{W}_p^l\}_{l=1}^L$ to fuse the low-rank weights of the $\lambda$ selected tasks via Eq.~\eqref{eq: merge_low_rank_training} for inference. To tackle this issue, we concatenate the semantic affinities of the $\lambda$ selected tasks to obtain $\mathcal{M}^l = \{\mathcal{R}_u^l[\xi_i]\}_{i=1}^\lambda \in\mathbb{R}^{\lambda}$, where  $\xi_i = \beta_{u}^l[i]$ and $\mathcal{R}_u^l[\xi_i]$ denotes the $\xi_i$-th element of $\mathcal{R}_u^l$. Subsequently, we employ $\mathcal{M}^l$ to
adaptively combine the low-rank weights of $\lambda$ selected tasks, yielding a new low-rank weight $\Delta\widehat{\mathbf{W}}_*^l$ at the $l$-th transformer layer for inference: 
\begin{align}
\Delta\widehat{\mathbf{W}}_*^l = \sum_{i=1}^\lambda \psi(\mathcal{M}^l)_i \cdot \mathbf{A}_{\xi_i}^l(\mathbf{B}_{\xi_i}^l\odot \mathbf{M}_{\xi_i}^l), 
\label{eq: merge_low_rank_inference}
\end{align}
where $\psi(\mathcal{M}^l) = (\mathcal{M}^l)^2/\|(\mathcal{M}^l)^2\|_F \in\mathbb{R}^\lambda$ aims to normalize $\mathcal{M}^l$, and $\psi(\mathcal{M}^l)_i$ denotes the $i$-th entry of $\psi(\mathcal{M}^l)$. After obtaining the new low-rank weights $\{\Delta\widehat{\mathbf{W}}_*^l\}_{l=1}^L$ across different layers via Eq.~\eqref{eq: merge_low_rank_inference}, we derive a new denoising UNet $\epsilon_{\theta_*^\prime}(\cdot)$ for inference. Here, $\theta^\prime_* = \theta_0+\Delta\theta_*$ and $\Delta \theta_* = \{\Delta \widehat{\mathbf{W}}_*^l\}_{l=1}^L$. 
It is worth mentioning that $\theta^\prime_*$ captures the task-specific distinctive properties of all learned personalized concepts that are semantically close to the given text prompt $\widehat{\mathbf{p}}$,
while discarding their irrelevant attributes under the CIVC setting. As a result, $\theta^\prime_*$ plays a critical role in addressing catastrophic forgetting of previously learned concepts during inference.

\begin{figure}[t]
\centering
\includegraphics[width = 1.0\linewidth]
{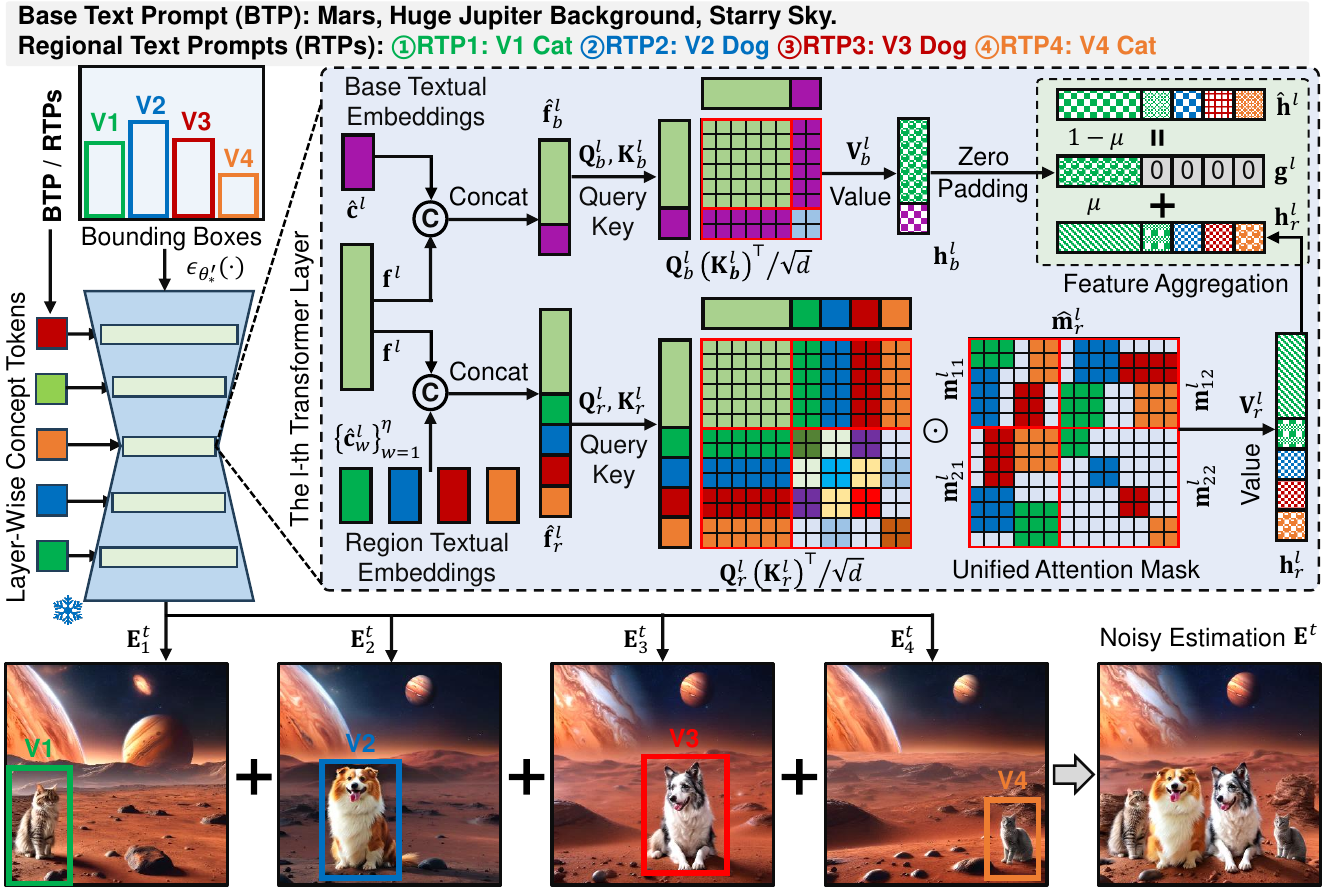}
\vspace{-22pt}
\caption{Demonstration of the controllable regional context synthesis module, using text-to-image customization as an example to illustrate the pipeline of our CCDM for mitigating concept neglect in the CIVC setting. } 
\label{fig: model_pipeline_c}
\end{figure}

\subsection{Controllable Regional Context Synthesis}\label{sec: regional_context_synthesis}
Under the CIVC setting, directly applying $\epsilon_{\theta^\prime_*}(\cdot)$ obtained in Sec.~\ref{sec: AD_LoRA_Aggregation} for multi-concept customization fails to produce high-quality images that adhere to user-specified conditions (such as bounding boxes \cite{101007978031_731167_14, Li_2023_CVPR}). This is because the generation of certain personalized concepts within specified regions is often overlooked, a phenomenon referred to as concept neglect \cite{11453592116_Chefer}. To mitigate concept neglect, several studies \cite{NEURIPS2024_DongJH, gu2023mixofshow} propose regional cross-attention between the image feature map and textual embeddings, aiming to independently associate each intended concept with its corresponding user-provided region. However, they fail to maintain seamless transitions at the boundaries of different bounding boxes within the holistic composition. To address this issue, we develop a controllable regional context synthesis module, as depicted in Fig.~\ref{fig: model_pipeline_c}. It uses a unified attention mask to restrict self-attention among image features to user-provided regions, while enabling cross-attention from image features to region textual embeddings. This module guarantees the semantic independence of distinct bounding boxes and facilitates smooth transitions at their boundaries, thereby enhancing the overall consistency of multi-concept composition.

\textbullet\textbf{Regional Concept Generation:}
In addition to the base text prompt (BTP) $\widehat{\mathbf{p}}$ introduced in Sec.~\ref{sec: AD_LoRA_Aggregation}, users can also specify $\eta$ region conditions, denoted as $\{\widehat{\mathbf{p}}_w, \widehat{\mathbf{b}}_w \}_{w=1}^{\eta}$, to control the context of multi-concept customization. Here, $\widehat{\mathbf{p}}_w$ is the $w$-th regional text prompt (RTP), which is used to generate the corresponding concept within the bounding box $\widehat{\mathbf{b}}_w\in\mathbb{R}^4$. The text encoder $\Psi(\cdot)$ is employed to derive the layer-wise textual embeddings $\widehat{\mathbf{c}}_w = \{\widehat{\mathbf{c}}_w^l \in \mathbb{R}^{N_w \times d}\}_{l=1}^L$ for $\widehat{\mathbf{p}}_w$, where $N_w$ is the prompt length of $\widehat{\mathbf{p}}_w$. Subsequently, the new denoising UNet $\epsilon_{\theta^\prime_*}(\cdot)$ acquired from Sec.~\ref{sec: AD_LoRA_Aggregation} extracts layer-wise image features $\{\mathbf{f}^l \in\mathbb{R}^{h^l\times w^l\times d}\}_{l=1}^L$ by incorporating with the base text prompt $\widehat{\mathbf{p}}$. $h^l$ and $w^l$ represent the height and width of $\mathbf{f}^l$. 
To improve the holistic coherence of multi-concept composition, we concatenate $\mathbf{f}^l$ with all region textual embeddings $\{\widehat{\mathbf{c}}_w^l\}_{w=1}^{\eta}$ to obtain $\widehat{\mathbf{f}}_r^l \in\mathbb{R}^{N_f^l\times d}$ at the $l$-th layer of $\epsilon_{\theta^\prime_*}(\cdot)$, and define its unified attention mask as $\widehat{\mathbf{m}}_r^l \in\mathbb{R}^{N_f^l\times N_f^l}$, where $N_f^l = h^lw^l+\sum_{w=1}^{\eta}N_w$. For multi-concept synthesis, we utilize $\widehat{\mathbf{m}}_r^l$ to mask the self-attention of $\widehat{\mathbf{f}}_r^l$, deriving a new region embedding $\mathbf{h}_r^l \in\mathbb{R}^{N_f^l\times d}$: 
\begin{align}
\!\mathbf{h}_r^l = \mathrm{Atten}(\mathbf{Q}_r^l, \mathbf{K}_r^l, \widehat{\mathbf{m}}_r^l)\mathbf{V}_r^l 
= \sigma\big(\frac{\mathbf{Q}_r^l (\mathbf{K}_r^l)^\top}{\sqrt{d}} \odot \widehat{\mathbf{m}}_r^l \big) \mathbf{V}_r^l, 
\label{eq: attention_mask_new_feature}
\end{align}
where $\mathrm{Atten}(\cdot)$ represents the self-attention operation. The query, key, and value matrices are defined as $\mathbf{Q}_r^l = \widehat{\mathbf{f}}_r^l\mathbf{w}_q \in \mathbb{R}^{N_f^l \times d}$, $\mathbf{K}_r^l = \widehat{\mathbf{f}}_r^l\mathbf{w}_k \in \mathbb{R}^{N_f^l \times d}$, and $\mathbf{V}_r^l = \widehat{\mathbf{f}}_r^l\mathbf{w}_v \in \mathbb{R}^{N_f^l \times d}$, respectively. $\mathbf{w}_q, \mathbf{w}_k, \mathbf{w}_v\in\mathbb{R}^{d\times d}$ are the projection matrices.

In this paper, $\widehat{\mathbf{m}}_r^l$ consists of four parts, and is defined as:
\begin{align}
\widehat{\mathbf{m}}_r^l = \left[\begin{matrix}
\mathbf{m}_{11}^l & \mathbf{m}_{12}^l \\
\mathbf{m}_{21}^l & \mathbf{m}_{22}^l  
\end{matrix}\right], 
\label{eq: four_part_mask}
\end{align}
where $\mathbf{m}_{11}^l \in\mathbb{R}^{h^lw^l\times h^lw^l}$ restricts self-attention among image features exclusively to user-provided regions. Notably, this strategy enables the generation of regional concepts to be more coherent with the overall creation, while also emphasizing the importance of each region concept, thereby mitigating concept neglect. Then $\mathbf{m}_{11}^l$ is formulated as:
\begin{align}
\mathbf{m}_{11}^l = \sum_{w=1}^{\eta} \mathrm{RS}\big( \Phi(\widehat{\mathbf{b}}_w)\big) \cdot \big(\mathrm{RS}(\Phi(\widehat{\mathbf{b}}_w)) \big)^\top, 
\label{eq: definition_m11}
\end{align}
where $\Phi(\widehat{\mathbf{b}}_w) \in\mathbb{R}^{h^l\times w^l}$ outputs a binary region mask for the $w$-th bounding box $\widehat{\mathbf{b}}_w$. The values of $\Phi(\widehat{\mathbf{b}}_w)$ are set to 1 within $\widehat{\mathbf{b}}_w$ and to 0 elsewhere. $\mathrm{RS}(\cdot)$ reshapes $\Phi(\widehat{\mathbf{b}}_w)$ into a vector of size $h^lw^l \times 1$. 
Moreover, $\mathbf{m}_{12}^l \in\mathbb{R}^{h^lw^l\times (\sum_{w=1}^{\eta}N_w)}$ serves as a mask for the cross-attention from image features to region textual embeddings, satisfying $\mathbf{m}_{12}^l = (\mathbf{m}_{21}^l)^\top$. It ensures that the $w$-th bounding box $\widehat{\mathbf{b}}_w$ generates its corresponding concept $\widehat{\mathbf{p}}_w$. As a result, we express $\mathbf{m}_{12}^l$ as:
\begin{align}
\!\!\mathbf{m}_{12}^l \!=\! \big[ \mathrm{RS}( \Phi(\widehat{\mathbf{b}}_1)) \!\otimes\! \mathbf{1}_{N_1\times 1}, \cdots, \mathrm{RS}( \Phi(\widehat{\mathbf{b}}_{\eta})) \!\otimes\! \mathbf{1}_{N_\eta\times 1} \big],\!\!\!
\label{eq: definition_m12}
\end{align}
where $\mathbf{1}_{N_w\times 1}$ is a $N_w\times 1$ all-ones vector ($w = 1, \cdots, \eta$). $\otimes$ is the outer product. To preserve the independence of each text prompt, we define $\mathbf{m}_{22}^l \in\mathbb{R}^{ (\sum_{w=1}^{\eta}N_w)\times (\sum_{w=1}^{\eta}N_w)}$ below:
\begin{align}
\mathbf{m}_{22}^l = \mathrm{diag}(\mathbf{1}_{N_1\times 1}, \mathbf{1}_{N_2\times 1}, \cdots, \mathbf{1}_{N_\eta\times 1}),
\label{eq: definition_m22}
\end{align}
where $\mathrm{diag}(\cdot)$ transforms a vector into a diagonal matrix.

Subsequently, the feature map $\mathbf{f}^l \in\mathbb{R}^{h^l\times w^l \times d}$ is concatenated with the textual embedding $\widehat{\mathbf{c}}^l \in\mathbb{R}^{N_p\times d}$ of base prompt $\widehat{\mathbf{p}}$ to derive $\widehat{\mathbf{f}}_b^l \in\mathbb{R}^{(h^lw^l+N_p)\times d}$. After computing the query, key, and value matrices $\mathbf{Q}_b^l, \mathbf{K}_b^l, \mathbf{V}_b^l\in\mathbb{R}^{(h^lw^l+N_p)\times d}$ for $\widehat{\mathbf{f}}_b^l$, we formulate the new base embedding as $\mathbf{h}_b^l = \mathrm{Atten}(\mathbf{Q}_b^l, \mathbf{K}_b^l)\mathbf{V}_b^l 
= \sigma(\mathbf{Q}_b^l (\mathbf{K}_b^l)^\top / \sqrt{d} ) \mathbf{V}_b^l \in\mathbb{R}^{(h^lw^l+N_p)\times d}$. Here, $\mathbf{Q}_b^l=\widehat{\mathbf{f}}_b^l\mathbf{w}_q$, $\mathbf{K}_b^l=\widehat{\mathbf{f}}_b^l\mathbf{w}_k$, and $\mathbf{V}_b^l=\widehat{\mathbf{f}}_b^l\mathbf{w}_v$. To further ensure smooth transitions at the boundaries across different bounding boxes, we fuse $\mathbf{h}_r^l \in\mathbb{R}^{N_f^l\times d}$ and $\mathbf{h}_b^l \in\mathbb{R}^{(h^lw^l+N_p)\times d}$ to obtain a new embedding $\widehat{\mathbf{h}}^l \in\mathbb{R}^{N_f^l\times d}$ at the $l$-th layer. To achieve this, we select the first $h^lw^l$ rows of $\mathbf{h}_b^l$ and pad them with an all-zero matrix $\mathbf{0}_{(\sum_{w=1}^\eta N_w) \times d}$ to derive $\mathbf{g}^l \in\mathbb{R}^{N_f^l\times d}$. Then we formulate $\widehat{\mathbf{h}}^l$ as follows: 
\begin{align}
\widehat{\mathbf{h}}^l = \mu\mathbf{h}_r^l + (1-\mu)\mathbf{g}^l,
\label{eq: feature_fusion} 
\end{align} 
where $\mu\in(0,1)$ is the balancing weight to fuse $\mathbf{h}_r^l$ and $\mathbf{g}^l$.

\textbullet\textbf{Noise Estimation:}
At the $t$-th timestep, by incorporating the base embeddings $\{\mathbf{h}_b^l\}_{l=1}^L$ into the new denoising UNet $\epsilon_{\theta^\prime_*}(\cdot)$, we obtain the output $\mathcal{O}_b^t \in\mathbb{R}^{(h^Lw^L+N_p)\times d}$ from the $L$-th layer, where $h^L$ and $w^L$ denote the height and width of $\mathbf{f}^L$ at the $L$-th layer. 
We then select the first $h^Lw^L$ rows of $\mathcal{O}_b^t$ as the noise estimation $\epsilon_{\theta^\prime_*} (\mathbf{z}_t| \widehat{\mathbf{c}}, t) \in\mathbb{R}^{h^Lw^L\times d}$, which is conditioned on the base textual embeddings $\widehat{\mathbf{c}}$ of $\widehat{\mathbf{p}}$ \cite{ho2021classifierfree}. After deriving the unconditional noise estimation $\epsilon_{\theta^\prime_*} (\mathbf{z}_t|t) \in\mathbb{R}^{h^Lw^L\times d}$ produced by the $L$-th layer, we formulate the base noise estimation $\mathbf{E}_b^t \in\mathbb{R}^{h^Lw^L\times d}$ for the base prompt $\widehat{\mathbf{p}}$ as: 
\begin{align}
\mathbf{E}_b^t = \epsilon_{\theta^\prime_*} (\mathbf{z}_t|t) + s(\epsilon_{\theta^\prime_*} (\mathbf{z}_t| \widehat{\mathbf{c}}, t) - \epsilon_{\theta^\prime_*} (\mathbf{z}_t|t)),
\label{eq: base_noisy_estimation}
\end{align}
where $s=7.5$ is the scaling factor.
Analogous to $\mathbf{E}_b^t$, we also incorporate the new embeddings $\{\widehat{\mathbf{h}}^l \}_{l=1}^L$ of the $\eta$ region conditions $\{\widehat{\mathbf{p}}_w, \widehat{\mathbf{b}}_w \}_{w=1}^{\eta}$, obtained from Eq.~\eqref{eq: feature_fusion}, into the UNet $\epsilon_{\theta_*^\prime}(\cdot)$ to produce the output $\mathcal{O}_r^t \in\mathbb{R}^{N_f^L\times d}$ of the $L$-th layer, where $N_f^L = h^Lw^L+\sum_{w=1}^{\eta}N_w$. Subsequently, we regard the first $h^Lw^L$ rows of $\mathcal{O}_r^t$ as the noise estimation $\epsilon_{\theta^\prime_*} (\mathbf{z}_t| \{\widehat{\mathbf{c}}_w, \widehat{\mathbf{b}}_w\}_{w=1}^{\eta}, t) \in\mathbb{R}^{h^Lw^L\times d}$, which is conditioned on $\{\widehat{\mathbf{c}}_w, \widehat{\mathbf{b}}_w \}_{w=1}^{\eta}$. Consequently, the region noise estimation $\mathbf{E}_r^t \in\mathbb{R}^{h^Lw^L\times d}$ at the $t$-th timestep is defined below:
\begin{align}
\!\!\mathbf{E}_r^t = \epsilon_{\theta^\prime_*} (\mathbf{z}_t|t) + s(\epsilon_{\theta^\prime_*} (\mathbf{z}_t| \{\widehat{\mathbf{c}}_w, \widehat{\mathbf{b}}_w\}_{w=1}^{\eta}, t) - \epsilon_{\theta^\prime_*} (\mathbf{z}_t|t)).\!\!
\label{eq: region_noisy_estimation}
\end{align}

\begin{figure}[t]
\centering
\includegraphics[width = 0.68\linewidth]
{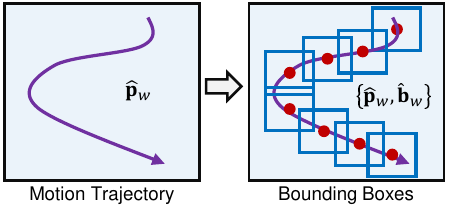}
\vspace{-15pt}
\caption{Illustration of transforming a motion trajectory into bounding boxes.  }
\label{fig: motion_trajectory}
\end{figure}

To surmount concept neglect during multi-concept customization, we leverage self-attention among user-provided regions to fuse
$\mathbf{E}_b^t$ and $\mathbf{E}_r^t$, thereby obtaining $\mathbf{E}^t \in\mathbb{R}^{h^Lw^L\times d}$:
\begin{align}
\mathbf{E}^t = \alpha\mathbf{E}_b^t + (1-\alpha)\cdot \mathrm{EXT}\big(\sigma(\frac{\mathbf{Q}_r^L (\mathbf{K}_r^L)^\top}{\sqrt{d}} \odot \widehat{\mathbf{m}}_r^L) \big)
\mathbf{E}_r^t,
\label{eq: aggregate_noisy_estimation}
\end{align}
where $\widehat{\mathbf{m}}_r^L \in\mathbb{R}^{N_f^L\times N_f^L}$, $\mathbf{Q}_r^L = \widehat{\mathbf{f}}_r^L\mathbf{w}_q \in\mathbb{R}^{N_f^L\times d}$ and $\mathbf{K}_r^L = \widehat{\mathbf{f}}_r^L\mathbf{w}_k \in\mathbb{R}^{N_f^L\times d}$ are the unified attention mask, query and key matrices of the feature $\widehat{\mathbf{f}}_r^L \in\mathbb{R}^{N_f^L\times d}$ at the $L$-th layer. Besides, $\mathrm{EXT}(\cdot)$ extracts the first $h^Lw^L$ rows and columns of the given matrix, thereby producing a $h^Lw^L\times h^Lw^L$ matrix. $\alpha=0.1$ is the combination weight. Subsequently, we reshape $\mathbf{E}^t$ into $\widehat{\mathbf{E}}_*^t \in\mathbb{R}^{h^L\times w^L\times d}$, and follow \cite{podell2024sdxl} to feed $\widehat{\mathbf{E}}_*^t$ into the denoising process for multi-concept composition.

\subsection{Extension to Text-to-Video Customization}
\label{sec: extension_text_to_video}
\begin{figure}[t]
\centering
\includegraphics[width = 1.0\linewidth]
{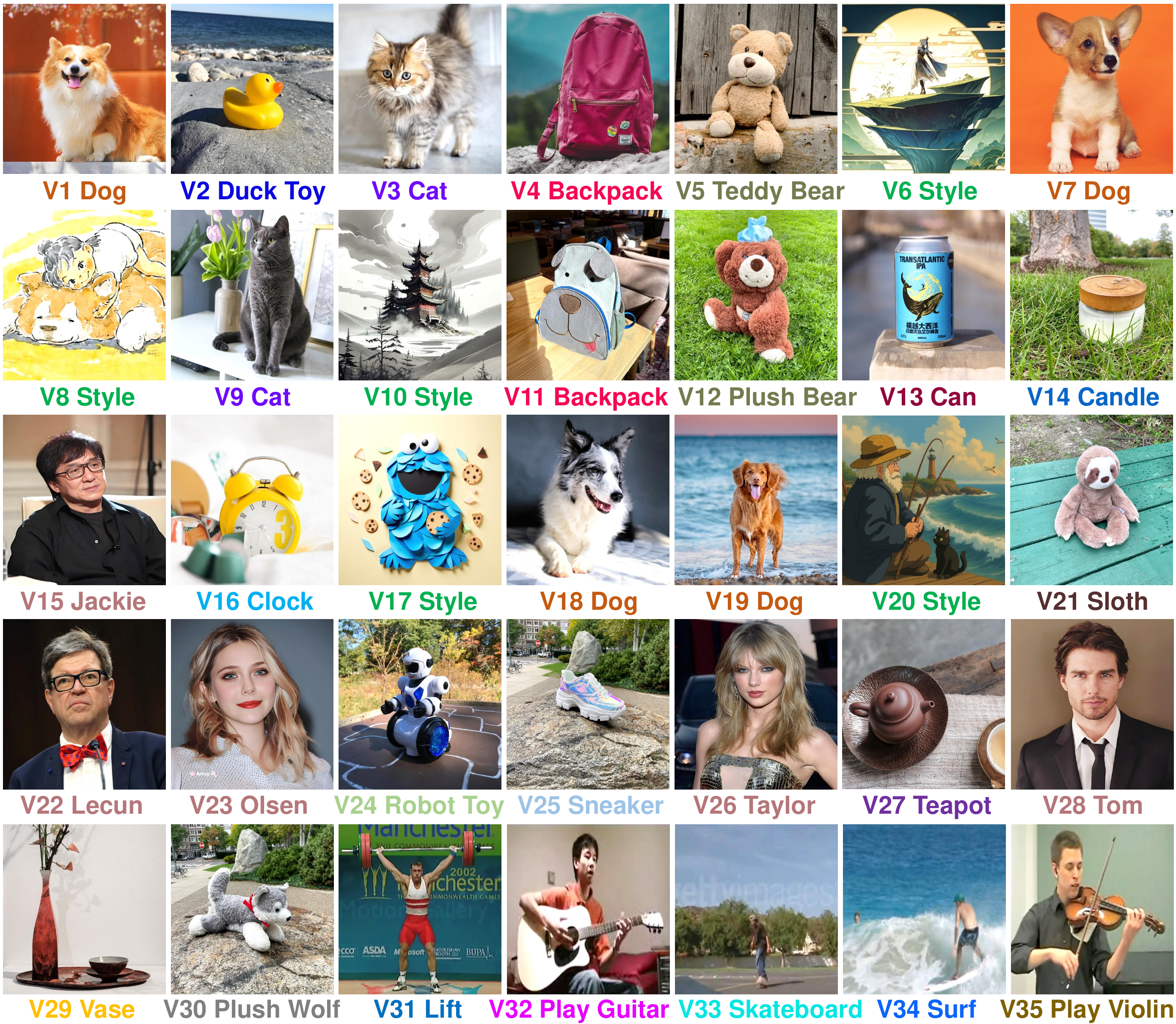}
\vspace{-25pt}
\caption{Visualization of exemplary cases from 35 continuous concept customization tasks in the CIL benchmark.  }
\label{fig: concept_dataset_vis}
\end{figure}

To extend the proposed CCDM to text-to-video customization under the CIVC setting, we introduce 5 motion concepts (V31--V35), in addition to the 30 subject concepts (V1--V30) used for text-to-image synthesis, as shown in Fig.~\ref{fig: concept_dataset_vis}. Specifically, we apply the proposed CCDM to the CogVideoX backbone \cite{yang2024cogvideox} to continually learn subject and motion concepts. Similar to text-to-image customization, when learning the $u$-th task $\mathcal{T}_u$, we initialize $L$ layer-wise text prompts $\{\mathbf{p}_u^{s,l}\}_{l=1}^L$ and $L$ corresponding layer-wise concept tokens $\{\mathbf{y}_u^{s,l}\}_{l=1}^L$ for each input tuple $\{\mathbf{x}_u^s, \mathbf{p}_u^s, \mathbf{y}_u^s\}$. Following Dualreal \cite{wang2025dualreal}, we apply AD-LoRA to the spatial transformer blocks of CogVideoX for learning subject concepts and to the temporal transformer blocks for learning motion concepts. Furthermore, we implement the proposed relevance-guided AD-LoRA aggregation strategy in both spatial and temporal transformer blocks, with each component optimized independently to learn subject and motion concepts, respectively. 
For multi-concept text-to-video customization, users can also provide a specific motion trajectory for each regional text prompt to control the generated contexts of videos. As presented in Fig.~\ref{fig: motion_trajectory}, given the $w$-th regional text prompt $\widehat{\mathbf{p}}_w$ and its motion trajectory, we automatically place a bounding box at each position along the motion trajectory to obtain frame-level bounding boxes $\widehat{\mathbf{b}}_w \in\mathbb{R}^{n_v\times 4}$ for $\widehat{\mathbf{p}}_w$. Here, $n_v$ is the number of frames in the synthesized video. Given $\eta$ region conditions $\{\widehat{\mathbf{p}}_w, \widehat{\mathbf{b}}_w \}_{w=1}^{\eta}$ and a base text prompt $\widehat{\mathbf{p}}$, we first use $\widehat{\mathbf{p}}$ to extract layer-wise video features $\{\mathbf{f}^l \in\mathbb{R}^{n_v\times h^l\times w^l\times d}\}_{l=1}^L$ via the denoising UNet $\epsilon_{\theta^\prime_*}(\cdot)$ in Sec.~\ref{sec: AD_LoRA_Aggregation}. For the $i$-th frame ($i=1, \cdots, n_v$), we utilize the frame feature $\mathbf{f}^l[i] \in \mathbb{R}^{h^l \times w^l \times d}$ together with the corresponding region conditions $\{\widehat{\mathbf{p}}_w, \widehat{\mathbf{b}}_w[i]\}_{w=1}^{\eta}$ to execute the proposed controllable regional context synthesis module (Sec.~\ref{sec: regional_context_synthesis}) within the $l$-th spatial transformer block, analogous to text-to-image customization. $\widehat{\mathbf{b}}_w[i] \in\mathbb{R}^4$ is the bounding box of the regional text prompt $\widehat{\mathbf{p}}_w$ at the $i$-th frame.

\subsection{Extension to Text-to-3D Customization}
\label{sec: extension_text_to_3D}
For text-to-3D customization under the CIVC setting, we adopt a two-stage pipeline: text-to-image generation followed by image-to-3D synthesis. Specifically, we first build the proposed CCDM on the SDXL backbone \cite{podell2024sdxl} to continually learn subject concepts (see V1--V30 in Fig.~\ref{fig: concept_dataset_vis}), allowing users to specify text prompts and some region conditions for single/multi-concept image generation and style transfer according to their preferences. After synthesizing reference images with our CCDM, we employ the Stable Video Diffusion (SVD) model \cite{blattmann2023stable} and the 3D-aware feedback module \cite{wen2025ouroboros3d} to generate multi-view-consistent images. Then, we use a large multi-view Gaussian model (LGM) \cite{tang2024lgm} to recover 3D Gaussian splatting \cite{tang2024lgm} for 3D representation reconstruction.

\begin{table*}[t]
\centering
\setlength{\tabcolsep}{1.1pt}
\renewcommand{\arraystretch}{1.2}
\caption{Quantitative performance of single-concept text-to-image customization under the CIVC setting. } 
\vspace{-3mm}
\resizebox{1.0\linewidth}{!}{
\begin{tabular}{l|c|cccc|c|cccc}
\toprule
\makecell[c]{\multirow{2}{*}{Methods}}  
& \multicolumn{5}{c|}{SDXL \cite{podell2024sdxl}} & \multicolumn{5}{c}{FLUX.1 \cite{flux2024}}  \\
& \#Params & ~~~~~IA ($\uparrow$)~~~~~ & ~~~~~TA ($\uparrow$)~~~~~ & ~~~~FIA ($\downarrow$)~~~~ & ~~~~FTA ($\downarrow$)~~~~ & \#Params  & ~~~~~IA ($\uparrow$)~~~~~ & ~~~~~TA ($\uparrow$)~~~~~ & ~~~~FIA ($\downarrow$)~~~~ & ~~~~FTA ($\downarrow$)~~~~  \\
\midrule
Finetuning   &	6.29M	&	60.41$\pm$0.22 	&	76.44$\pm$0.05 	&	14.84$\pm$0.02 	&	6.40$\pm$0.04 
&	10.27M	&	60.91$\pm$0.16 	&	78.60$\pm$0.10 	&	13.97$\pm$0.06 		&3.44$\pm$0.12 
 \\

EWC \cite{kirkpatrick2017overcoming} (NAS'2017) &	6.29M	&	65.67$\pm$0.28 	&	78.19$\pm$0.03 	&\ \ 	9.08$\pm$0.07 	&	3.68$\pm$0.03 &	10.27M	&	64.28$\pm$0.20 	&	78.06$\pm$0.09 	&	\ \ 7.51$\pm$0.06 	&	3.19$\pm$0.05 

\\

LWF \cite{li2017learning} (TPAMI'2017)&	6.29M & 64.03$\pm$0.21 	&	77.25$\pm$0.11 	&	10.33$\pm$0.10 	&	5.39$\pm$0.02 &	10.27M	&	64.30$\pm$0.16 	&	79.04$\pm$0.16 	&	\ \ 7.07$\pm$0.05 	&	2.86$\pm$0.06 

 \\

LoRA-M \cite{zhong2024multi} (TMLR'2024) &	6.29M	&	64.89$\pm$0.05 	&	80.28$\pm$0.07 	&\ \ 	9.44$\pm$0.07 	&	2.28$\pm$0.04 &	10.27M	&	63.18$\pm$0.09 	&	78.56$\pm$0.12 	&\ \	9.56$\pm$0.08 	&	3.50$\pm$0.11 

 \\
LoRA-C \cite{zhong2024multi} (TMLR'2024) &	6.29M	&	65.27$\pm$0.08 	&	80.06$\pm$0.08 	&\ \	9.57$\pm$0.05 	&	2.26$\pm$0.03 &	10.27M	&	62.09$\pm$0.08 	&	78.34$\pm$0.09 	&	10.21$\pm$0.09 	&	3.15$\pm$0.06 

  \\
CLoRA \cite{smith2023continual} (TMLR'2024) &	6.29M	&	69.32$\pm$0.14 	&	79.29$\pm$0.10 	&\ \	7.42$\pm$0.07 	&	3.31$\pm$0.02 &	10.27M	&	64.34$\pm$0.15 	&	78.20$\pm$0.14 	&\ \	8.80$\pm$0.10 	&	3.21$\pm$0.10 

  \\
L2DM \cite{sun2024create} (TPAMI'2024) &	6.29M	&	66.41$\pm$0.25 	&	78.35$\pm$0.08 	&\ \	8.78$\pm$0.04 	&	5.19$\pm$0.02 &	10.27M	&	65.42$\pm$0.24 	&	78.08$\pm$0.08 	&\ \	6.67$\pm$0.05 	&	3.05$\pm$0.08 

   \\
CIDM \cite{NEURIPS2024_DongJH} (NeurIPS'2024) &	6.43M	&	67.81$\pm$0.08 	&	80.13$\pm$0.04 	&\ \	8.35$\pm$0.05 	&	2.08$\pm$0.05 &	10.31M	&	65.82$\pm$0.10 	&	77.89$\pm$0.12 	&\ \	5.86$\pm$0.04 	&	4.01$\pm$0.04 

  \\
CPG \cite{Guo_2025_CVPR} (CVPR'2025) &	6.29M	&	66.08$\pm$0.17 	&	78.59$\pm$0.04 	&	10.14$\pm$0.04 	&	4.83$\pm$0.05 &	10.27M	&	65.40$\pm$0.08 	&	78.12$\pm$0.06 	&\ \	7.18$\pm$0.08 	&	3.54$\pm$0.06 

 \\
\midrule
\rowcolor{lightgray}
\textbf{Ours} (\textbf{CCDM}) &	\textcolor{deepred}{\textbf{4.23M}}&	\textcolor{deepred}{\textbf{72.84$\pm$0.11}} &\textcolor{deepred}{\textbf{80.75$\pm$0.05}} &\ \	\textcolor{deepred}{\textbf{3.84$\pm$0.03}} & \textcolor{deepred}{\textbf{1.89$\pm$0.02}} &	\  \	\textcolor{deepred}{\textbf{6.72M}}	&	\textcolor{deepred}{\textbf{68.09$\pm$0.12}} 	&	\textcolor{deepred}{\textbf{80.11$\pm$0.08}} 	&\ \	\textcolor{deepred}{\textbf{3.04$\pm$0.04}} &	\textcolor{deepred}{\textbf{1.64$\pm$0.03}} \\
\midrule 
\end{tabular}}
\label{tab: quantitative_comp_T2I}
\vspace{-3pt}
\end{table*}

\begin{table*}[t]
\centering
\setlength{\tabcolsep}{1.1pt}
\renewcommand{\arraystretch}{1.2}
\caption{Quantitative performance of single-concept text-to-video and text-to-3D customization under the CIVC setting. }
\vspace{-3mm}
\resizebox{1.0\linewidth}{!}{
\begin{tabular}{l|c|ccccc|c|cccc}
\toprule
\makecell[c]{\multirow{2}{*}{Methods}}  
& \multicolumn{6}{c|}{Text-to-Video} & \multicolumn{5}{c}{Text-to-3D}  \\
& \#Params & ~~~~~IA ($\uparrow$)~~~~~ & ~~~~~TA ($\uparrow$)~~~~~ & ~~~~~DI ($\uparrow$)~~~~~ & ~~~~FIA ($\downarrow$)~~~~ & ~~~~FTA ($\downarrow$)~~~~ & \#Params  & ~~~~~IA ($\uparrow$)~~~~~ & ~~~~~TA ($\uparrow$)~~~~~ & ~~~~FIA ($\downarrow$)~~~~ & ~~~~FTA ($\downarrow$)~~~~  \\
\midrule
Finetuning   &	18.91M	&	60.23$\pm$0.28 	&	74.56$\pm$0.16 	&	41.05$\pm$0.20 	&	11.67$\pm$0.08 	&	3.54$\pm$0.04 
&	6.29M	&	60.98$\pm$0.18 	&61.34$\pm$0.14 	&11.48$\pm$0.13 	&9.56$\pm$0.08 

 \\

EWC \cite{kirkpatrick2017overcoming} (NAS'2017)&	18.91M	&	65.78$\pm$0.35 	&	73.89$\pm$0.08 	&	48.06$\pm$0.13 	&	\  \ 7.06$\pm$0.06 	&	3.18$\pm$0.06 &	6.29M	&	65.81$\pm$0.23 	&	70.18$\pm$0.05 	&\ \ 8.05$\pm$0.06 	&	2.54$\pm$0.10 

 \\

LWF \cite{li2017learning} (TPAMI'2017) &	18.91M	&	66.48$\pm$0.20 	&74.18$\pm$0.19 	&	48.37$\pm$0.14 	&\ \ 6.48$\pm$0.06 	&	3.10$\pm$0.08 &	6.29M	&	66.38$\pm$0.14 	&	69.40$\pm$0.16 	&\ \ 7.16$\pm$0.08 	&	3.48$\pm$0.04 

 \\

LoRA-M \cite{zhong2024multi} (TMLR'2024) &	18.91M	&	65.85$\pm$0.18 	&	75.06$\pm$0.04 	&	47.55$\pm$0.08 	&\ \	7.86$\pm$0.08 	&	2.03$\pm$0.06 &	6.29M	&	65.43$\pm$0.16 	&	70.05$\pm$0.08 	&\ \ 8.04$\pm$0.08 	&	2.38$\pm$0.06 

 \\
LoRA-C \cite{zhong2024multi} (TMLR'2024) &	18.91M	&	65.33$\pm$0.19 	&	75.38$\pm$0.06 	&	47.06$\pm$0.08 	&\ \	8.34$\pm$0.04 	&	1.98$\pm$0.04 &	6.29M	&	65.58$\pm$0.16 	&	70.37$\pm$0.09 	&\ \ 7.49$\pm$0.04 	&	2.45$\pm$0.12 

  \\
CLoRA \cite{smith2023continual} (TMLR'2024) &	18.91M	&	67.76$\pm$0.23 	&	74.90$\pm$0.12 	&	50.25$\pm$0.20 	&	\ \ 4.44$\pm$0.04 	&	2.19$\pm$0.12 &	6.29M	&	67.44$\pm$0.12 	&	69.71$\pm$0.12 	&\ \ 5.45$\pm$0.12 	&	3.04$\pm$0.03 

  \\
L2DM \cite{sun2024create} (TPAMI'2024) &	18.91M	&	67.49$\pm$0.14 	&	74.02$\pm$0.10 	&	50.53$\pm$0.12 	&\  \	4.85$\pm$0.06 	&	2.24$\pm$0.10 &	6.29M	&	67.25$\pm$0.08 	&	70.25$\pm$0.08 	&\ \ 5.18$\pm$0.07 	&	2.29$\pm$0.05 

  \\
CIDM \cite{NEURIPS2024_DongJH} (NeurIPS'2024) &	19.25M	&	68.49$\pm$0.18 	&	75.06$\pm$0.12 	&	50.36$\pm$0.08 	&	\  \ 3.04$\pm$0.08 	&	1.93$\pm$0.02 &	6.43M	&	68.15$\pm$0.13 	&	70.32$\pm$0.14 	&\ \ 	4.67$\pm$0.05 	&	2.03$\pm$0.03 \\
CPG \cite{Guo_2025_CVPR} (CVPR'2025)&	18.91M	&	67.06$\pm$0.16 	&	74.28$\pm$0.08 	&	50.01$\pm$0.12 	&\ \ 5.08$\pm$0.10 	&	2.83$\pm$0.04 &	6.29M	&	66.48$\pm$0.19 	&	69.37$\pm$0.18 &\ \  6.89$\pm$0.06 	&	4.01$\pm$0.04  \\
\midrule
\rowcolor{lightgray}
\textbf{Ours} (\textbf{CCDM}) & \textcolor{deepred}{\textbf{12.63M}}
 & \textcolor{deepred}{\textbf{70.96$\pm$0.14}} & \textcolor{deepred}{\textbf{76.30$\pm$0.08}} & \textcolor{deepred}{\textbf{52.98$\pm$0.16}} &\ \  \textcolor{deepred}{\textbf{1.69$\pm$0.03}} & \textcolor{deepred}{\textbf{1.34$\pm$0.02}} & \textcolor{deepred}{\textbf{4.23M}} & \textcolor{deepred}{\textbf{69.03$\pm$0.12}} & \textcolor{deepred}{\textbf{71.25$\pm$0.03}} & \ \ \textcolor{deepred}{\textbf{2.89$\pm$0.03}} & \textcolor{deepred}{\textbf{1.49$\pm$0.04}}  \\
\midrule 
\end{tabular}}
\label{tab: quantitative_comp_T2V_T23D}
\end{table*}

\begin{table}[t]
\centering
\setlength{\tabcolsep}{1.1pt}
\renewcommand{\arraystretch}{1.2}
\caption{Quantitative performance of multi-concept text-to-image customization under the CIVC setting using SDXL \cite{podell2024sdxl} as the backbone.  }
\vspace{-3mm}
\resizebox{1.0\linewidth}{!}{
\begin{tabular}{l|c|cccc}
\toprule
\makecell[c]{Methods} & \#Params & ~~~~~IA ($\uparrow$)~~~~~ & ~~~~~TA ($\uparrow$)~~~~~ & ~~~~FIA ($\downarrow$)~~~~ & ~~~~FTA ($\downarrow$)~~~~ \\
\midrule
Finetuning &6.29M &60.15$\pm$0.08 &72.13$\pm$0.10 &16.19$\pm$0.12 &7.96$\pm$0.17 \\

EWC \cite{kirkpatrick2017overcoming} (NAS'2017) &6.29M &65.87$\pm$0.13 &73.85$\pm$0.06 &10.56$\pm$0.14 &6.42$\pm$0.05 \\

LWF \cite{li2017learning} (TPAMI'2017)&	6.29M &66.57$\pm$0.06 &73.20$\pm$0.11 &10.18$\pm$0.10 &5.68$\pm$0.07 \\

LoRA-M \cite{zhong2024multi} (TMLR'2024) &	6.29M	&63.19$\pm$0.08 &71.14$\pm$0.12 &13.16$\pm$0.07 &7.50$\pm$0.13 \\
LoRA-C \cite{zhong2024multi} (TMLR'2024) &	6.29M	&64.20$\pm$0.15 &72.58$\pm$0.12 &12.19$\pm$0.09 &7.05$\pm$0.06 \\
CLoRA \cite{smith2023continual} (TMLR'2024) &	6.29M	&71.18$\pm$0.13 &73.31$\pm$0.16 &\ \ 6.44$\pm$0.22 &6.18$\pm$0.20 \\
L2DM \cite{sun2024create} (TPAMI'2024) &	6.29M	&70.18$\pm$0.10 &72.16$\pm$0.12 &\ \ 7.54$\pm$0.05 &6.65$\pm$0.11  \\
CIDM \cite{NEURIPS2024_DongJH} (NeurIPS'2024) &	6.43M &72.12$\pm$0.08 &74.23$\pm$0.04 &\ \ 6.05$\pm$0.06 &5.02$\pm$0.03  \\
CPG \cite{Guo_2025_CVPR} (CVPR'2025) &	6.29M	&70.25$\pm$0.16 &73.10$\pm$0.11 &\ \ 6.96$\pm$0.09 &5.67$\pm$0.14 \\
\midrule
\rowcolor{lightgray}
\textbf{Ours} (\textbf{CCDM}) &\textcolor{deepred}{\textbf{4.23M}} &	\textcolor{deepred}{\textbf{77.25$\pm$0.03}} &\textcolor{deepred}{\textbf{75.29$\pm$0.07}} &\ \ \textcolor{deepred}{\textbf{2.86$\pm$0.02}} &\textcolor{deepred}{\textbf{3.54$\pm$0.05}} \\
\midrule 
\end{tabular}}
\label{tab: quantitative_comp_multi_concept_image}
\vspace{-10pt}
\end{table}

\section{Experiments}

\subsection{Implementation Details}
\textbf{Dataset:}
Drawing motivation from previous methods \cite{NEURIPS2024_DongJH, sun2024create, smith2023continual, kumari2022customdiffusion}, we propose a novel concept-incremental learning (CIL) benchmark to evaluate model performance in text-to-image, text-to-video, and text-to-3D concept customization under the CIVC setting. Specifically, this benchmark comprises 20 object concepts, 5 style concepts (V6, V8, V10, V17 and V20), 5 person concepts (V15, V22, V23, V26 and V28), and 5 motion concepts (V31--V35), resulting in a total of 35 personalized concepts. Under the CIVC setting, as presented in Fig.~\ref{fig: concept_dataset_vis}, we utilize 30 continual concept generation tasks (V1–V30), excluding motion concepts, for both text-to-image and text-to-3D customization, while all concepts (V1–V35) are used for text-to-video customization. To reflect real-world CIVC scenarios, each task contains approximately 3--5 text-image pairs, with text prompts generated using BLIP \cite{li2022blip}. More importantly, we intentionally include several semantically similar concepts (\emph{e.g.}, V1, V7, V18, and V19 dogs; V4 and V11 backpacks; V3 and V9 cats) to increase the difficulty and challenge of the proposed CIL benchmark.

\textbf{Experimental Setups:}
Following \cite{wang2025msdiffusion}, we adopt SDXL \cite{podell2024sdxl} and FLUX.1 \cite{flux2024} as the backbone architectures for text-to-image customization, CogVideoX \cite{yang2024cogvideox} for text-to-video customization, and SDXL \cite{podell2024sdxl} for text-to-3D customization. To ensure fairness, all state-of-the-art baselines and our proposed model are trained under identical settings, including the same backbone and the Adam optimizer. The learning rates are configured as $1.0 \times 10^{-3}$ for textual embedding updates and $1.0 \times 10^{-4}$ for the denoising optimization of UNet. Following \cite{flux2024, NEURIPS2024_DongJH}, we set the low-rank parameter to $r=8$ for FLUX.1 and SDXL while configuring $r=32$ for CogVideoX. Furthermore, we set $\rho = 0.2$ to mask 20\% of the trainable parameters in $\mathbf{B}_u^l$, and $\mu = 0.1$ in Eq.~\eqref{eq: feature_fusion}. Additionally, based on empirical results in Tab.~\ref{tab: analysis_AD_LoRA_variants}, we fix the parameters of $\{\mathbf{A}_u^l\}_{l=1}^L$ in the even layers. 
To showcase our model's superiority in addressing the CIVC problem, we evaluate it against eight state-of-the-art methods spanning three categories: 1) continual learning approaches (LWF \cite{li2017learning} and EWC \cite{kirkpatrick2017overcoming}); 2) multi-LoRA integration methods (LoRA-M \cite{zhong2024multi} and LoRA-C \cite{zhong2024multi}); and 3) continual diffusion models (L2DM \cite{sun2024create}, CLoRA \cite{smith2023continual}, CIDM \cite{NEURIPS2024_DongJH} and CPG \cite{Guo_2025_CVPR}). 
For fair comparisons in multi-concept synthesis, we apply bounded attention \cite{101007978303172630_9_25} to CIDM \cite{NEURIPS2024_DongJH} (denoted as BoundAtt) and incorporate regionally controllable sampling \cite{gu2023mixofshow} into the other baselines. Since bounded attention is specifically designed for SDXL-based image generation, we only compare BoundAtt with our CCDM on SDXL-based text-to-image customization tasks under the CIVC setting.

\begin{figure*}[t]
\centering
\includegraphics[width = 1.0\linewidth]
{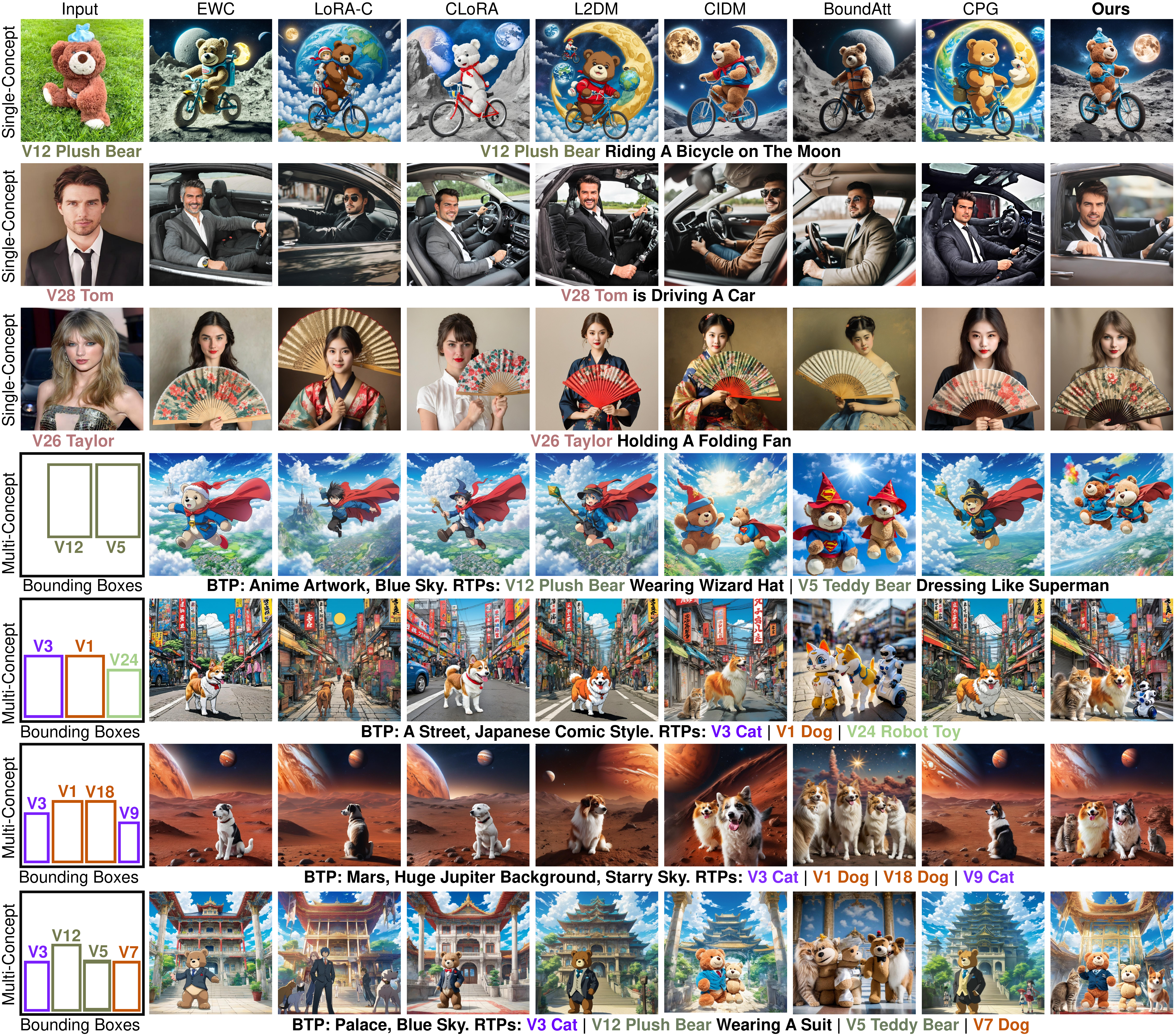}
\vspace{-25pt}
\caption{Qualitative comparisons of single- and multi-concept text-to-image customization generated by SDXL \cite{podell2024sdxl} under the CIVC setting.  }
\label{fig: comparison_single_multi_SDXL}
\vspace{-1pt}
\end{figure*}

\begin{figure*}[t]
\centering
\includegraphics[width = 1.0\linewidth]
{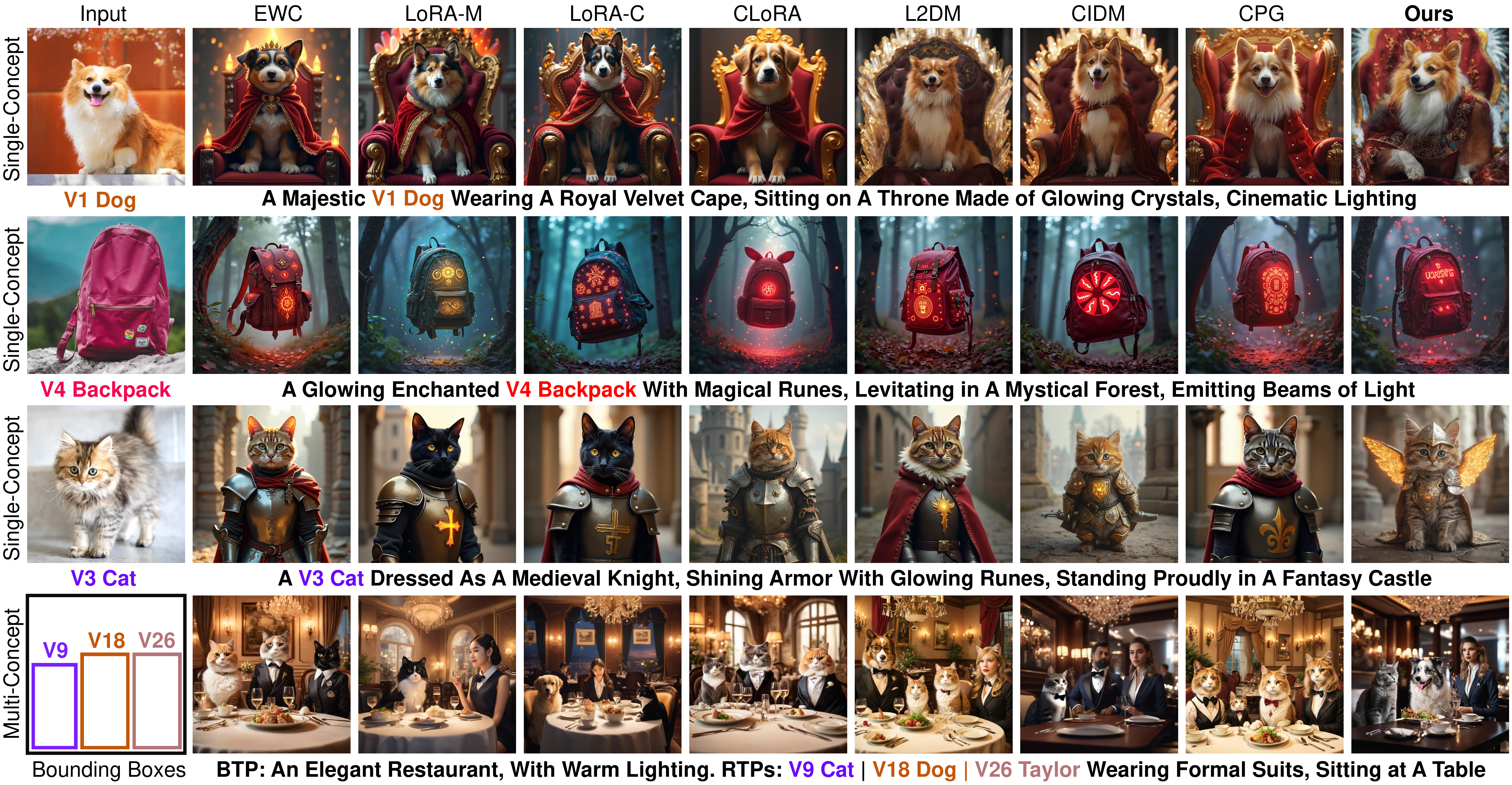}
\vspace{-25pt}
\caption{Qualitative comparisons of single- and multi-concept text-to-image customization generated by FLUX.1 \cite{flux2024} under the CIVC setting.  }
\label{fig: comparison_single_multi_FLUX}
\vspace{-3mm}
\end{figure*}

\begin{figure}[t]
\centering
\includegraphics[width = 1.0\linewidth]
{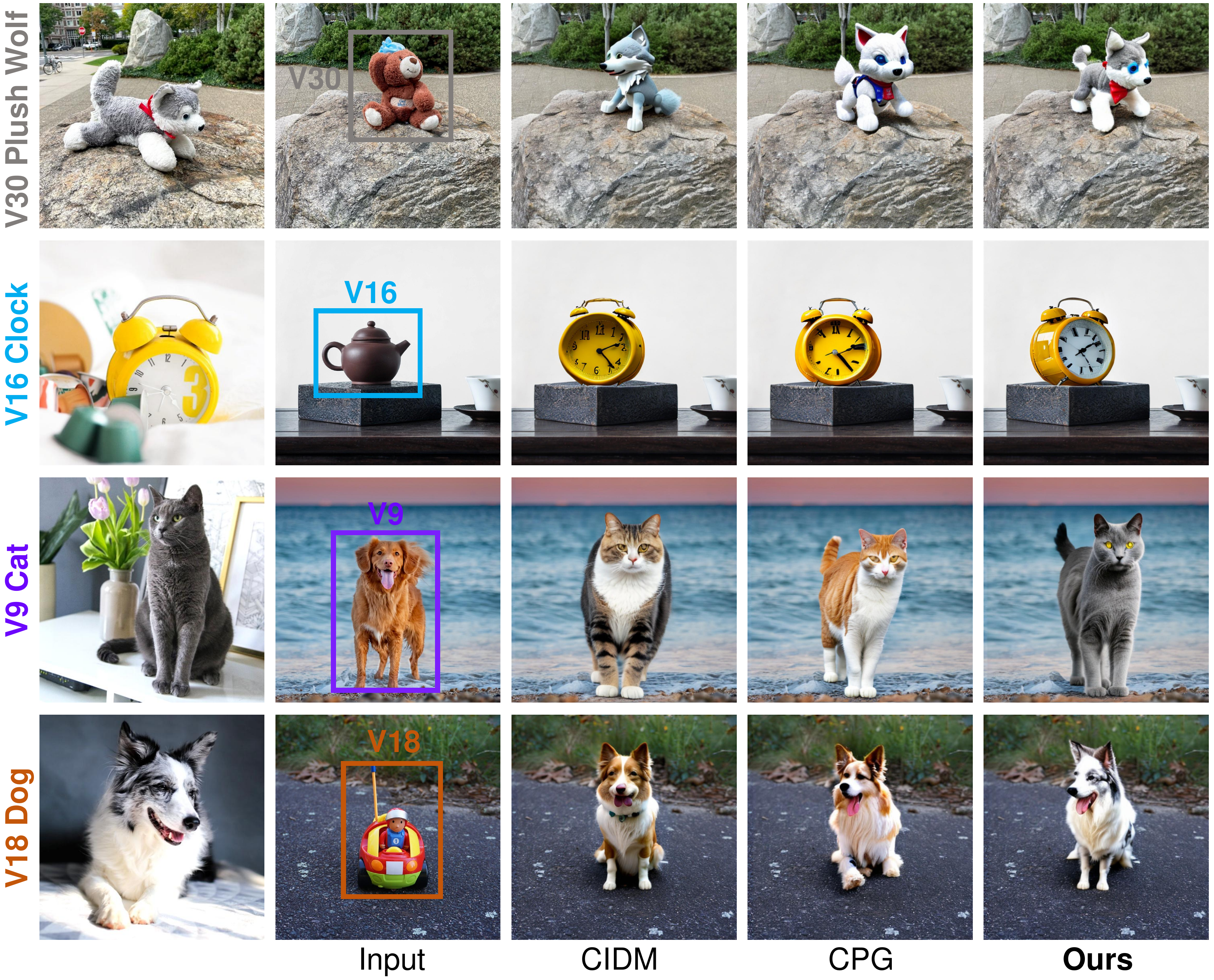}
\vspace{-25pt}
\caption{Qualitative comparisons of editing-based text-to-image customization results generated by SDXL \cite{podell2024sdxl} under the CIVC setting.  }
\label{fig: comparison_editing_SDXL} 
\vspace{-3mm}
\end{figure}

\begin{figure}[t]
\centering
\includegraphics[width = 1.0\linewidth]
{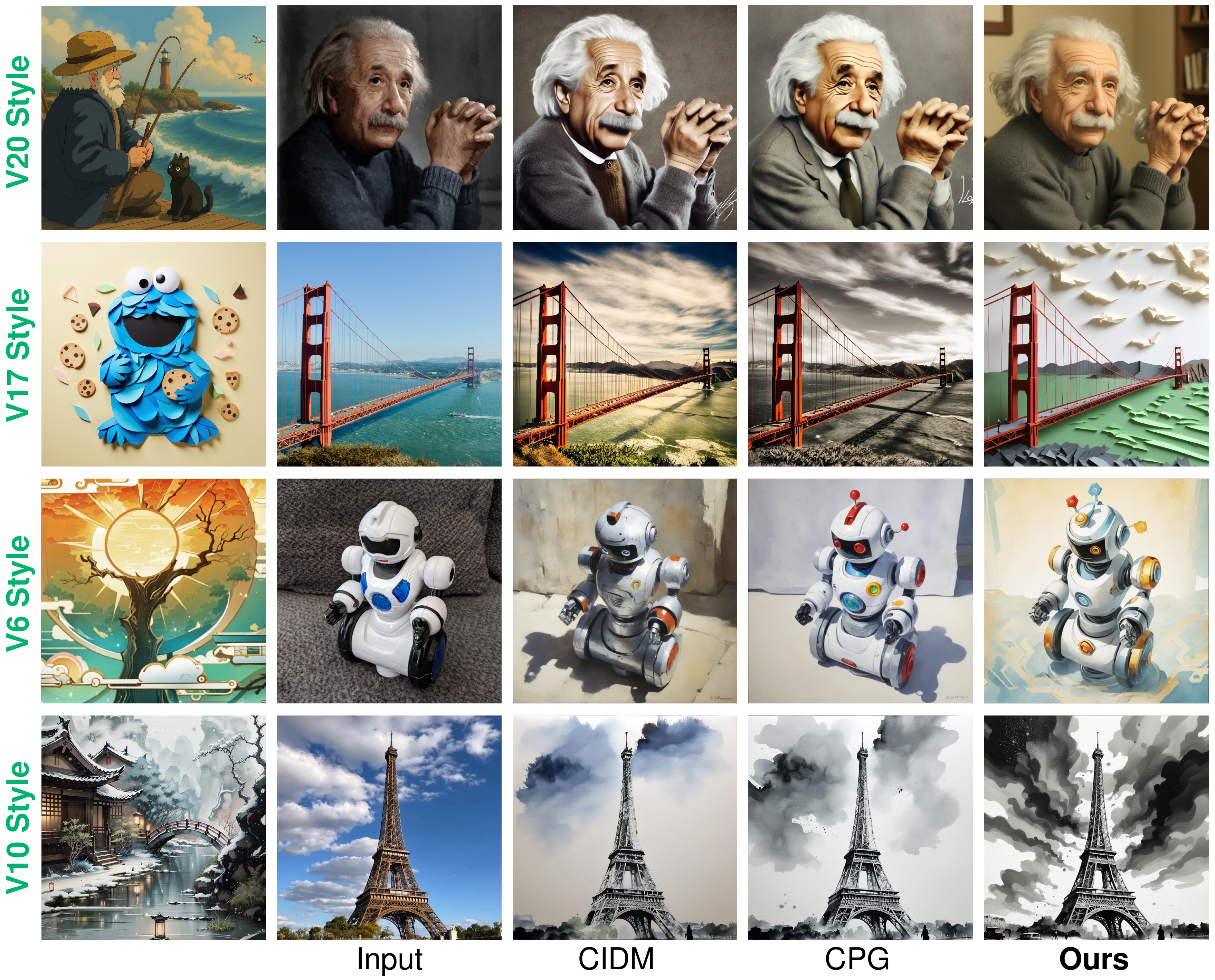}
\vspace{-25pt}
\caption{Qualitative comparisons of style-transfer text-to-image customization results generated by SDXL \cite{podell2024sdxl} under the CIVC setting.  }
\label{fig: comparison_style_SDXL}
\vspace{-3mm}
\end{figure}

\begin{figure*}[t]
\centering
\includegraphics[width = 1.0\linewidth]
{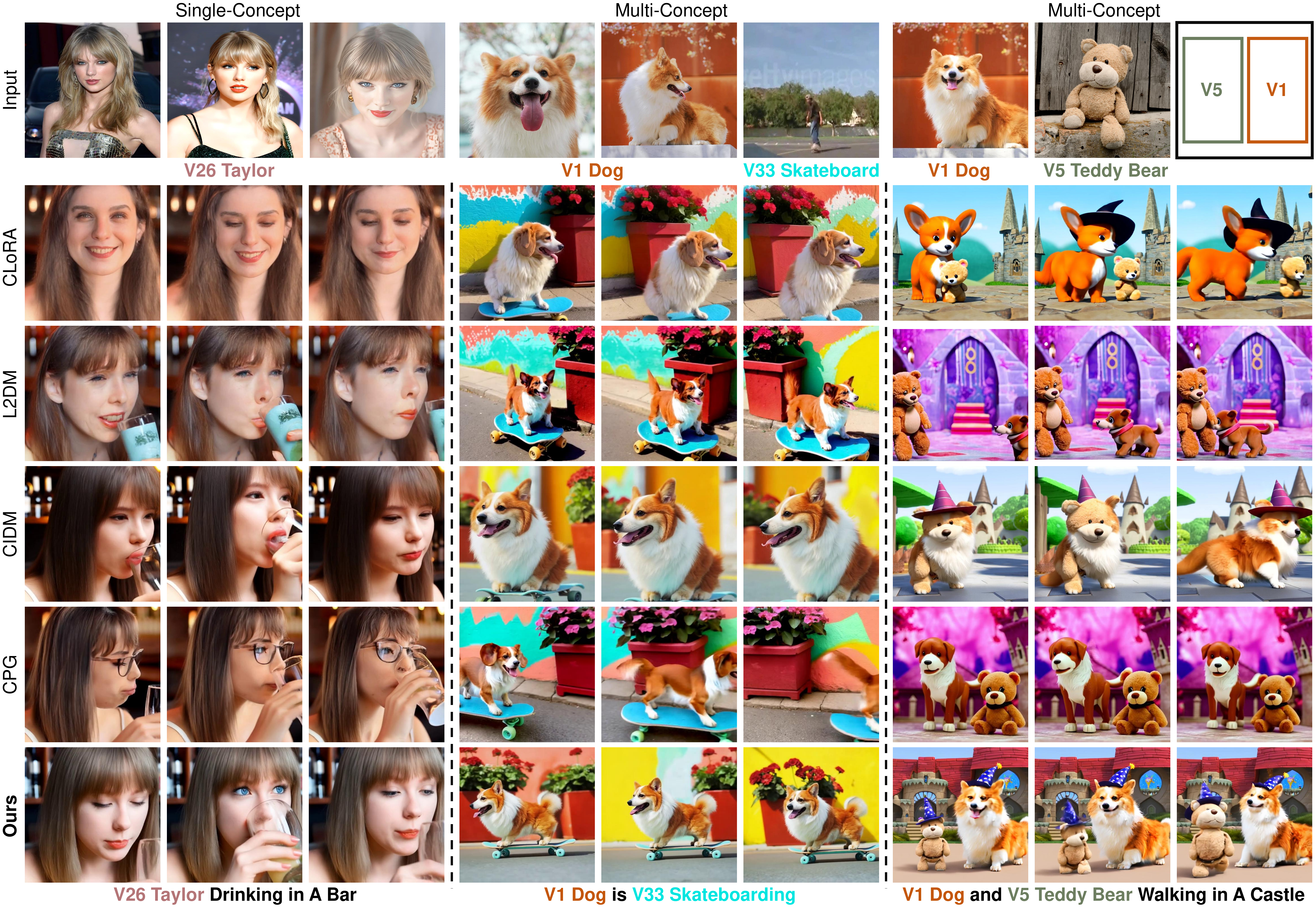}
\vspace{-25pt}
\caption{Qualitative comparisons of single- and multi-concept text-to-video customization under the CIVC setting.  }
\label{fig: comparison_single_multi_video}
\vspace{-5pt}
\end{figure*}

\begin{figure}[t]
\centering
\includegraphics[width = 1.0\linewidth]
{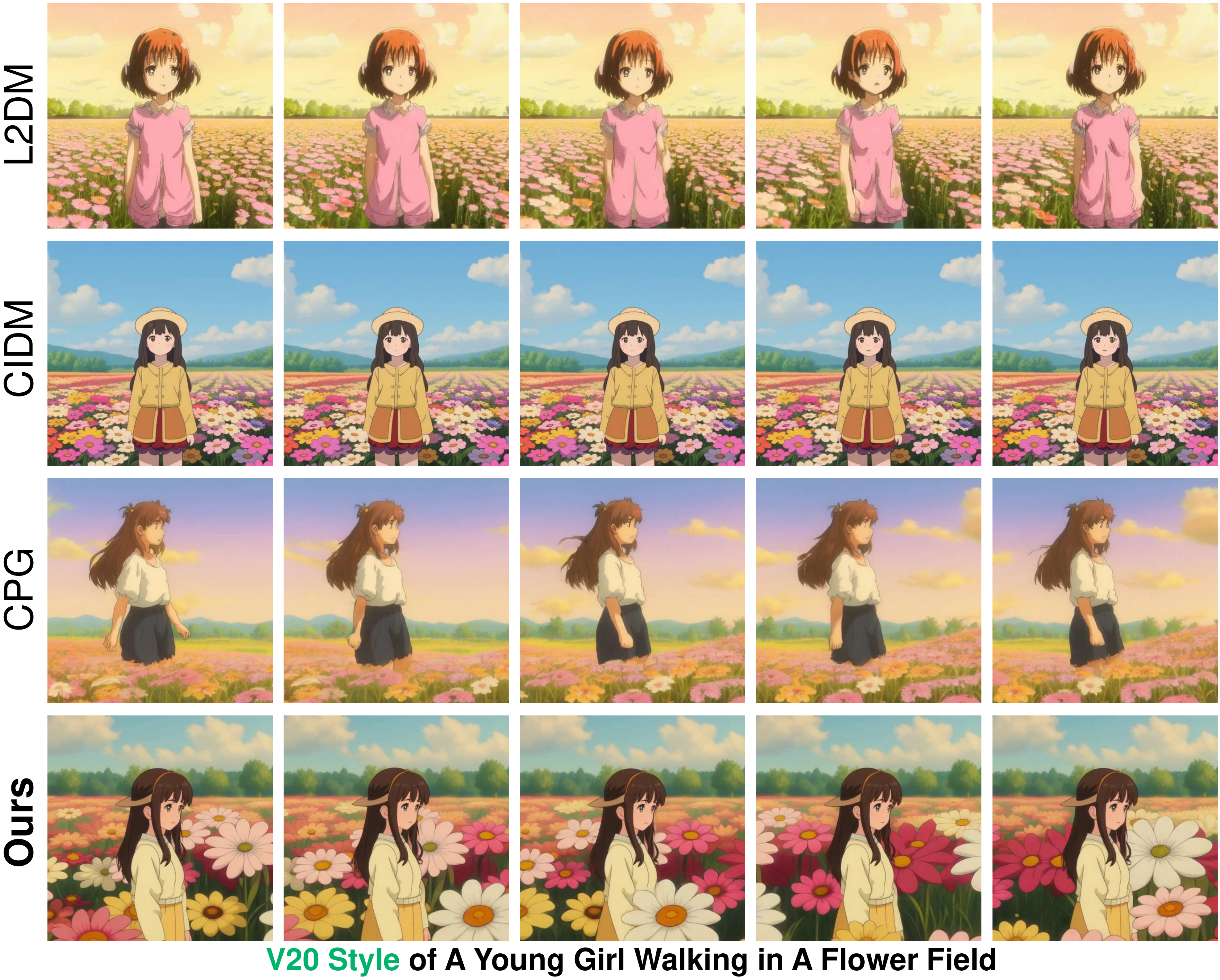}
\vspace{-25pt}
\caption{Qualitative comparisons of style-transfer text-to-video customization results under the CIVC setting. }
\label{fig: comparison_style_video}
\end{figure}

\begin{figure}[t]
\centering
\includegraphics[width = 1.0\linewidth]
{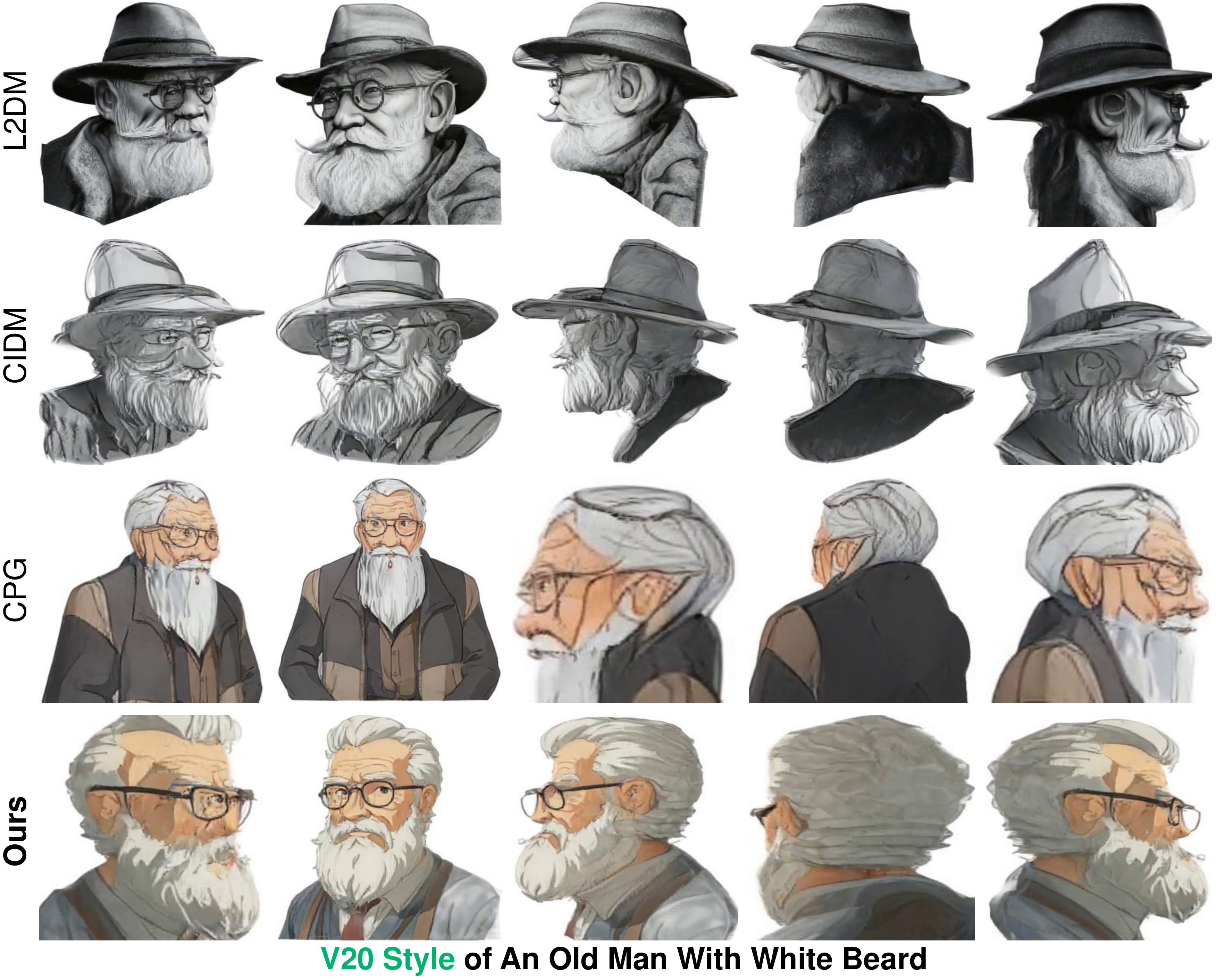}
\vspace{-25pt}
\caption{Qualitative comparisons of style-transfer text-to-3D customization results under the CIVC setting.  }
\label{fig: comparison_style_3D}
\end{figure}

\begin{figure*}[t]
\centering
\includegraphics[width = 1.0\linewidth]
{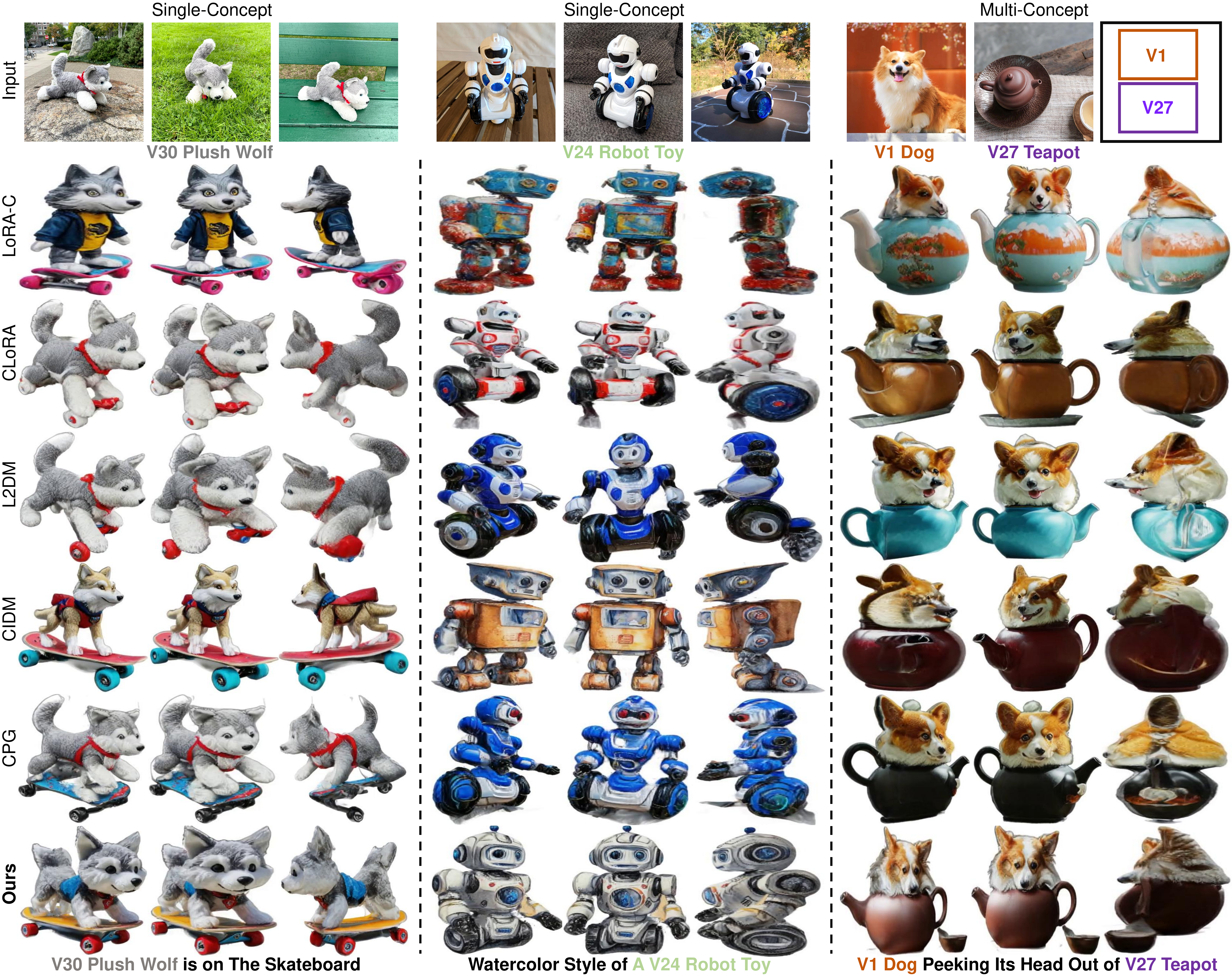}
\vspace{-25pt}
\caption{Qualitative comparisons of single- and multi-concept text-to-3D customization under the CIVC setting.  }
\label{fig: comparison_single_multi_3D}
\vspace{-1pt}
\end{figure*}

\textbf{Evaluation Metrics:}
In this paper, we perform both qualitative and quantitative assessments on a variety of concept customization tasks, including single/multi-concept synthesis, editing, and style transfer across text-to-image, text-to-video, and text-to-3D domains \cite{Raj_2023_ICCV}. 
\textbf{1) Single-Concept Quantitative Evaluation:}
We adopt image alignment (IA), text alignment (TA), forgetting of IA (FIA), and forgetting of TA (FTA) as evaluation metrics for text-to-image generation. Here, $\mathrm{FIA} =
\frac{1}{U-1}\sum_{i=1}^{U-1}\left(\mathrm{IA}_{i,i}-\mathrm{IA}_{i,U}\right)$, where $\mathrm{IA}_{i,i}$ and $\mathrm{IA}_{i,U}$ denote the IA of the $i$-th task evaluated after completing the learning of the $i$-th and $U$-th tasks, respectively. $\mathrm{FTA} = 
\frac{1}{U-1}\sum_{i=1}^{U-1}\left(\mathrm{TA}_{i,i}-\mathrm{TA}_{i,U}\right)$, where $\mathrm{TA}_{i,i}$ and $\mathrm{TA}_{i,U}$ represent the TA of the $i$-th task measured after learning the $i$-th and $U$-th tasks, respectively. Besides, each concept is tested using 20 diverse prompts, each accompanied by a consistent negative prompt, with 25 samples generated per prompt. This results in 500 images per concept for evaluating IA, TA, FIA, and FTA. 
For text-to-video customization, inspired by \cite{10656166}, we introduce an additional metric, DINOI (DI), alongside IA, TA, FIA, and FTA. For evaluation, we construct 10 distinct prompts for each concept and produce 10 video samples per prompt, yielding an aggregate of 100 generated videos per concept. 
To assess the performance on text-to-3D concept customization, we employ 20 diverse prompts for each learned concept, generating 25 samples per prompt, resulting in a total of 500 synthesized 3D assets per concept. Following \cite{Raj_2023_ICCV, poole2023dreamfusion}, we calculate the scores of IA, TA, FIA, and FTA using 40 uniformly distributed azimuth renderings at a fixed elevation angle of 40 degrees. These diverse metrics enable rigorous evaluation across various customization tasks in CIVC. 
\textbf{2) Multi-Concept Quantitative Evaluation:} To assess the generalization of multi-concept synthesis, we conduct quantitative evaluation on two- and three-concept generation. Specifically, we randomly sample 10 concept pairs and 10 concept triplets to generate 20 concept combinations. Each combination is associated with 10 different prompts and corresponding bounding boxes, resulting in 200 distinct settings. For each setting, we generate 25 samples, yielding a total of 5,000 samples. Then, we crop the generated multi-concept samples into single-concept samples based on their bounding boxes, and evaluate them using the single-concept quantitative evaluation protocol.

\subsection{Quantitative Comparisons}
As shown in Tabs.~\ref{tab: quantitative_comp_T2I}--\ref{tab: quantitative_comp_multi_concept_image}, this subsection evaluates the quantitative performance of our model against state-of-the-art approaches for single-concept and multi-concept generation across the text-to-image, text-to-video, and text-to-3D domains, with results averaged over three random runs. 
Compared with the previous conference version (\emph{i.e.}, CIDM \cite{NEURIPS2024_DongJH}), the proposed model not only achieves superior quantitative results on text-to-image, text-to-video, and text-to-3D concept customization tasks, but also requires less than 66\% of the trainable parameters used by CIDM \cite{NEURIPS2024_DongJH}, thereby reducing computational complexity. Furthermore, the quantitative comparisons in Tabs.~\ref{tab: quantitative_comp_T2I}--\ref{tab: quantitative_comp_multi_concept_image} demonstrate that our model significantly surpasses other baselines across all evaluation metrics for text-to-image, text-to-video, and text-to-3D customization. These findings confirm that our CCDM effectively maintains the unique identities of learned concepts, thereby addressing the CIVC problem. Additionally, our model exhibits superior resistance to catastrophic forgetting and concept neglect compared to existing baselines. This can be attributed to our relevance-guided AD-LoRA aggregation, which merges low-rank updates from new and previous tasks to exploit positive inter-task relationships, as well as the proposed controllable regional context synthesis, which ensures the semantic independence of distinct bounding boxes and facilitates smooth transitions at their boundaries.

\begin{table}[t]
\centering
\footnotesize
\setlength{\tabcolsep}{0.75mm}
\renewcommand{\arraystretch}{1.04}
\caption{Quantitative ablation studies of single-concept text-to-image and text-to-3D customization under the CIVC setting, both based on the SDXL backbone \cite{podell2024sdxl}.  }  
\vspace{-3mm}
\resizebox{\linewidth}{!}{
\begin{tabular}{c|c|ccccc}
\toprule
& \multirow{2}{*}{~Variants~} & \multirow{2}{*}{Base} & \multirow{2}{*}{Base+LCT} & \multirow{2}{*}{\makecell{Base+LCT \\ ~~~\quad+FGA}} & \multirow{2}{*}{\makecell{Base+LCT+ \\ FGA+MKB}} & \multirow{2}{*}{\textbf{Ours}} \\
& & \\
\midrule
\multirow{4}{*}{\makebox[0.35cm]{\rotatebox{90}{Modules}}}
& LCT & \xmarkg & \cmark & \cmark & \cmark & \cmark  \\
& FGA & \xmarkg & \xmarkg & \cmark & \cmark & \cmark  \\
& MKB & \xmarkg & \xmarkg & \xmarkg & \cmark & \cmark \\
& RGA & \xmarkg & \xmarkg & \xmarkg & \xmarkg & \cmark \\
\midrule
\multirow{5}{*}{\rotatebox{90}{Text-to-Image}}
& \#Params & 6.29M & 6.43M & 4.72M & \textcolor{deepred}{\textbf{4.23M}} & \textcolor{deepred}{\textbf{4.23M}} \\

& IA ($\uparrow$) & 64.89$\pm$0.05 & 65.18$\pm$0.08 & 65.76$\pm$0.03 & 66.25$\pm$0.05 &\textcolor{deepred}{\textbf{72.84$\pm$0.11}} \\
& TA ($\uparrow$) & 80.28$\pm$0.07 & 80.05$\pm$0.02 & 80.15$\pm$0.07 & 80.20$\pm$0.04 &\textcolor{deepred}{\textbf{80.75$\pm$0.05}} \\
& FIA ($\downarrow$) &\ \ 9.44$\pm$0.07 & \ \ 8.36$\pm$0.11 & \ \ 8.06$\pm$0.09 & \ \ 7.18$\pm$0.03 &\ \ \textcolor{deepred}{\textbf{3.84$\pm$0.03}} \\
& FTA ($\downarrow$) &\ \ 2.28$\pm$0.04 &\ \ 2.32$\pm$0.02 &\ \ 2.01$\pm$0.01 &\ \ 2.08$\pm$0.03 &\ \ \textcolor{deepred}{\textbf{1.89$\pm$0.02}} \\
\midrule

\multirow{5}{*}{\rotatebox{90}{Text-to-3D}}
& \#Params & 6.29M &6.43M & 4.72M& \textcolor{deepred}{\textbf{4.23M}} & \textcolor{deepred}{\textbf{4.23M}} \\
& IA ($\uparrow$) & 65.43$\pm$0.16 & 65.78$\pm$0.12 &66.40$\pm$0.09 &67.21$\pm$0.15  & \textcolor{deepred}{\textbf{69.03$\pm$0.12}} \\
& TA ($\uparrow$) & 70.05$\pm$0.08 &70.54$\pm$0.06 & 70.39$\pm$0.05 & 70.88$\pm$0.06 & \textcolor{deepred}{\textbf{71.25$\pm$0.03}} \\
& FIA ($\downarrow$) &\ \ 8.04$\pm$0.08 &\ \ 7.54$\pm$0.05 &\ \ 7.05$\pm$0.07  & \ \ 5.54$\pm$0.05 &\ \ \textcolor{deepred}{\textbf{2.89$\pm$0.03}} \\
& FTA ($\downarrow$) &\ \ 2.38$\pm$0.06 & \ \ 2.40$\pm$0.04 & \ \ 2.25$\pm$0.04 & \ \ 1.88$\pm$0.06 &\ \ \textcolor{deepred}{\textbf{1.49$\pm$0.04}} \\

\bottomrule
\end{tabular}
}
\label{tab: ablation_studies_single_T2I_T23D}
\end{table}

\subsection{Qualitative Comparisons}
To illustrate our model's efficacy in CIVC, we present qualitative comparisons (Figs.~\ref{fig: comparison_single_multi_SDXL}--\ref{fig: comparison_single_multi_3D}) across various tasks, including single/multi-concept synthesis, editing, and style transfer in text-to-image, text-to-video, and text-to-3D domains.

\begin{figure}[t]
\centering
\includegraphics[width = 1.0\linewidth]
{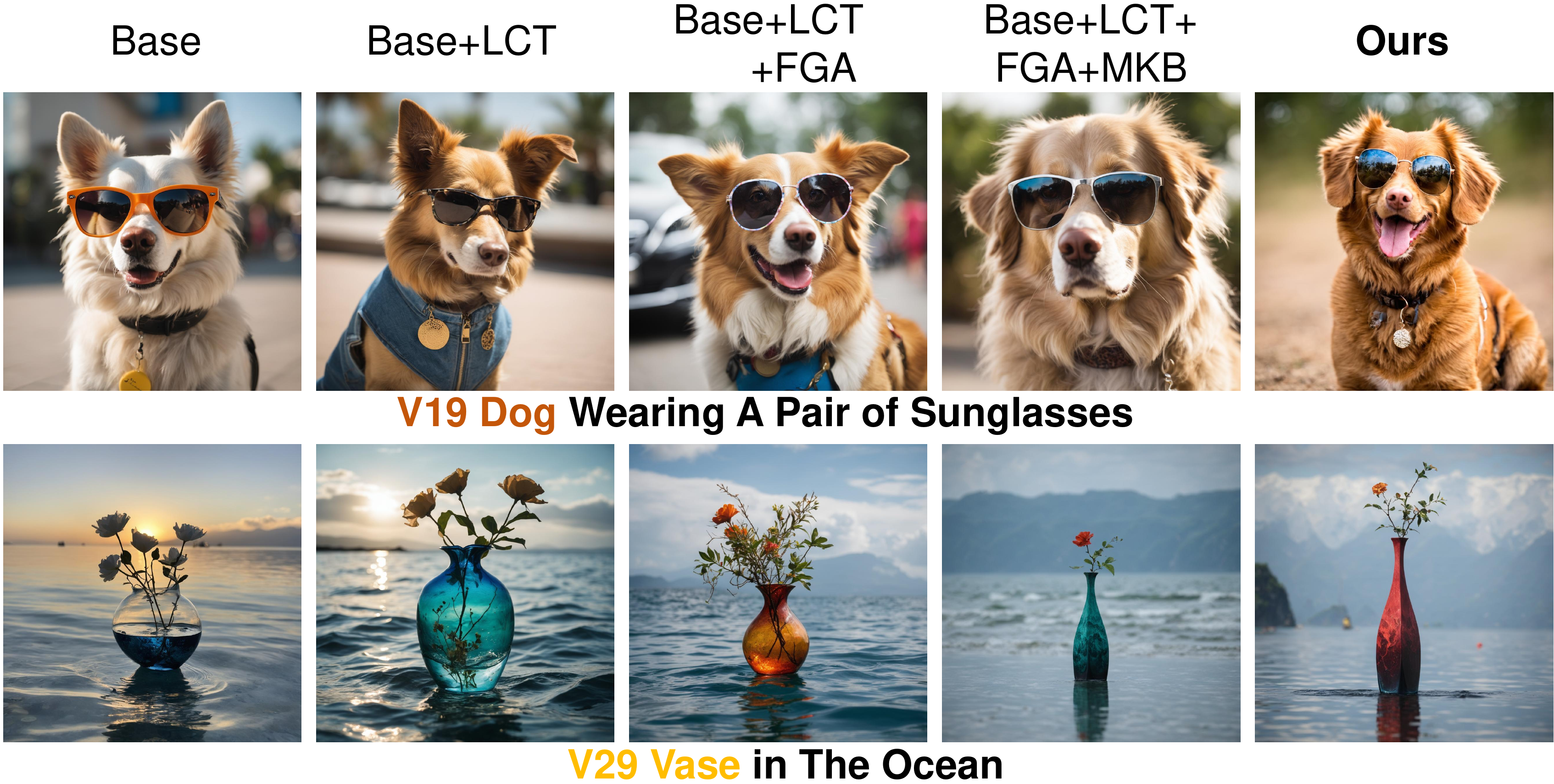}
\vspace{-25pt}
\caption{Qualitative ablation studies of single-concept text-to-image customization results generated by SDXL \cite{podell2024sdxl}.  }
\label{fig: ablation_single_SDXL}
\end{figure}

\textbf{Text-to-Image:} 
In Figs.~\ref{fig: comparison_single_multi_SDXL}--\ref{fig: comparison_single_multi_FLUX}, compared with the conference method CIDM \cite{NEURIPS2024_DongJH}, our model reveals superior capability in maintaining the identities of previous concepts for single/multi-concept image generation. This is attributed to the proposed attribute-decoupled LoRA (AD-LoRA), which selectively preserves concept-unique characteristics to mitigate forgetting. The qualitative comparisons in Figs.~\ref{fig: comparison_single_multi_SDXL}--\ref{fig: comparison_single_multi_FLUX} show that all baselines fail to properly synthesize multiple concepts simultaneously, whereas our model can overcome concept neglect through the proposed controllable regional context synthesis module.  
Besides, AnyDoor \cite{chen2023anydoor} is employed as a plug-in for all competing baselines to perform custom image editing. The visual results in Fig.~\ref{fig: comparison_editing_SDXL} confirm our model's superior editing capabilities, which are attributable to the relevance-guided AD-LoRA aggregation that preserves each concept's distinctive attributes in CIVC.  
Fig.~\ref{fig: comparison_style_SDXL} illustrates that our model has the best style conversion performance by preserving style identities under the CIVC setting.

\begin{figure*}[t]
\centering
\includegraphics[width = 1.0\linewidth]
{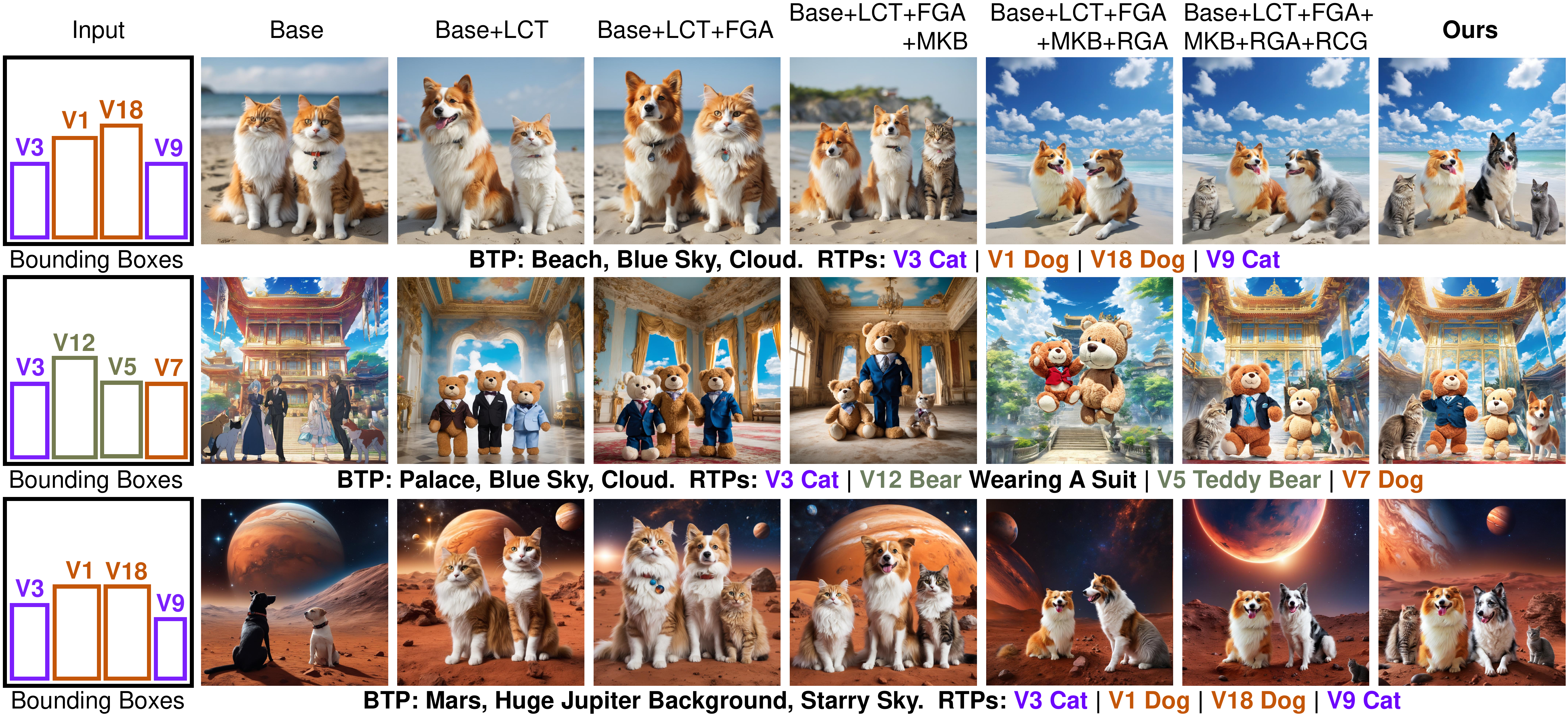}
\vspace{-25pt}
\caption{Qualitative ablation studies of multi-concept text-to-image customization results generated by SDXL \cite{podell2024sdxl}.  }
\label{fig: ablation_multi_SDXL}
\end{figure*}

\textbf{Text-to-Video:}
When applying the proposed CCDM to text-to-video concept customization under the CIVC setting, as shown in Figs.~\ref{fig: comparison_single_multi_video}--\ref{fig: comparison_style_video}, our model significantly outperforms the conference approach (CIDM \cite{NEURIPS2024_DongJH}) and other competing baselines in single/multi-concept video synthesis, as well as in video style transfer tasks. Notably, for fair comparisons in Fig.~\ref{fig: comparison_single_multi_video}, all comparative methods utilize regionally controllable sampling \cite{gu2023mixofshow} in each frame to achieve multi-concept video generation. 
Such performance improvement is attributed to the proposed relevance-guided AD-LoRA aggregation and controllable regional context synthesis, as they fully exploit beneficial inter-task relationships to continually learn new customization tasks while mitigating the challenge of concept neglect within user-specified regions.

\begin{figure}[t]
\centering
\includegraphics[width = 1.0\linewidth]
{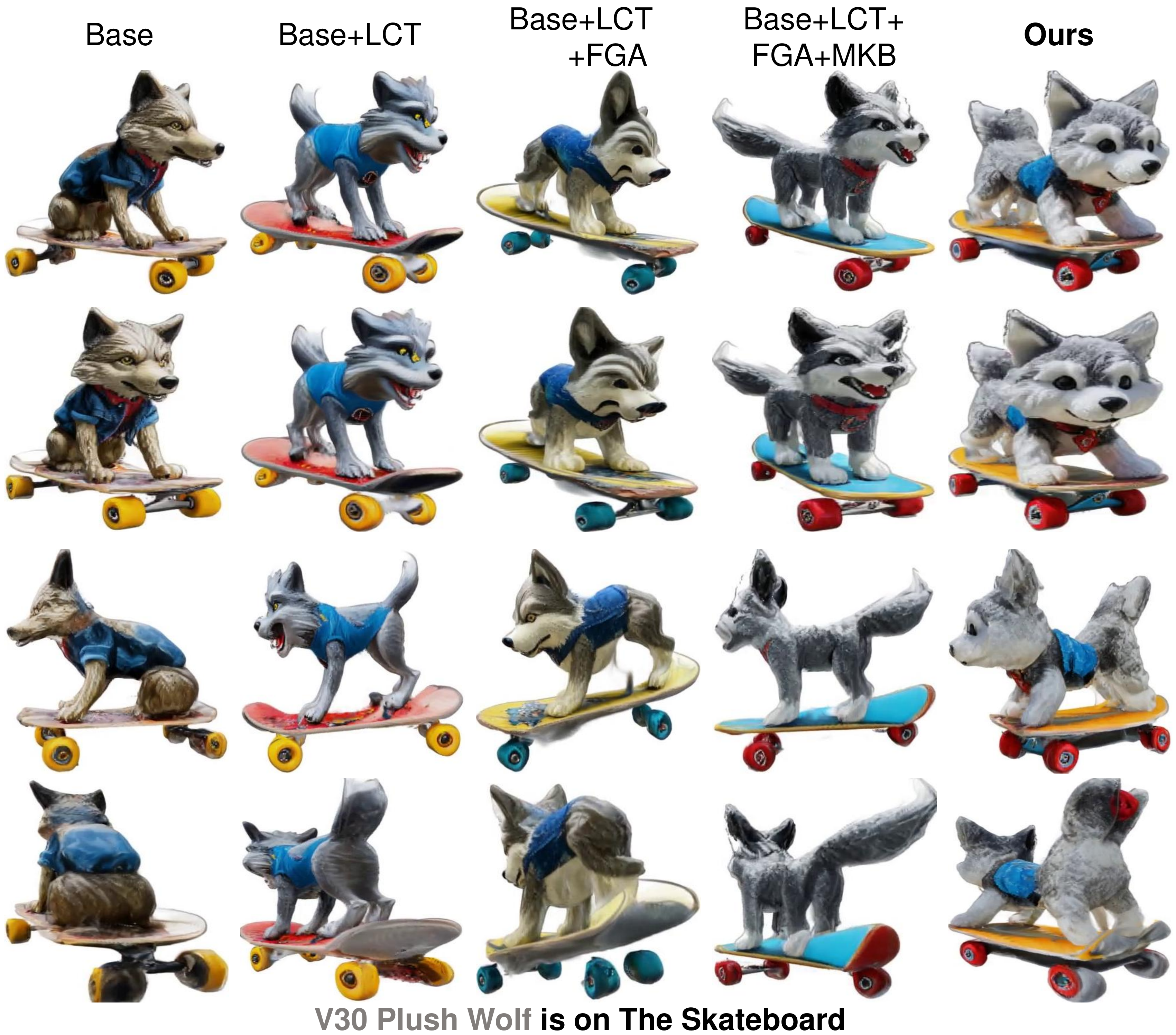}
\vspace{-25pt}
\caption{Ablation studies of single-concept text-to-3D customization.  }
\label{fig: ablation_single_3D}
\end{figure}

\begin{figure}[t]
\centering
\includegraphics[width = 1.0\linewidth]
{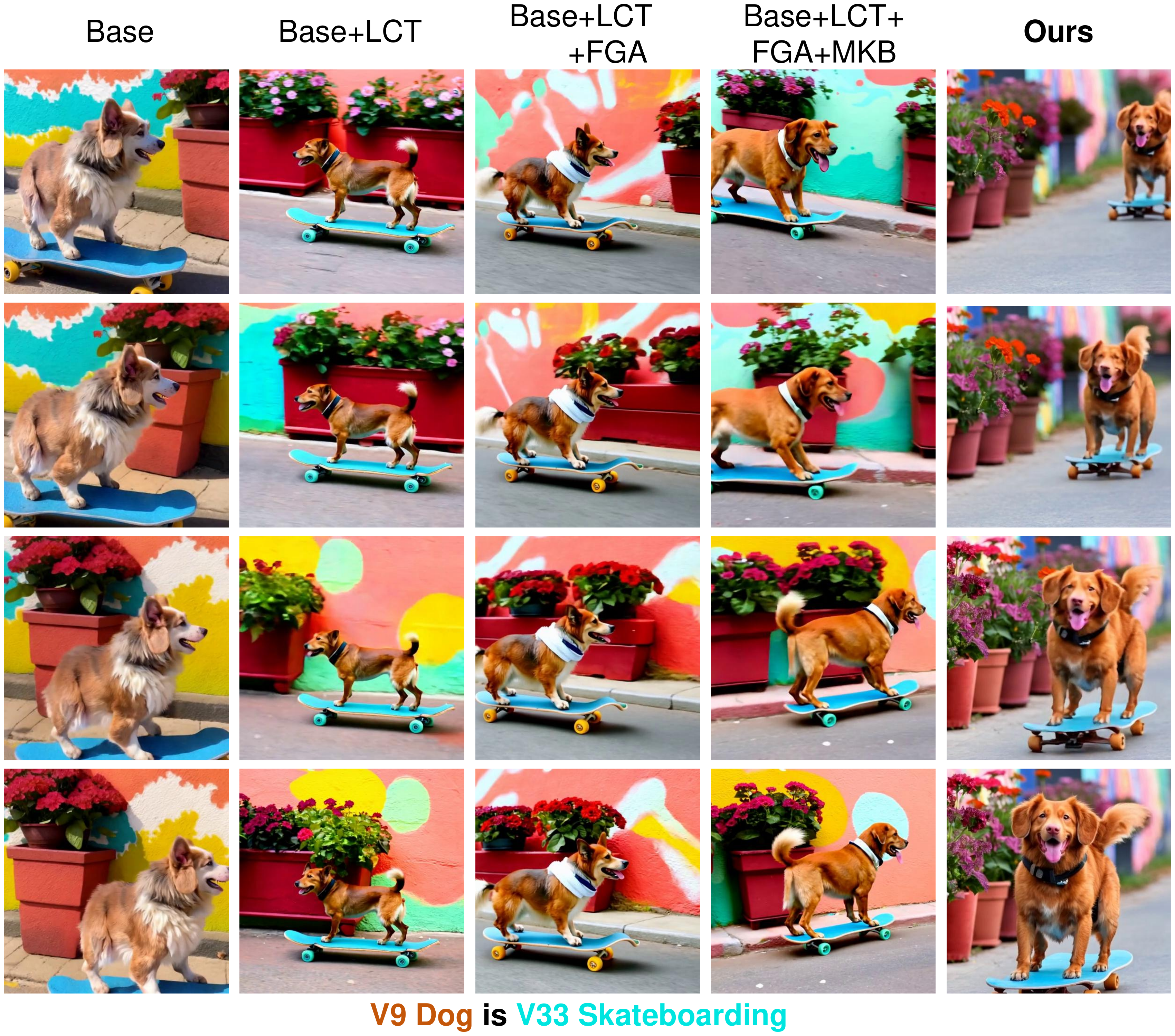}
\vspace{-25pt}
\caption{Ablation studies of multi-concept text-to-video customization.  }
\label{fig: ablation_multi_video}
\end{figure}

\textbf{Text-to-3D:}
Figs.~\ref{fig: comparison_style_3D}--\ref{fig: comparison_single_multi_3D} present qualitative comparisons on text-to-3D customization tasks, including style transfer, single- and multi-concept 3D generation. Compared with other state-of-the-art baselines, our model achieves superior customization performance in synthesizing both single and multiple 3D objects under the CIVC setting. As presented in Figs.~\ref{fig: comparison_style_3D}--\ref{fig: comparison_single_multi_3D}, the comparative methods overlook the unique attributes of previously learned concepts, thereby leading to catastrophic forgetting of old concepts. In contrast, our model preserves distinctive concept characteristics by incorporating the attribute-decoupled LoRA (AD-LoRA) and relevance-guided AD-LoRA aggregation to address the CIVC problem.

\begin{figure*}[t]
\centering
\includegraphics[width = 1.0\linewidth]
{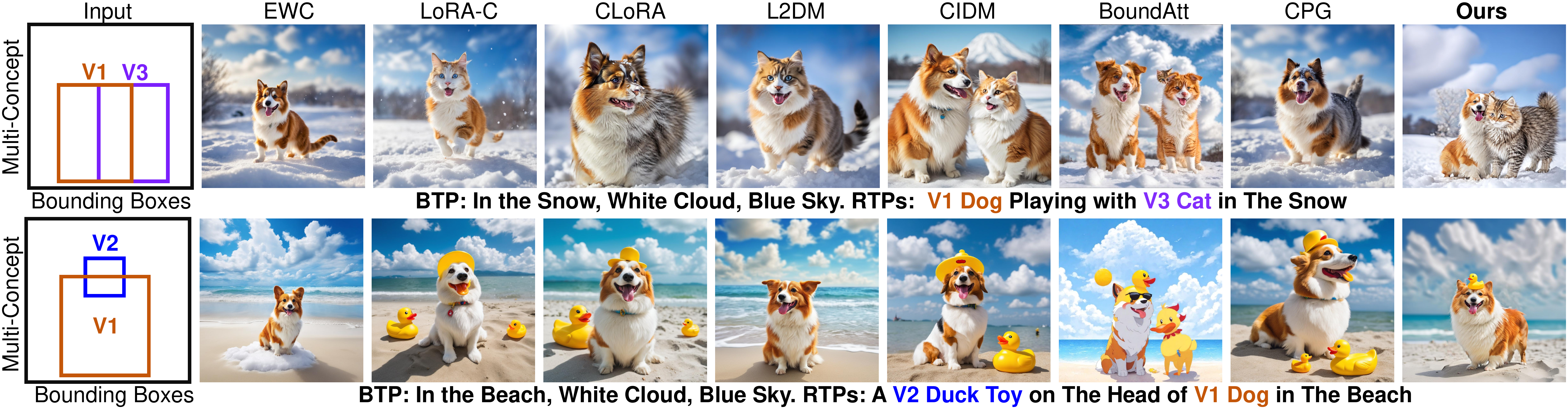}
\vspace{-25pt}
\caption{Qualitative results of multi-concept text-to-image customization generated by SDXL \cite{podell2024sdxl} when provided with overlapping bounding boxes.  }
\label{fig: analysis_overlapping_BBox}
\end{figure*}

\subsection{Ablation Studies}
This subsection examines the efficacy of individual components within the proposed CCDM in addressing the CIVC problem, including the layer-wise concept token (LCT), attribute-decoupled LoRA (AD-LoRA), relevance-guided AD-LoRA aggregation (RGA), and controllable regional context synthesis (CRS) modules. To better verify the effectiveness of the AD-LoRA module, we perform ablation studies on $\{\mathbf{A}_u^l\}_{l=1}^L$ with parameters fixed in the even layers and initialized from a standard Gaussian distribution (denoted  as FGA), as well as on the masked $\{\mathbf{B}_u^l\}_{l=1}^L$ (denoted as MKB). ``Base'' represents the performance without using the above modules. 
\textbf{1) Single-Concept:} Tab.~\ref{tab: ablation_studies_single_T2I_T23D} introduces quantitative ablation experiments of single-concept customization across text-to-image and text-to-3D domains, while Figs.~\ref{fig: ablation_single_SDXL} and \ref{fig: ablation_single_3D} provide qualitative ablation analyses of single-concept text-to-image and text-to-3D customization. The incorporation of the LCT, FGA, MKB, and RGA modules leads to significant performance gains across all evaluation metrics compared to ``Base''. The AD-LoRA module reduces the number of trainable model parameters by approximately 35\%.
The results in Figs.~\ref{fig: ablation_single_SDXL} and \ref{fig: ablation_single_3D} also validate our CCDM's effectiveness in tackling the issue of catastrophic forgetting. 
\textbf{2) Multi-Concept:} For the ablation studies of multi-concept customization, we further analyze the efficacy of regional concept generation (RCG) and noise estimation (NE) in the controllable regional context synthesis (CRS) module. As presented in Figs.~\ref{fig: ablation_multi_SDXL} and \ref{fig: ablation_multi_video}, the qualitative ablation results on multi-concept text-to-image and text-to-video customization demonstrate the superiority of incorporating all proposed modules in addressing concept neglect and catastrophic forgetting under the CIVC setting.

\begin{table}[t]
\centering
\footnotesize
\setlength{\tabcolsep}{0.75mm}
\renewcommand{\arraystretch}{1.04}
\caption{Analysis of different AD-LoRA variants on single-concept text-to-image customization, where the backbone is SDXL \cite{podell2024sdxl}.  }  
\vspace{-3mm}
\resizebox{0.9\linewidth}{!}{
\begin{tabular}{c|cccc}
\toprule
~Variants~ & ~~~~NFA~~~~ & ~~~~RFA~~~~ & ~~Ours-Odd~~ & ~~\textbf{Ours}~~ \\
\midrule
IA ($\uparrow$) & 72.23$\pm$0.14 & 71.20$\pm$0.08 & 72.68$\pm$0.13 & \textcolor{deepred}{\textbf{72.84$\pm$0.11}} \\
FIA ($\downarrow$) &\ \ 4.28$\pm$0.07 &\ \  4.45$\pm$0.06 &\ \  3.96$\pm$0.02 &\ \ \textcolor{deepred}{\textbf{3.84$\pm$0.03}}\\

\bottomrule
\end{tabular}
}
\label{tab: analysis_AD_LoRA_variants}
\end{table}

\subsection{Analysis of Overlapping Bounding Boxes}
This subsection analyzes the efficacy of our CCDM in handling inter-object physical interactions during multi-concept generation with overlapping bounding boxes. As presented in Fig.~\ref{fig: analysis_overlapping_BBox}, in contrast to existing competing baselines that struggle to jointly generate multiple personalized concepts in a coherent manner, the proposed model effectively addresses the issue of concept neglect in user-specified overlapping bounding boxes by leveraging our controllable regional context synthesis module. These comparative results verify the effectiveness and robustness of our model in addressing  inter-object physical interactions (\emph{i.e.}, overlapping bounding boxes) by enforcing clear semantic separation between different bounding boxes while ensuring seamless blending along their borders. They also demonstrate that the proposed controllable regional context synthesis module can effectively improve the overall coherence and harmony of multi-concept text-to-image compositions, even when the bounding boxes of different personalized concepts overlap.

\subsection{Analysis of AD-LoRA Variants}
As shown in Tab.~\ref{tab: analysis_AD_LoRA_variants}, we analyze the performance of different AD-LoRA variants for fixing $\{\mathbf{A}_u^l\}_{l=1}^L$. Specifically, Ours and Ours-Odd denote the variants that fix the even-numbered and odd-numbered layers of $\{\mathbf{A}_u^l\}_{l=1}^L$, respectively. NFA represents the performance without fixing $\mathbf{A}_u^l$, while RFA randomly fixes 50\% of the layers in $\{\mathbf{A}_u^l\}_{l=1}^L$. As reported in Tab.~\ref{tab: analysis_AD_LoRA_variants}, Ours and Ours-Odd outperform NFA and RFA. These results illustrate that fixing the odd- or even-numbered layers of $\{\mathbf{A}_u^l\}_{l=1}^L$ preserves the orthogonality of AD-LoRA across non-adjacent layers and different tasks, thereby mitigating the negative influence of irrelevant attributes.

\subsection{Analysis of the Projection Matrix $\mathbf{W}_p^l$}
To analyze the effectiveness of the trainable projection matrix $\mathbf{W}_p^l$ in dynamically learning the combination weights for AD-LoRA within the network, we replace $\mathbf{W}_p^l$ with the task similarity measure $\mathcal{M}^l$ to merge AD-LoRA weights of semantically similar tasks during training. This variant is denoted as Ours-w/oWP. As shown in Tab.~\ref{tab: analysis_Wp}, our model outperforms Ours-w/oWP across all evaluation metrics. These results demonstrate the effectiveness of adaptively combining AD-LoRA weights within the network, compared with using textual embeddings to measure task correlations.

\begin{table}[t]
\centering
\footnotesize
\setlength{\tabcolsep}{0.75mm}
\renewcommand{\arraystretch}{1.04}
\caption{Quantitative analysis of the projection matrix $\mathbf{W}_p^l$ for single-concept text-to-image customization using the SDXL backbone \cite{podell2024sdxl}. } 
\vspace{-3mm}
\resizebox{0.6\linewidth}{!}{
\begin{tabular}{c|cc}
\toprule
~Variants~ & Ours-w/oWP & \textbf{Ours} (\textbf{CCDM}) \\
\midrule
IA ($\uparrow$) & 71.49$\pm$0.05 & \textcolor{deepred}{\textbf{72.84$\pm$0.11}} \\
FIA ($\downarrow$) & \ \  4.95$\pm$0.06 & \ \  \textcolor{deepred}{\textbf{3.84$\pm$0.03}} \\

\bottomrule
\end{tabular}
}
\label{tab: analysis_Wp}
\end{table}

\section{Conclusion}
This paper presents a novel Continually Customizable Diffusion Model (CCDM) that effectively mitigates catastrophic forgetting and concept neglect in Concept-Incremental Versatile Customization (CIVC). On one hand, to address the catastrophic forgetting of previously learned concepts, we devise an attribute-decoupled LoRA (AD-LoRA) module and a relevance-guided AD-LoRA aggregation strategy. These two components preserve distinctive concept attributes while leveraging inter-task correlations to incrementally learn new customization tasks. On the other hand, to mitigate concept neglect, we introduce a controllable regional context synthesis module that composes multiple personalized concepts in accordance with user-specified requirements. This strategy enhances the overall quality of multi-concept synthesis by maintaining semantic independence across different regions and ensuring their smooth boundary blending. Through comprehensive evaluations, we validate our CCDM's superiority over existing baselines in diverse CIVC scenarios spanning text-to-image, text-to-video, and text-to-3D domains.
In future work, we will explore more robust memory mechanisms to retain long-term personalized knowledge and incorporate real-time user feedback to guide concept customization.

\ifCLASSOPTIONcaptionsoff
  \newpage
\fi

\bibliographystyle{IEEEtranS}
\bibliography{arxiv}

@article{yang2024cogvideox,
  title={Cogvideox: Text-to-video diffusion models with an expert transformer},
  author={Yang, Zhuoyi and Teng, Jiayan and Zheng, Wendi and Ding, Ming and others},
  journal={arXiv preprint arXiv:2408.06072},
  year={2024}
}

@inproceedings{tang2024lgm,
  title={Lgm: Large multi-view gaussian model for high-resolution 3d content creation},
  author={Tang, Jiaxiang and Chen, Zhaoxi and Chen, Xiaokang and others},
  booktitle={ECCV},
  year={2024},
  organization={Springer}
}

@inproceedings{wen2025ouroboros3d,
  title={Ouroboros3d: Image-to-3d generation via 3d-aware recursive diffusion},
  author={Wen, Hao and Huang, Zehuan and Wang, Yaohui and Chen, Xinyuan and Sheng, Lu},
  booktitle={CVPR},
  pages={21631--21641},
  year={2025},
}

@article{blattmann2023stable,
  title={Stable video diffusion: Scaling latent video diffusion models to large datasets},
  author={Blattmann, Andreas and Dockhorn, Tim and Kulal, Sumith and others},
  journal={arXiv preprint arXiv:2311.15127},
  year={2023}
}

@InProceedings{Lin_2023_CVPR,
    author    = {Lin, Chen-Hsuan and Gao, Jun and Tang, Luming and Takikawa, Towaki and Zeng, Xiaohui and others},
    title     = {Magic3D: High-Resolution Text-to-3D Content Creation},
    booktitle = {CVPR},
    month     = {June},
    year      = {2023},
    pages     = {300-309}
}

@InProceedings{101007978031_72907_23,
author="Liu, Fangfu
and Wang, Hanyang
and Chen, Weiliang
and Sun, Haowen
and Duan, Yueqi",
title="Make-Your-3D: Fast and Consistent Subject-Driven 3D Content Generation",
booktitle="ECCV",
year="2024",
pages="389--406",
}

@InProceedings{Petrov_2025_CVPR,
    author    = {Petrov, Dmitry and Goyal, Pradyumn and Shivashok, Divyansh and Tao, Yuanming and Averkiou, Melinos and Kalogerakis, Evangelos},
    title     = {ShapeWords: Guiding Text-to-Image Synthesis with 3D Shape-Aware Prompts},
    booktitle = {CVPR},
    month     = {June},
    year      = {2025},
    pages     = {13305-13314},
}

@inproceedings{chen2024disenstudio,
title={DisenStudio: Customized Multi-subject Text-to-Video Generation with Disentangled Spatial Control},
author={Hong Chen and Xin Wang and Yipeng Zhang and Yuwei Zhou and Zeyang Zhang and Siao Tang and Wenwu Zhu},
booktitle={ACM MM},
year={2024},
}

@article{wu2024customcrafter,
  title={Customcrafter: Customized video generation with preserving motion and concept composition abilities},
  author={Wu, Tao and Zhang, Yong and Wang, Xintao and Zhou, Xianpan and others},
  journal={arXiv preprint arXiv:2408.13239},
  year={2024}
}

@InProceedings{Huang_2025_CVPR,
    author    = {Huang, Chi-Pin and Wu, Yen-Siang and Chung, Hung-Kai and Chang, Kai-Po and Yang, Fu-En and Wang, Yu-Chiang Frank},
    title     = {VideoMage: Multi-Subject and Motion Customization of Text-to-Video Diffusion Models},
    booktitle = {CVPR},
    month     = {June},
    year      = {2025},
    pages     = {17603-17612}
}

@InProceedings{Dong_2022_CVPR,
	author    = {Dong, Jiahua and Wang, Lixu and Fang, Zhen and Sun, Gan and Xu, Shichao and Wang, Xiao and Zhu, Qi},
	title     = {Federated Class-Incremental Learning},
	booktitle = {CVPR},
	month     = {June},
	year      = {2022},
	pages     = {10164-10173}
}

@misc{flux2024,
    author={Black Forest Labs},
    title={FLUX},
    year={2024},
    howpublished={\url{https://github.com/black-forest-labs/flux}},
}

@inproceedings{xie2024sana,
  title={Sana: Efficient High-Resolution Image Synthesis with Linear Diffusion Transformer},
      author={Enze Xie and Junsong Chen and Junyu Chen and Han Cai and Haotian Tang and Yujun Lin and Zhekai Zhang and others},
      year={2024},
  booktitle={ICLR},
}

@article{articleJinUniCanvas,
author = {Jin, Jian and Shen, Yang and Zhao, Xinyang and Fu, Zhenyong and Yang, Jian},
year = {2025},
month = {01},
pages = {3456-3480},
title = {UniCanvas: Affordance-Aware Unified Real Image Editing via Customized Text-to-Image Generation},
volume = {133},
journal = {International Journal of Computer Vision},
}

@article{Wang_Bai_Xie_Yi_Wang_Ma_2025, 
title={SigStyle: Signature Style Transfer via Personalized Text-to-Image Models}, 
volume={39}, 
number={8}, 
journal={AAAI}, 
author={Wang, Ye and Bai, Tongyuan and Xie, Xuping and Yi, Zili and Wang, Yilin and Ma, Rui}, year={2025}, month={Apr.}, 
pages={8051-8059},
}

@article{Choi_Park_Baek_2025, 
title={DynASyn: Multi-Subject Personalization Enabling Dynamic Action Synthesis}, volume={39},number={3}, 
journal={AAAI}, 
author={Choi, Yongjin and Park, Chanhun and Baek, Seung Jun}, 
year={2025}, month={Apr.}, pages={2564-2572} 
}

@inproceedings{rombach2022high,
  author = {Rombach, Robin and Blattmann, Andreas and Lorenz, Dominik and Esser, Patrick and Ommer, Bj{\"o}rn},
  booktitle = {CVPR},
  pages = {10684--10695},
  title = {High-resolution image synthesis with latent diffusion models},
  year = 2022
}

@article{goldstein2018phasemax,
  title={Phasemax: Convex phase retrieval via basis pursuit},
  author={Goldstein, Tom and Studer, Christoph},
  journal={IEEE Transactions on Information Theory},
  volume={64},
  number={4},
  pages={2675--2689},
  year={2018},
  publisher={IEEE}
}

@InProceedings{101007978031_731167_14,
author="Li, Pengzhi
and Nie, Qiang
and Chen, Ying
and Jiang, Xi
and Wu, Kai
and Lin, Yuhuan
and Liu, Yong
and Peng, Jinlong
and Wang, Chengjie
and Zheng, Feng",
title="Tuning-Free Image Customization with Image and Text Guidance",
booktitle="ECCV",
year="2024",
pages="233--250",
}

@inproceedings{li2022blip,
  title={Blip: Bootstrapping language-image pre-training for unified vision-language understanding and generation},
  author={Li, Junnan and Li, Dongxu and Xiong, Caiming and Hoi, Steven},
  booktitle={ICML},
  pages={12888--12900},
  year={2022},
}

@inproceedings{kumari2022customdiffusion,
  title={Multi-Concept Customization of Text-to-Image Diffusion},
  author={Kumari, Nupur and Zhang, Bingliang and Zhang, Richard and Shechtman, Eli and Zhu, Jun-Yan},
  booktitle = {CVPR},
  year      = {2023}
}

@inproceedings{ho2021classifierfree,
title={Classifier-Free Diffusion Guidance},
author={Jonathan Ho and Tim Salimans},
booktitle={NeurIPS 2021 Workshop on Deep Generative Models and Downstream Applications},
year={2021},
}

@InProceedings{Li_2023_CVPR,
    author    = {Li, Yuheng and Liu, Haotian and Wu, Qingyang and Mu, Fangzhou and Yang, Jianwei and Gao, Jianfeng and Li, Chunyuan and Lee, Yong Jae},
    title     = {GLIGEN: Open-Set Grounded Text-to-Image Generation},
    booktitle = {CVPR},
    month     = {June},
    year      = {2023},
    pages     = {22511-22521}
}

@inproceedings{DBLPconfaaaiWuPZPYZLM25,
  author={Feize Wu and Yun Pang and Junyi Zhang and Lianyu Pang and Jian Yin and Baoquan Zhao and Qing Li and Xudong Mao},
  title={CoRe: Context-Regularized Text Embedding Learning for Text-to-Image Personalization},
  year={2025},
  pages={8377-8385},
  booktitle={AAAI},
}

@article{Zhu_Li_Ma_He_Li_2025, 
title={MultiBooth: Towards Generating All Your Concepts in an Image from Text}, volume={39},
number={10}, 
journal={AAAI}, 
author={Zhu, Chenyang and Li, Kai and Ma, Yue and He, Chunming and Li, Xiu}, year={2025}, month={Apr.}, pages={10923-10931} 
}

@inproceedings{podell2024sdxl,
title={{SDXL}: Improving Latent Diffusion Models for High-Resolution Image Synthesis},
author={Dustin Podell and Zion English and Kyle Lacey and Andreas Blattmann and Tim Dockhorn and Jonas M{\"u}ller and Joe Penna and Robin Rombach},
booktitle={ICLR},
year={2024},
}

@article{yang2024loracomposer,
  title={LoRA-Composer: Leveraging Low-Rank Adaptation for Multi-Concept Customization in Training-Free Diffusion Models}, 
  author={Yang Yang and Wen Wang and Liang Peng and Chaotian Song and Yao Chen and Hengjia Li and others},
  year={2024},
 journal={arxiv preprint arxiv:2403.11627},
}

@ARTICLE{10323204,
  author={Dong, Jiahua and Li, Hongliu and Cong, Yang and Sun, Gan and Zhang, Yulun and Van Gool, Luc},
  journal={IEEE Transactions on Pattern Analysis and Machine Intelligence}, 
  title={No One Left Behind: Real-World Federated Class-Incremental Learning}, 
  year={2024},
  volume={46},
  number={4},
  pages={2054-2070},
}

@inproceedings{NEURIPS2024_DongJH,
 author = {Dong, Jiahua and Liang, Wenqi and Li, Hongliu and Zhang, Duzhen and Cao, Meng and Ding, Henghui and Khan, Salman and Khan, Fahad Shahbaz},
 booktitle = {NeurIPS},
 pages = {130057--130083},
 title = {How to Continually Adapt Text-to-Image Diffusion Models for Flexible Customization?},
 volume = {37},
 year = {2024}
}

@article{li2017learning,
  title={Learning without forgetting},
  author={Li, Zhizhong and Hoiem, Derek},
  journal={IEEE Transactions on Pattern Analysis and Machine Intelligence},
  volume={40},
  number={12},
  pages={2935--2947},
  year={2017},
  publisher={IEEE}
}

@article{kirkpatrick2017overcoming,
  title={Overcoming catastrophic forgetting in neural networks},
  author={Kirkpatrick, James and Pascanu, Razvan and Rabinowitz, Neil and others},
  journal={Proceedings of the National Academy of Sciences},
  volume={114},
  number={13},
  pages={3521--3526},
  year={2017},
  publisher={National Acad Sciences}
}

@article{zhong2024multi,
title={Multi-Lo{RA} Composition for Image Generation},
author={Ming Zhong and yelong shen and Shuohang Wang and Yadong Lu and Yizhu Jiao and Siru Ouyang and Donghan Yu and Jiawei Han and Weizhu Chen},
journal={Transactions on Machine Learning Research},
issn={2835-8856},
year={2024},
}

@article{smith2023continual,
  title={Continual diffusion: Continual customization of text-to-image diffusion with c-lora},
  author={Smith, James Seale and Hsu, Yen-Chang and Zhang, Lingyu and Hua, Ting and Kira, Zsolt and Shen, Yilin and Jin, Hongxia},
  journal={Transactions on Machine Learning Research},
  year={2024}
}

@ARTICLE{sun2024create,
  author={Sun, Gan and Liang, Wenqi and Dong, Jiahua and Li, Jun and Ding, Zhengming and Cong, Yang},
  journal={IEEE Transactions on Pattern Analysis and Machine Intelligence}, 
  title={Create Your World: Lifelong Text-to-Image Diffusion}, 
  year={2024},
  volume={46},
  number={9},
  pages={6454-6470},
}

@inproceedings{saharia2022photorealistic,
title={Photorealistic Text-to-Image Diffusion Models with Deep Language Understanding},
author={Chitwan Saharia and William Chan and Saurabh Saxena and Lala Li and Jay Whang and Emily Denton and others},
booktitle={NeurIPS},
year={2022},
}

@InProceedings{Gu_2022_CVPR,
    author    = {Gu, Shuyang and Chen, Dong and Bao, Jianmin and Wen, Fang and Zhang, Bo and Chen, Dongdong and Yuan, Lu and Guo, Baining},
    title     = {Vector Quantized Diffusion Model for Text-to-Image Synthesis},
    booktitle = {CVPR},
    month     = {June},
    year      = {2022},
    pages     = {10696-10706}
}

@inproceedings{liu2023customizable,
title={Customizable Image Synthesis with Multiple Subjects},
author={Zhiheng Liu and Yifei Zhang and Yujun Shen and Kecheng Zheng and Kai Zhu and Ruili Feng and Yu Liu and others},
booktitle={NeurIPS},
year={2023},
}

@inproceedings{10114536805283687625,
author = {Sauer, Axel and Boesel, Frederic and Dockhorn, Tim and Blattmann, Andreas and Esser, Patrick and Rombach, Robin},
title = {Fast High-Resolution Image Synthesis with Latent Adversarial Diffusion Distillation},
year = {2024},
booktitle = {SIGGRAPH Asia 2024 Conference Papers},
}

@inproceedings{hu2022lora,
title={Lo{RA}: Low-Rank Adaptation of Large Language Models},
author={Edward J Hu and yelong shen and Phillip Wallis and Zeyuan Allen-Zhu and Yuanzhi Li and Shean Wang and Lu Wang and Weizhu Chen},
booktitle={ICLR},
year={2022},
}

@InProceedings{Lu_2024_CVPR,
    author    = {Lu, Yanzuo and Zhang, Manlin and Ma, Andy J and Xie, Xiaohua and Lai, Jianhuang},
    title     = {Coarse-to-Fine Latent Diffusion for Pose-Guided Person Image Synthesis},
    booktitle = {CVPR},
    month     = {June},
    year      = {2024},
    pages     = {6420-6429}
}

@inproceedings{1555536920703693369,
author = {Liu, Shih-Yang and Wang, Chien-Yi and Yin, Hongxu and Molchanov, Pavlo and Wang, Yu-Chiang Frank and Cheng, Kwang-Ting and Chen, Min-Hung},
title = {DoRA: weight-decomposed low-rank adaptation},
year = {2024},
booktitle = {ICML},
numpages = {22},
}

@InProceedings{Zhou_2024_CVPR,
    author    = {Zhou, Yufan and Zhang, Ruiyi and Gu, Jiuxiang and Sun, Tong},
    title     = {Customization Assistant for Text-to-Image Generation},
    booktitle = {CVPR},
    month     = {June},
    year      = {2024},
    pages     = {9182-9191}
}

@inproceedings{ruiz2023dreambooth,
  title={Dreambooth: Fine tuning text-to-image diffusion models for subject-driven generation},
  author={Ruiz, Nataniel and Li, Yuanzhen and Jampani, Varun and Pritch, Yael and Rubinstein, Michael and Aberman, Kfir},
  booktitle={CVPR},
  year={2023}
}

@article{Ramesh2022HierarchicalTI,
title={Hierarchical Text-Conditional Image Generation with CLIP Latents},
author={Aditya Ramesh and Prafulla Dhariwal and Alex Nichol and Casey Chu and Mark Chen},
journal={arxiv preprint arxiv:2204.06125},
year={2022},
}

@InProceedings{Zeng_2024_CVPR,
    author    = {Zeng, Yu and Patel, Vishal M. and Wang, Haochen and Huang, Xun and Wang, Ting-Chun and Liu, Ming-Yu and Balaji, Yogesh},
    title     = {JeDi: Joint-Image Diffusion Models for Finetuning-Free Personalized Text-to-Image Generation},
    booktitle = {CVPR},
    month     = {June},
    year      = {2024},
    pages     = {6786-6795}
}

@inproceedings{10555536920703692573,
author = {Esser, Patrick and Kulal, Sumith and Blattmann, Andreas and Entezari, Rahim and M\"{u}ller, Jonas and Saini, Harry and others},
title = {Scaling rectified flow transformers for high-resolution image synthesis},
year = {2024},
booktitle = {ICML},
}

@inproceedings{wang2025msdiffusion,
  title={{MS}-Diffusion: Multi-subject Zero-shot Image Personalization with Layout Guidance},
  author={Xierui Wang and Siming Fu and Qihan Huang and Wanggui He and Hao Jiang},
  booktitle={ICLR},
  year={2025},
}

@INPROCEEDINGS{10377873,
  author={Han, Ligong and Li, Yinxiao and Zhang, Han and Milanfar, Peyman and Metaxas, Dimitris and Yang, Feng},
  booktitle={ICCV}, 
  title={SVDiff: Compact Parameter Space for Diffusion Fine-Tuning}, 
  year={2023},
  volume={},
  number={},
  pages={7289-7300},
}

@inproceedings{10114536646473681391,
author = {Xie, Zhenyu and Dong, Haoye and Gao, Yufei and Ma, Zehua and Liang, Xiaodan},
title = {DreamVTON: Customizing 3D Virtual Try-on with Personalized Diffusion Models},
year = {2024},
booktitle = {ACM MM},
pages = {10784–10793},
}

@article{song2025mult,
      title={MultiDreamer3D: Multi-concept 3D Customization with Concept-Aware Diffusion Guidance}, 
      author={Wooseok Song and Seunggyu Chang and Jaejun Yoo},
      year={2025},
      journal={arXiv preprint arXiv:2501.13449},
}

@INPROCEEDINGS{wu2024motionbooth,
  title={MotionBooth: Motion-Aware Customized Text-to-Video Generation},
  author={Jianzong Wu and Xiangtai Li and Yanhong Zeng and Jiangning Zhang and Qianyu Zhou and Yining Li and Yunhai Tong and Kai Chen},
  booktitle={NeurIPS},
  year={2024},
}

@InProceedings{Zhang_2023_inst,
 author    = {Zhang, Yuxin and Huang, Nisha and Tang, Fan and Huang, Haibin and Ma, Chongyang and others},
 title     = {Inversion-Based Style Transfer With Diffusion Models},
 booktitle = {CVPR},
 year      = {2023},
 pages     = {10146-10156}
}

@article{11453592116_Chefer,
author = {Chefer, Hila and Alaluf, Yuval and Vinker, Yael and Wolf, Lior and Cohen-Or, Daniel},
title = {Attend-and-Excite: Attention-Based Semantic Guidance for Text-to-Image Diffusion Models},
year = {2023},
volume = {42},
number = {4},
journal = {ACM Transactions on Graphics},
month = {jul},
}

@inproceedings{gu2023mixofshow,
title={Mix-of-Show: Decentralized Low-Rank Adaptation for Multi-Concept Customization of Diffusion Models},
author={Gu, Yuchao and Wang, Xintao and Wu, Jay Zhangjie and Shi, Yujun and Chen Yunpeng and Fan, Zihan and others},
booktitle = {NeurIPS},
year={2023},
}

@inproceedings{gal2023an,
title={An Image is Worth One Word: Personalizing Text-to-Image Generation using Textual Inversion},
author={Rinon Gal and Yuval Alaluf and Yuval Atzmon and Or Patashnik and Amit Haim Bermano and Gal Chechik and Daniel Cohen-or},
booktitle={ICLR},
year={2023},
}

@InProceedings{Ruiz_2024_CVPR,
    author    = {Ruiz, Nataniel and Li, Yuanzhen and Jampani, Varun and Wei, Wei and Hou, Tingbo and Pritch, Yael and Wadhwa, Neal and Rubinstein, Michael and Aberman, Kfir},
    title     = {HyperDreamBooth: HyperNetworks for Fast Personalization of Text-to-Image Models},
    booktitle = {CVPR},
    month     = {June},
    year      = {2024},
    pages     = {6527-6536}
}

@InProceedings{Nam_2024_CVPR,
    author    = {Nam, Jisu and Kim, Heesu and Lee, DongJae and Jin, Siyoon and Kim, Seungryong and Chang, Seunggyu},
    title     = {DreamMatcher: Appearance Matching Self-Attention for Semantically-Consistent Text-to-Image Personalization},
    booktitle = {CVPR},
    month     = {June},
    year      = {2024},
    pages     = {8100-8110}
}

@ARTICLE{11081388,
  author={Yang, Jiahui and Di, Donglin and Ma, Baorui and others},
  journal={IEEE Transactions on Pattern Analysis and Machine Intelligence}, 
  title={TV-3DG: Mastering Text-to-3D Customized Generation with Visual Prompt}, 
  year={2025},
  volume={},
  number={},
  pages={1-18},
}

@InProceedings{Raj_2023_ICCV,
    author    = {Raj, Amit and Kaza, Srinivas and Poole, Ben and Niemeyer, Michael and Ruiz, Nataniel and others},
    title     = {DreamBooth3D: Subject-Driven Text-to-3D Generation},
    booktitle = {ICCV},
    month     = {October},
    year      = {2023},
    pages     = {2349-2359}
}

@inproceedings{poole2023dreamfusion,
title={DreamFusion: Text-to-3D using 2D Diffusion},
author={Ben Poole and Ajay Jain and Jonathan T. Barron and Ben Mildenhall},
booktitle={ICLR},
year={2023},
}

@INPROCEEDINGS{10656166,
  author={Wei, Yujie and Zhang, Shiwei and Qing, Zhiwu and Yuan, Hangjie and Liu, Zhiheng and Liu, Yu and Zhang, Yingya and Zhou, Jingren and Shan, Hongming},
  booktitle={CVPR}, 
  title={Dream Video: Composing Your Dream Videos with Customized Subject and Motion}, 
  year={2024},
  volume={}, 
  number={},
  pages={6537-6549}
}

@article{chen2023anydoor,
  title={Anydoor: Zero-shot object-level image customization},
  author={Chen, Xi and Huang, Lianghua and Liu, Yu and Shen, Yujun and Zhao, Deli and Zhao, Hengshuang},
  journal={arxiv preprint arxiv:2307.09481},
  year={2023},
}

@INPROCEEDINGS{10377881,
  author={Zhang, Lvmin and Rao, Anyi and Agrawala, Maneesh},
  booktitle={ICCV}, 
  title={Adding Conditional Control to Text-to-Image Diffusion Models}, 
  year={2023},
  volume={},
  number={},
  pages={3813-3824},
}

@InProceedings{Wu_2025_CVPR,
    author    = {Wu, Pingyu and Zhu, Kai and Liu, Yu and Zhao, Liming and Zhai, Wei and Cao, Yang and Zha, Zheng-Jun},
    title     = {Improved Video VAE for Latent Video Diffusion Model},
    booktitle = {CVPR},
    month     = {June},
    year      = {2025},
    pages     = {18124-18133}
}

@InProceedings{Henschel_2025_CVPR,
    author    = {Henschel, Roberto and Khachatryan, Levon and Poghosyan, Hayk and Hayrapetyan, Daniil and others},
    title     = {StreamingT2V: Consistent, Dynamic, and Extendable Long Video Generation from Text},
    booktitle = {CVPR},
    month     = {June},
    year      = {2025},
    pages     = {2568-2577}
}

@inproceedings{karras2022elucidating,
title={Elucidating the Design Space of Diffusion-Based Generative Models},
author={Tero Karras and Miika Aittala and Timo Aila and Samuli Laine},
booktitle={NeurIPS},
year={2022},
}

@inproceedings{he2025cameractrl,
title={CameraCtrl: Enabling Camera Control for Video Diffusion Models},
author={Hao He and Yinghao Xu and Yuwei Guo and Gordon Wetzstein and Bo Dai and Hongsheng Li and Ceyuan Yang},
booktitle={ICLR},
year={2025},
}

@inproceedings{mou2023t2i,
  title={T2i-adapter: Learning adapters to dig out more controllable ability for text-to-image diffusion models},
  author={Mou, Chong and Wang, Xintao and Xie, Liangbin and Wu, Yanze and Zhang, Jian and Qi, Zhongang and Shan, Ying and Qie, Xiaohu},
  booktitle={AAAI},
  year={2024},
}

@inproceedings{li2024motrans,
title={MoTrans: Customized Motion Transfer with Text-driven Video Diffusion Models},
author={Xiaomin Li and Xu Jia and Qinghe Wang and Haiwen Diao and mengmeng Ge and Pengxiang Li and You He and Huchuan Lu},
booktitle={ACM MM},
year={2024},
}

@article{hu2024storyagentcus,
      title={StoryAgent: Customized Storytelling Video Generation via Multi-Agent Collaboration}, 
      author={Panwen Hu and Jin Jiang and Jianqi Chen and Mingfei Han and Shengcai Liao and Xiaojun Chang and Xiaodan Liang},
      year={2024},
      journal={arXiv preprint arXiv:2411.04925},
}

@InProceedings{Yang_2025_CVPR,
    author    = {Yang, Yuanbo and Shao, Jiahao and Li, Xinyang and Shen, Yujun and Geiger, Andreas and Liao, Yiyi},
    title     = {Prometheus: 3D-Aware Latent Diffusion Models for Feed-Forward Text-to-3D Scene Generation},
    booktitle = {CVPR},
    month     = {June},
    year      = {2025},
    pages     = {2857-2869}
}

@InProceedings{Meng_2025_CVPR,
    author    = {Meng, Quan and Li, Lei and Nie{\ss}ner, Matthias and Dai, Angela},
    title     = {LT3SD: Latent Trees for 3D Scene Diffusion},
    booktitle = {CVPR},
    month     = {June},
    year      = {2025},
    pages     = {650-660}
}

@InProceedings{Hu_2025_CVPR,
    author    = {Hu, Hanzhe and Yin, Tianwei and Luan, Fujun and Hu, Yiwei and Tan, Hao and Xu, Zexiang and Bi, Sai and Tulsiani, Shubham and Zhang, Kai},
    title     = {Turbo3D: Ultra-fast Text-to-3D Generation},
    booktitle = {CVPR},
    month     = {June},
    year      = {2025},
    pages     = {23668-23678}
}

@article{wang2023lavie,
  title={LAVIE: High-Quality Video Generation with Cascaded Latent Diffusion Models},
  author={Wang, Yaohui and Chen, Xinyuan and Ma, Xin and Zhou, Shangchen and others},
  journal={International Journal of Computer Vision},
  year={2025}
}

@InProceedings{Blattmann_2023_CVPR,
    author    = {Blattmann, Andreas and Rombach, Robin and Ling, Huan and Dockhorn, Tim and Kim, Seung Wook and Fidler, Sanja and Kreis, Karsten},
    title     = {Align Your Latents: High-Resolution Video Synthesis With Latent Diffusion Models},
    booktitle = {CVPR},
    month     = {June},
    year      = {2023},
    pages     = {22563-22575}
}

@InProceedings{Ma_2025_CVPR,
    author    = {Ma, Zhiyuan and Liang, Xinyue and Wu, Rongyuan and Zhu, Xiangyu and Lei, Zhen and Zhang, Lei},
    title     = {Progressive Rendering Distillation: Adapting Stable Diffusion for Instant Text-to-Mesh Generation without 3D Data},
    booktitle = {CVPR},
    month     = {June},
    year      = {2025},
    pages     = {11036-11050}
}

@inproceedings{wu2024mixture,
title={Mixture of Lo{RA} Experts},
author={Xun Wu and Shaohan Huang and Furu Wei},
booktitle={ICLR},
year={2024},
}

@inproceedings{le_etal2023antidreambooth,
  title={Anti-DreamBooth: Protecting users from personalized text-to-image synthesis},
  author={Van Le, Thanh and Phung, Hao and Nguyen, Thuan Hoang and Dao, Quan and Tran, Ngoc N and Tran, Anh},
  booktitle={ICCV},
  pages={2116--2127},
  year={2023}
}

@InProceedings{Qin_2025_CVPR,
    author    = {Qin, Yiming and Xu, Zhu and Liu, Yang},
    title     = {Apply Hierarchical-Chain-of-Generation to Complex Attributes Text-to-3D Generation},
    booktitle = {CVPR},
    month     = {June},
    year      = {2025},
    pages     = {18521-18530}
}

@inproceedings{yoon2018lifelong,
  title={Lifelong Learning with Dynamically Expandable Networks},
  author={Yoon, Jaehong and Yang, Eunho and Lee, Jeongtae and Hwang, Sung Ju},
  year={2018},
  booktitle={ICLR},
}

@inproceedings{101007978_3_03119809_0_36,
author = {Wang, Zifeng and Zhang, Zizhao and Ebrahimi, Sayna and Sun, Ruoxi and Zhang, Han and others},
title = {DualPrompt: Complementary Prompting for Rehearsal-Free Continual Learning},
year = {2022},
booktitle = {ECCV},
pages = {631–648},
}

@inproceedings{madaan2022representational,
title={Representational Continuity for Unsupervised Continual Learning},
author={Divyam Madaan and Jaehong Yoon and Yuanchun Li and Yunxin Liu and Sung Ju Hwang},
booktitle={ICLR},
year={2022},
}

@InProceedings{Douillard_2022_CVPR,
	author    = {Douillard, Arthur and Ram\'e, Alexandre and Couairon, Guillaume and Cord, Matthieu},
	title     = {DyTox: Transformers for Continual Learning With DYnamic TOken eXpansion},
	booktitle = {CVPR},
	month     = {June},
	year      = {2022},
	pages     = {9285-9295}
}

@ARTICLE{10965859,
  author={Liu, Chenxi and Sun, Gan and Liang, Wenqi and Dong, Jiahua and Qin, Can and Cong, Yang},
  journal={IEEE Transactions on Image Processing}, 
  title={MuseumMaker: Continual Style Customization Without Catastrophic Forgetting}, 
  year={2025},
  volume={34},
  number={},
  pages={2499-2512}
}

@InProceedings{Wu_2025_CVPR_Elegant_GIFT,
    author    = {Wu, Bin and Shi, Wuxuan and Wang, Jinqiao and Ye, Mang},
    title     = {Synthetic Data is an Elegant GIFT for Continual Vision-Language Models},
    booktitle = {CVPR},
    month     = {June},
    year      = {2025},
    pages     = {2813-2823}
}

@InProceedings{wang2025dualreal,
      title={DualReal: Adaptive Joint Training for Lossless Identity-Motion Fusion in Video Customization}, 
      author={Wenchuan Wang and Mengqi Huang and Yijing Tu and Zhendong Mao},
      year={2025},
      booktitle={ICCV},
      month={October},
}

@inproceedings{ye2025stylemaster,
  title={Stylemaster: Stylize your video with artistic generation and translation},
  author={Ye, Zixuan and Huang, Huijuan and Wang, Xintao and Wan, Pengfei and Zhang, Di and Luo, Wenhan},
  booktitle={CVPR},
  pages={2630--2640},
  year={2025}
}

@INPROCEEDINGS{900901934533,
  author={Belouadah, Eden and Popescu, Adrian},
  booktitle={ICCV}, 
  title={IL2M: Class Incremental Learning With Dual Memory}, 
  year={2019},
  volume={},
  number={},
  pages={583-592},
}

@InProceedings{Guo_2025_CVPR,
    author    = {Guo, Zirun and Jin, Tao},
    title     = {ConceptGuard: Continual Personalized Text-to-Image Generation with Forgetting and Confusion Mitigation},
    booktitle = {CVPR},
    month     = {June},
    year      = {2025},
    pages     = {2945-2954},
}

@InProceedings{101007978303172630_9_25,
author="Dahary, Omer
and Patashnik, Or
and Aberman, Kfir
and Cohen-Or, Daniel",
title="Be Yourself: Bounded Attention for Multi-subject Text-to-Image Generation",
booktitle="ECCV",
year="2024",
pages="432--448",
}

@inproceedings{liu2025cclip,
title={C-{CLIP}: Multimodal Continual Learning for Vision-Language Model},
author={Wenzhuo Liu and Fei Zhu and Longhui Wei and Qi Tian},
booktitle={ICLR},
year={2025},
}

@InProceedings{Truong_2025_CVPR,
    author    = {Truong, Thanh-Dat and Prabhu, Utsav and Raj, Bhiksha and Cothren, Jackson and Luu, Khoa},
    title     = {FALCON: Fairness Learning via Contrastive Attention Approach to Continual Semantic Scene Understanding},
    booktitle = {CVPR},
    month     = {June},
    year      = {2025},
    pages     = {15065-15075}
}

\begin{IEEEbiography}[{\includegraphics[width=1in,height=1.25in,clip,keepaspectratio]{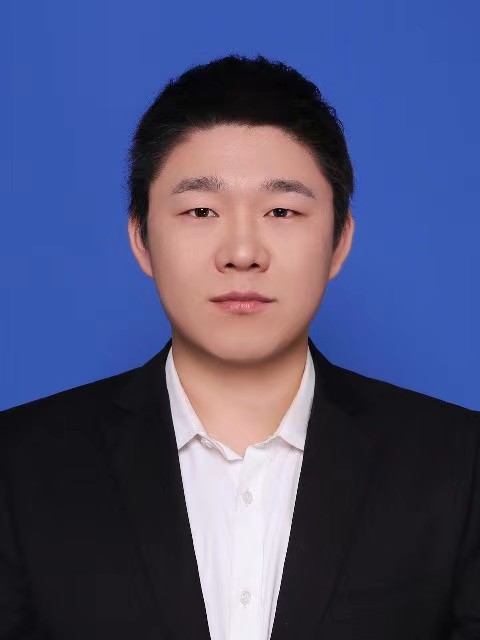}}]{Jiahua Dong} is a postdoctoral researcher in the Mohamed bin Zayed University of Artificial Intelligence, United Arab Emirates. He received the Ph.D. degree from the University of Chinese Academy of Sciences in 2024. He visited the Computer Vision Lab, ETH Zurich, Switzerland from April 2022 to August 2022, and Max Planck Institute for Informatics, Germany from September 2022 to January 2023. Before that, he received the B.S. degree from Jilin University in 2017. His current research interests include computer vision, machine learning and medical image analysis. 
\end{IEEEbiography}

\begin{IEEEbiography}[{\includegraphics[width=1in,height=1.25in,clip,keepaspectratio]{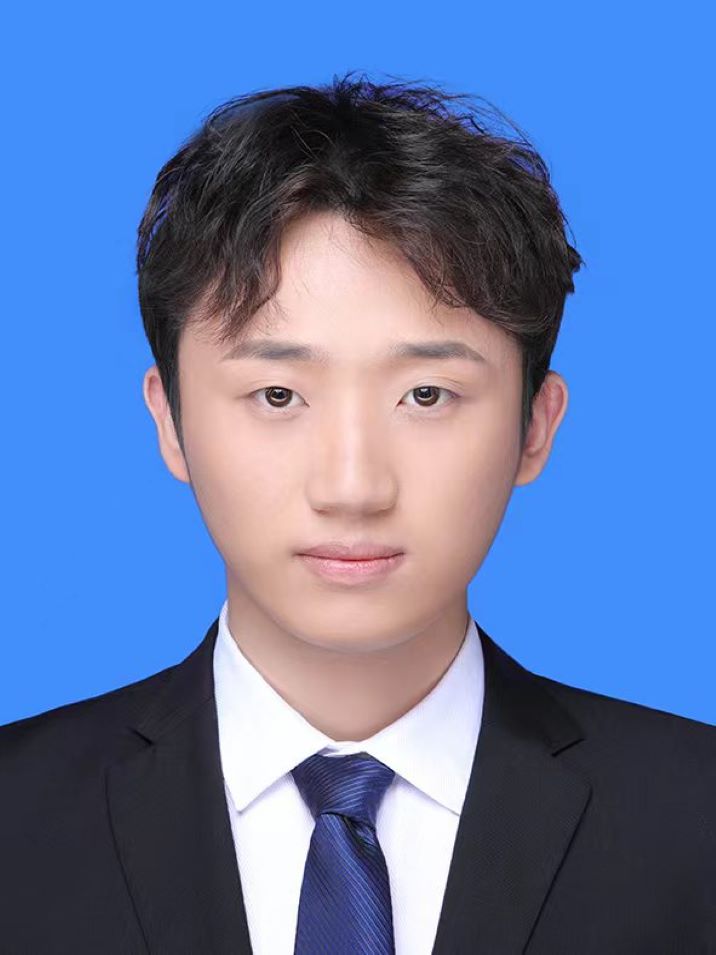}}]{Wenqi Liang} received the B.S. degree from Beijing Jiaotong University, Beijing, China, in 2022, and the M.S. degree from the State Key Laboratory of Robotics, Shenyang Institute of Automation, University of Chinese Academy of Sciences, Shenyang, China, in 2025. From 2025 to 2028, he is pursuing the Ph.D. degree with the University of Trento, Trento, Italy, as a member of the ELLIS Ph.D. program. His work has been accepted or published in top-tier venues such as NeurIPS, ICLR, TPAMI, TIP, CVPR, ICCV, and AAAI. His research interests include continual learning, generative AI, and robotic learning. 
\end{IEEEbiography}

\begin{IEEEbiography}[{\includegraphics[width=1in,height=1.25in,clip,keepaspectratio]{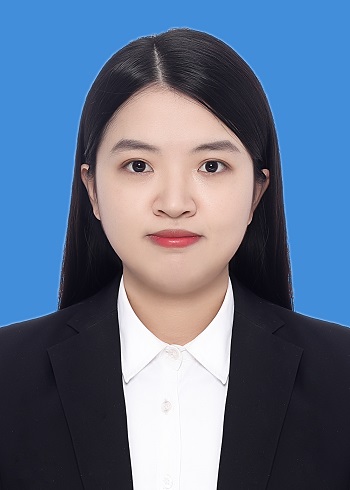}}]{Hongliu Li} is currently a postdoctoral fellow in the Department of Civil and Environmental Engineering, Hong Kong Polytechnic University. She received the B.S. degree from University of Petroleum of China in 2017, and the Ph.D degree from University of Science and Technology of China, and the Ph.D. degree from City University of Hong Kong in 2022. Her current research interests include computer vision, machine learning, pedestrian and evacuation dynamics. 
\end{IEEEbiography}

\begin{IEEEbiography}[{\includegraphics[width=1in,height=1.25in,clip,keepaspectratio]{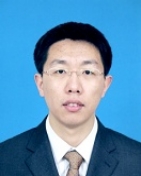}}]{Yang Cong} received the B.Sc. degree from Northeast University in 2004 and the Ph.D. degree from the State Key Laboratory of Robotics, Chinese Academy of Sciences, in 2009. From 2009 to 2011, he was a Research Fellow with the National University of Singapore (NUS) and Nanyang Technological University (NTU). He was a Visiting Scholar with the University of Rochester. He was the professor until 2023 with Shenyang Institute of Automation, Chinese Academy of Sciences. He is currently the full professor with South China University of Technology. His current research interests include robot, computer vision, and machine learning.
\end{IEEEbiography}

\begin{IEEEbiography}[{\includegraphics[width=1in,height=1.25in,clip,keepaspectratio]{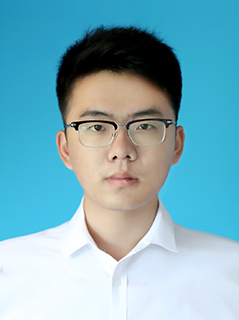}}]{Duzhen Zhang} received his B.Sc. degree from Shandong University in June 2019. He completed his Ph.D. at the Institute of Automation, Chinese Academy of Sciences in June 2024. Since September 2024, he has been a postdoctoral researcher at the Mohamed bin Zayed University of Artificial Intelligence. His current research interests include large language models, continual learning, multi-modal learning, and AI for science.
\end{IEEEbiography}

\begin{IEEEbiography}[{\includegraphics[width=1in,height=1.25in,clip,keepaspectratio]{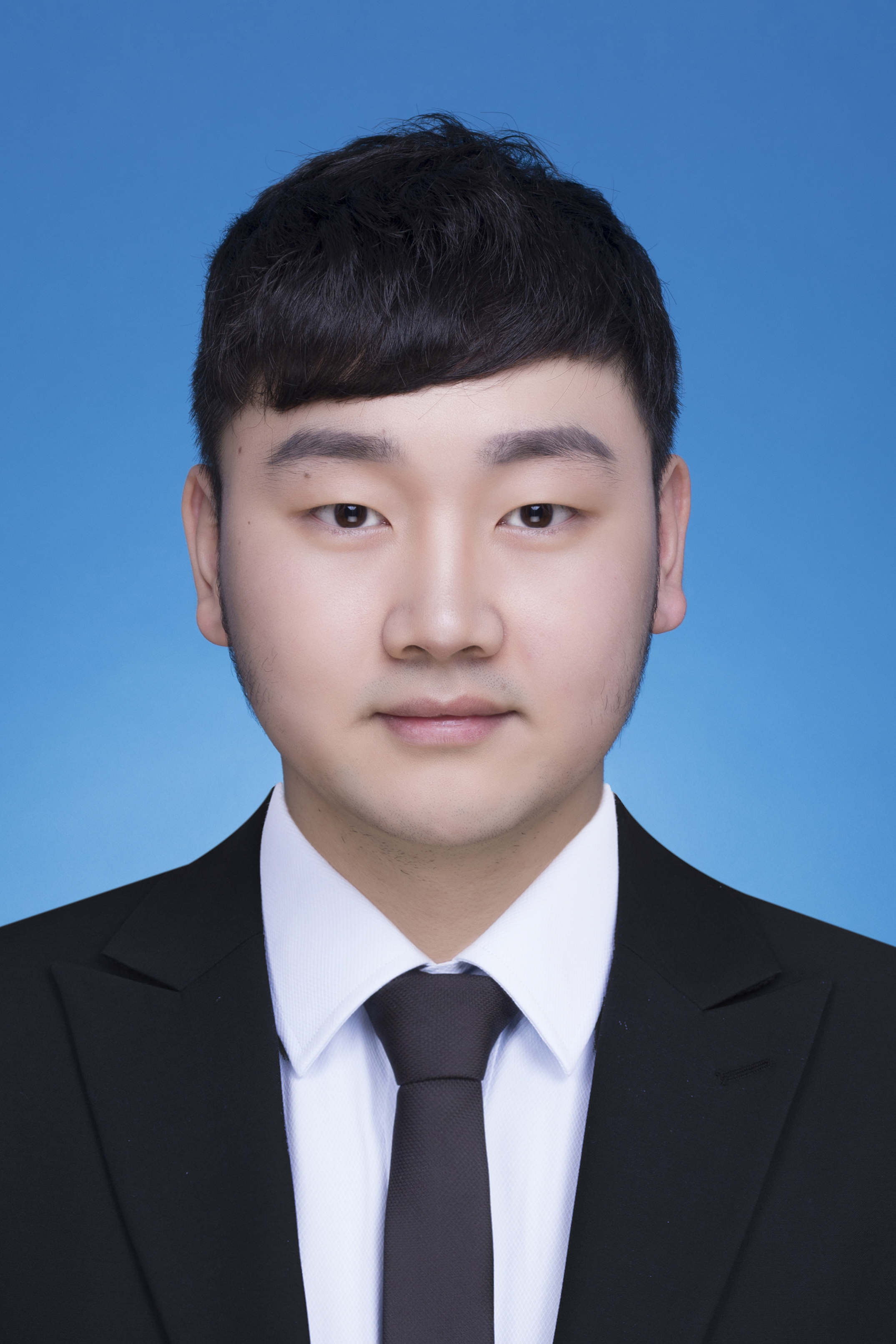}}]{Hanbin Zhao} is a research assistant professor in Zhejiang University, Hangzhou, China. He received the Ph.D. degree from the College of Computer Science and Technology, Zhejiang University in 2023. Before that, he received the B.S. degree from Zhejiang University in 2018. His current research interests include continual learning/incremental learning/lifelong learning, generative models and computer vision.
\end{IEEEbiography}

\begin{IEEEbiography}[{\includegraphics[width=1in,height=1.25in,clip,keepaspectratio]{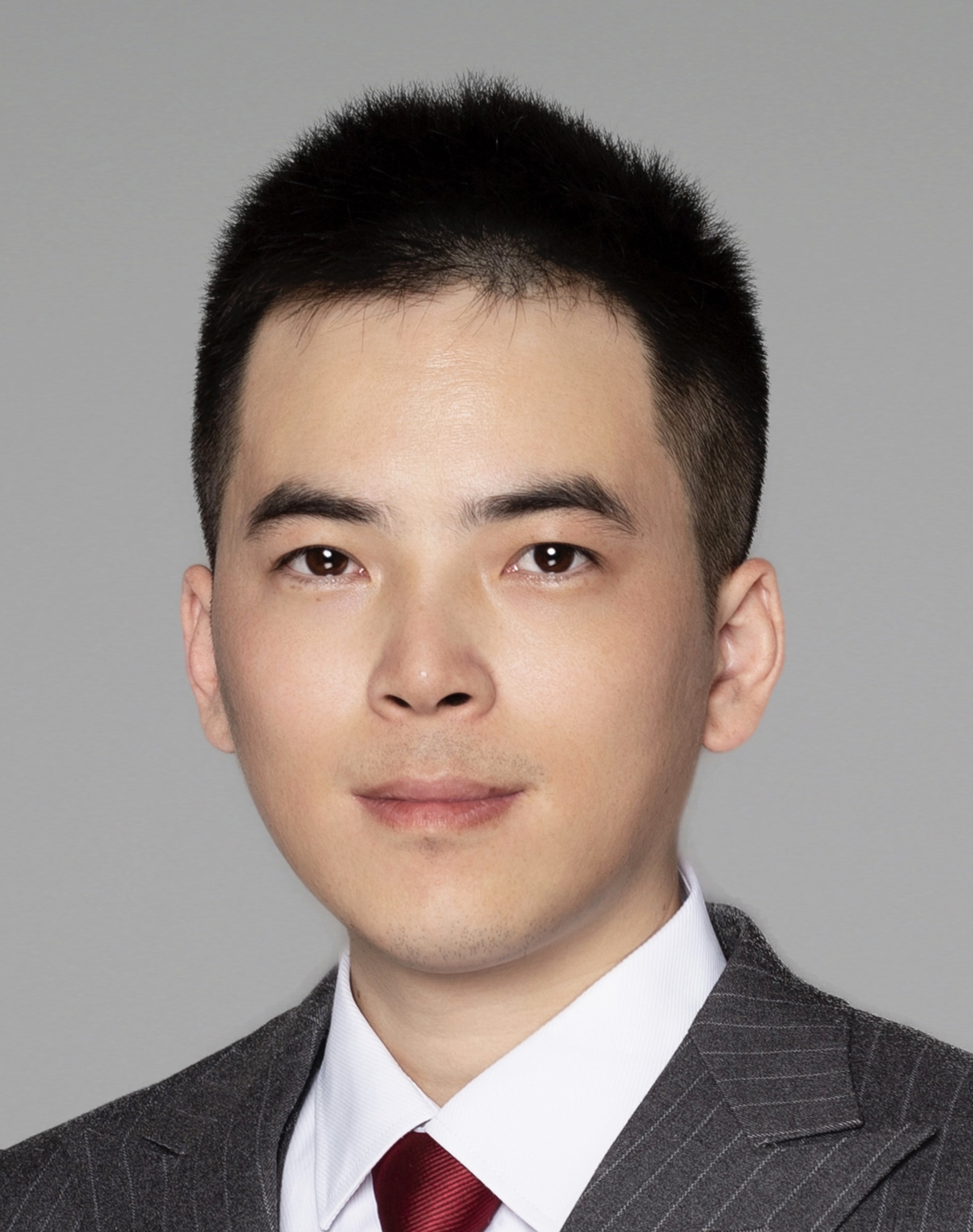}}]{Henghui Ding} received the B.E. degree from Xi'an Jiaotong University, China, in 2016. He received the Ph.D. degree from Nanyang Technological University (NTU), Singapore, in 2020. He was a Research Scientist at ByteDance, a Postdoctoral Researcher at ETH Zurich and NTU. He is currently a Professor at Fudan University, Shanghai, China. He serves as an Associate Editor for IEEE Transactions on Image Processing (TIP), and regularly serves as a Senior Area Chair or Area Chair of top conferences such as CVPR, NeurIPS, ICML, ICLR, AAAI, and ACM MM. 
His research interests include computer vision and machine learning.
\end{IEEEbiography}

\begin{IEEEbiography}[{\includegraphics[width=1in,height=1.25in,clip,keepaspectratio]{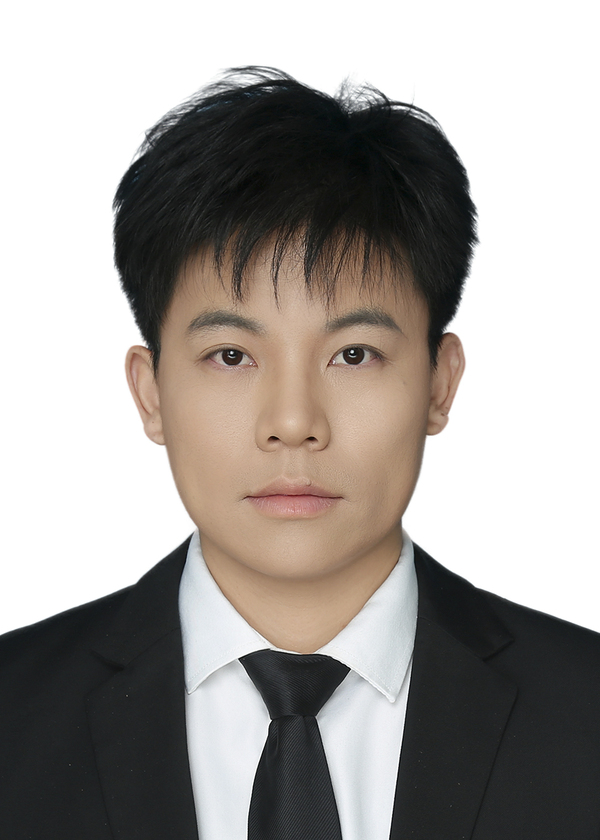}}]{Yulun Zhang} is a tenure-track associate professor at Shanghai Jiao Tong University. He was a postdoctoral researcher at Computer Vision Lab, ETH Zürich, Switzerland. He obtained the Ph.D. degree from the Department of ECE, Northeastern University, USA, in 2021. He also worked as a research fellow in Harvard University. Before that, he received the B.E. degree from the School of Electronic Engineering, Xidian University, China, in 2013 and the M.E. degree from the Department of Automation, Tsinghua University, China, in 2017. His research interests include image/video restoration and synthesis, biomedical image analysis and computational imaging.
\end{IEEEbiography}

\begin{IEEEbiography}[{\includegraphics[width=1in,height=1.25in,clip,keepaspectratio]{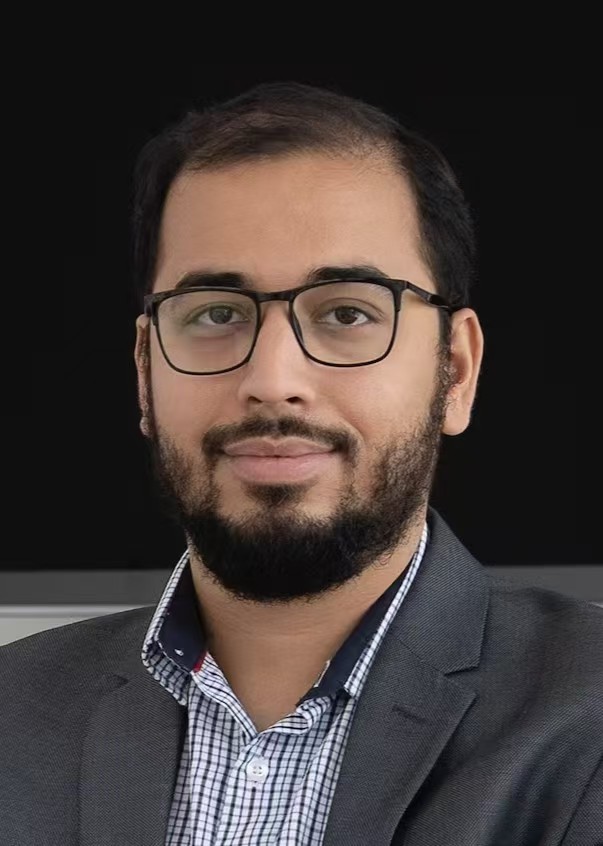}}]{Salman Khan} is an Associate Professor at Mohamed bin Zayed University of Artificial Intelligence, United Arab Emirates. He is also an adjunct faculty member in Australian National University since 2016. Before that, he was a research visitor at National ICT Australia in 2015 and also served as a research affiliate with Australian Center for Robotic Vision. 
His current interests include model generalization, adversarial robustness of deep neural networks and continual learning systems. 
\end{IEEEbiography}

\begin{IEEEbiography}[{\includegraphics[width=1in,height=1.25in,clip,keepaspectratio]{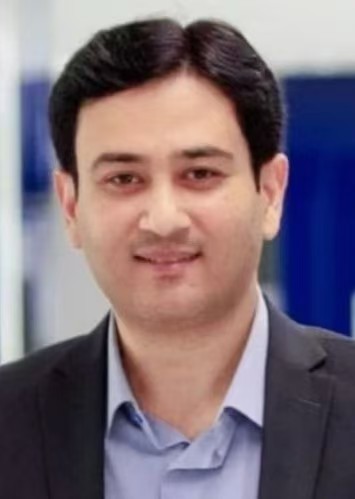}}]{Fahad Shahbaz Khan} is currently a full professor and deputy department chair of computer vision at Mohamed bin Zayed University of Artificial Intelligence, United Arab Emirates. He received the M.Sc. degree in Chalmers University of Technology, Sweden and a Ph.D. degree in Autonomous University of Barcelona, Spain. From 2012 to 2014, he was a postdoctoral fellow at Computer Vision Laboratory, Linköping University, Sweden. From 2014 to 2018, he was a research fellow at Linköping University, Sweden. His research interests include a wide range of topics within computer vision and machine learning. He serves as a regular senior program committee member for leading conferences such as, CVPR, ICCV, ECCV.
\end{IEEEbiography}

\end{document}